\documentclass[journal]{IEEEtran}

\usepackage[utf8]{inputenc} 
\usepackage[T1]{fontenc}    
\usepackage{nicefrac}       
\usepackage{microtype}      

\usepackage{color}
\usepackage{xcolor}
\definecolor{myblue}{rgb}{0.0, 0.0, 0.5}

\usepackage[pagebackref=false,colorlinks=true,bookmarks=false,linkcolor=red,citecolor=green,urlcolor=red]{hyperref}
\usepackage{url}            
\usepackage{booktabs}
\usepackage{arydshln} 
\usepackage{multirow}
\usepackage{graphicx}
\usepackage{wrapfig}
\usepackage{subcaption}
\usepackage{dsfont}
\usepackage{comment}
\usepackage{flushend}

\usepackage{amssymb, bbm} 

\usepackage{amsmath,amsfonts,bm}

\def\eqref#1{equation~\ref{#1}}

\def\1{\bm{1}}

\def\vf{{\bm{f}}}

\def\vm{{\bm{m}}}

\def\vp{{\bm{p}}}

\def\vx{{\bm{x}}}
\def\vy{{\bm{y}}}

\def\mM{{\bm{M}}}

\def\mW{{\bm{W}}}

\DeclareMathAlphabet{\mathsfit}{\encodingdefault}{\sfdefault}{m}{sl}
\SetMathAlphabet{\mathsfit}{bold}{\encodingdefault}{\sfdefault}{bx}{n}

\newcommand{\E}{\mathbb{E}}

\newcommand{\R}{\mathbb{R}}

\usepackage{cite}
\usepackage[font=normalsize,labelfont=bf]{caption}
\usepackage{overpic}

\usepackage{color,colortbl}
\definecolor{Gray}{gray}{0.9}
\newcolumntype{a}{>{\columncolor{Gray}}c}

\begin{document}

\title{Adversarial Domain Adaptation with Prototype-Based Normalized Output Conditioner}

\author{
{Dapeng Hu, Jian Liang*, Qibin Hou, Hanshu Yan, Yunpeng Chen}
\thanks{D. Hu and H. Yan are with the Department of Electrical and Computer Engineering, National University of Singapore, Singapore.}
\thanks{J. Liang is with National Laboratory of Pattern Recognition (NLPR), Chinese Academy of Sciences (CASIA), Beijing, China.}
\thanks{Q. Hou is with Nankai University, Tianjin, China.}
\thanks{Y. Chen is with Meitu Inc., Beijing, China.}
\thanks{The first two authors share co-first authorship. Correspondence to: Jian Liang <liangjian92@gmail.com>.}
}

\markboth{Transactions on Image Processing}{}
\maketitle

\begin{abstract}
Domain adversarial training has become a prevailing and effective paradigm for unsupervised domain adaptation (UDA). To successfully align the multi-modal data structures across domains, the following works exploit discriminative information in the adversarial training process, e.g., using multiple class-wise discriminators and involving conditional information in the input or output of the domain discriminator. However, these methods either require non-trivial model designs or are inefficient for UDA tasks.
In this work, we attempt to address this dilemma by devising simple and compact conditional domain adversarial training methods. We first revisit the simple concatenation conditioning strategy where features are concatenated with output predictions as the input of the discriminator.
We find the concatenation strategy suffers from the weak conditioning strength. We further demonstrate that enlarging the norm of concatenated predictions can effectively energize the conditional domain alignment.
Thus we improve concatenation conditioning by normalizing the output predictions to have the same norm of features, and term the derived method as \underline{N}ormalized \underline{O}utp\underline{U}t co\underline{N}ditioner~(NOUN). 
However, conditioning on raw output predictions for domain alignment, NOUN suffers from inaccurate predictions of the target domain.
To this end, we propose to condition the cross-domain feature alignment in the prototype space rather than in the output space. Combining the novel prototype-based conditioning with NOUN, we term the enhanced method as \underline{PRO}totype-based \underline{N}ormalized \underline{O}utp\underline{U}t co\underline{N}ditioner~(PRONOUN).
Experiments on both object recognition and semantic segmentation show that NOUN can effectively align the multi-modal structures across domains and even outperform state-of-the-art domain adversarial training methods. Together with prototype-based conditioning, PRONOUN further improves the adaptation performance over NOUN on multiple object recognition benchmarks for UDA.
\footnote{Code is available at \url{https://github.com/tim-learn/NOUN}.}

\end{abstract}


%
\IEEEpeerreviewmaketitle

\section{Introduction} \label{introduction}

\IEEEPARstart{D}{eep} neural networks (DNNs) have achieved impressive success in many fields, relying on large-scale labeled datasets. However, DNNs models have difficulty generalizing to new data with a large distribution shift. As a result, given an unlabeled dataset, the typical learning paradigm of DNNs is to perform laborious manual labeling work for the supervised training. In contrast, unsupervised domain adaptation (UDA) provides an efficient and effective approach to improving the performance of DNNs on unlabeled target data. UDA aims to leverage the knowledge of a labeled dataset (source domain) to help train a predictive model for another unlabeled dataset (target domain).
Recently, deep UDA methods have significantly improved the performance of many tasks, including image classification \cite{long2015learning,saito2017asymmetric} and semantic segmentation \cite{hoffman2016fcns,adaptseg,siban,clan,asm}, by exploiting supervisions from heterogeneous sources.

The closed-set UDA problem discussed in this paper assumes the shared label space but different data distributions across two domains. Previous methods exploit maximum mean discrepancy (MMD) \cite{gretton2008kernel,long2015learning} or other distribution statistics like central moments \cite{sun2016deep,zellinger2017central,koniusz2017domain} for domain adaptation. Recently, adversarial learning \cite{goodfellow2014generative} provides a promising alternative solution by introducing an extra binary domain discriminator. The task model in domain adversarial training aims to learn domain-invariant features through promoting domain confusion. However, the confusion of the discriminator does not always guarantee the discriminative alignment of multi-class distributions between domains. The probable case is that the multi-modal structures are misaligned between the source and target domains.

Following adversarial methods~\cite{madacls,cdan,idda,rca, dannca} are proposed to pursue discriminative domain alignment.
However, concerning these methods, \cite{madacls, idda, rca, dannca} require complex designs on either model architecture or objective function, and \cite{cdan} involves the inefficient Cartesian product between features and predictions. In contrast, in this work, we pursue simple and compact conditional domain adversarial training methods. We first revisit a simple conditional domain adversarial training strategy, i.e., the concatenation conditioning strategy~\cite{cdan}, where cross-domain feature alignment is conditioned on output predictions by concatenating features with the output of the model.
As demonstrated in \cite{cdan}, the concatenation conditioning strategy is heavily inferior to the informative multi-linear conditioning strategy. Similarly, we also have such a consistent observation across various object recognition UDA tasks.

\cite{cdan} ascribes the failure of the naive concatenation conditioning to the independence between the concatenated features and predictions. Instead, we argue that the failure is because the weak conditioning strength renders conditional domain alignment ineffective. We observe that the Euclidean norm of the feature is usually tens of times that of the concatenated output prediction.
The effect is that the contribution of concatenated predictions to the domain confusion objective is heavily ignored. Thus the discriminator would be dominantly optimized with the domain confusion loss between marginal features of two domains, making the conditional domain adversarial training ineffective. 
As expected, we observe obvious performance improvement over the naive concatenation conditioning by enlarging the norm of concatenated output predictions.
Though effective, there is no guarantee to find a suitable range for the norm of conditioned output.
Instead of directly tuning the norm of output, we first normalize the output predictions to have the same norm of features. We then tune the norm of the normalized output with a norm control factor if necessary.
This method is called Normalized OutpUt coNditioner~(NOUN).

Similar to previous conditional methods \cite{madacls, madaseg, rca, cdan}, NOUN applies raw output predictions for the conditional feature alignment across domains. However, due to the domain shift, target predictions generated by the source-trained classifier may be inaccurate, leading to possible misalignment between multi-modal structures across domains. 
We thus propose to condition the domain alignment in the prototype space rather than in the prediction space. Specifically, we leverage accurate source prototypes to project predictions into the same prototype space. The projected outputs are discriminative, simultaneously considering network predictions and semantic structures shared across domains. 
Incorporating the structure-aware conditioning into NOUN, we derive a simple yet strong UDA method called PROtotype-based Normalized OutpUt coNditioner~(PRONOUN).

Experimental results on both object recognition and semantic segmentation tasks demonstrate the advantages of the proposed methods over previous state-of-the-arts~\cite{madacls, cdan, clan, adaptpatch, siban} of domain adversarial training methods. The main contributions of this work are summarized as follows:

\begin{table*}[t]
    \centering
    \caption{Comparison between related conditional domain adversarial training methods and our methods, i.e., NOUN and PRONOUN, where DANN \cite{dann, dannjmlr} is as a common baseline. $D$ means the domain classifier, i.e., the discriminator. 
    We denote the feature dimension as $d$ and the class number as $c$. In the first column, `Input dimension of $D$': the input feature dimension of the discriminator; `Output dimension of $D$': the output prediction dimension of the discriminator; `Number of $D$': the number of discriminators involved in the method; 
    ``Sensitivity to pseudo-labels": the sensitivity of the method to pseudo-labels. We classify this sensitivity into three degrees: low, medium, and high. Methods not using pseudo labels are classified as the `low' sensitivity. Methods with `high' sensitivity directly rely on pseudo labels to perform class-wise feature alignment across domains. Methods with the `medium' sensitivity implicitly leverage pseudo labels for conditional domain alignment.
    }
    \scalebox{1.0}{
    \begin{tabular}{lccccccccc}
    \toprule
             &DANN \cite{dann} &MADA \cite{madacls} &CDAN \cite{cdan} & DANN-CA \cite{dannca} &RCA \cite{rca} &IDDA \cite{idda} &NOUN &PRONOUN\\
    \midrule
    Input dimension of $D$ & $d$ & $d$ & $cd$ & $d$ & $d$ & $d$ & $c+d$ & $2d$ \\ 
    Output dimension of $D$ & $1$ & $1$ & $1$ & $c+1$ & $2c$ & $c+1$ & $1$ & $1$ \\
    Amount of $D$ & $1$ & $c$ & $1$ & $0$ & $1$ & $1$ & $1$ & $1$ \\ 
    Sensitivity to pseudo-labels& low & high & medium & low & high & low & medium & medium \\
    \bottomrule
    \end{tabular}
    }
    \label{tab:cmp_adv}
\end{table*}

\begin{itemize}
  \item We revisit the concatenation conditioning strategy for conditional domain alignment and provide a novel explanation to its previous failure in~\cite{cdan}, i.e., the smaller norm of conditioned output predictions.
  \item We propose a novel and effective conditional domain adversarial training method NOUN. NOUN enlarges the norm of conditioned output by normalizing the output to have the same norm with features and involving an extra norm control factor for better flexibility.
  \item To deal with the possible misalignment due to inaccurate pseudo-labels, we further propose a novel prototype-based conditioning strategy and combine it with NOUN to be a strong UDA method called PRONOUN.
  \item Both NOUN and PRONOUN are simple and compact. The effectiveness of our methods is further validated on multiple cross-domain object recognition benchmarks. In addition, NOUN is shown to be generic and competitive for synthetic-to-real semantic segmentation tasks.
  \end{itemize} 

The rest of this paper is organized as follows. In Section \ref{sec:related_works}, we review related domain adaptation works. In Section \ref{sec:method}, we first introduce the settings of UDA and the vanilla domain adversarial training solution \cite{dann}. Then we revisit the naive concatenation conditioning strategy and explain why it fails in conditional domain alignment. Finally, we introduce NOUN and PRONOUN. In Section \ref{sec:exp}, we conduct extensive experiments on cross-domain object recognition benchmarks and synthetic-to-real semantic segmentation benchmarks to evaluate NOUN and PRONOUN. Besides, we provide extensive ablations and analyses on our methods. In Section \ref{sec:conclusion}, we briefly summarize our work and discuss the future work.

\section{Related Work} \label{sec:related_works}

\subsubsection{Unsupervised Domain Adaptation} 
Unsupervised domain adaptation (UDA) is first modeled as the covariate shift problem \cite{shimodaira2000improving} where different domains have different marginal distributions but identical conditional distributions.
At early times, researchers address UDA by instance re-weighting~\cite{dudik2006correcting,huang2007correcting}, feature space alignment~\cite{gong2012geodesic, fernando2013unsupervised, sun2016return} or domain-invariant feature transformation \cite{pan2010domain,long2013transfer,herath2017learning}.
In the past years, various deep UDA works \cite{long2015learning,dann,adda,cdan,saito2018maximum,adaptseg} have been developed and have significantly boosted the UDA performance in visual tasks. Generally, they can be classified into discrepancy-based methods and adversary-based methods.
Discrepancy-based methods address the dataset shift by mitigating specific discrepancies defined on different layers of a shared model between domains, e.g., \cite{tzeng2014deep,long2015learning,long2017deep} adopt MMD measure and \cite{zellinger2017central,koniusz2017domain} leverage central moments.
Recently, adversarial learning~\cite{goodfellow2014generative} has become an increasingly popular solution to domain adaptation problems. It aims to learn domain-invariant features by introducing an extra domain discriminator to promote domain confusion.
\cite{dann} optimizes a minimax objective via inserting a gradient reversal layer between the classification network and the discriminator. Differently, \cite{adda} adopts two independent optimization objectives with inverted labels following GAN~\cite{goodfellow2014generative}.

\subsubsection{Pseudo-labeling}
UDA can be also regarded as a semi-supervised learning (SSL) problem where unlabeled data from the same domain are replaced by data from another target domain.
As a simple yet effective technique for SSL, pseudo-labeling~\cite{lee2013pseudo} has been widely used in UDA methods. For example, \cite{saito2017asymmetric,li2019bidirectional,atdoc} exploit intermediate pseudo-labels with tri-training and self-training, respectively. Recently, curriculum learning \cite{choi2019pseudo}, self-paced learning \cite{zou2018unsupervised} and re-weighting schemes \cite{cdan} are further leveraged to protect the domain adaptation from inaccurate pseudo-labels.

\subsubsection{Conditional Domain Adaptation}
Apart from being explicitly used to re-train the classification network, pseudo-labels can also be exploited to enhance the feature-level domain alignment.
\cite{long2013transfer,zhang2017joint} show that pseudo-labels can help mitigate the joint distribution discrepancy via minimizing multiple class-wise MMD measures.
\cite{long2017deep} proposes to align joint distributions of multiple domain-specific layers across domains based on a joint MMD criterion.
Recent methods resort to conditional domain adversarial training for better alignment of multi-modal structures across domains. \cite{madaseg, madacls} leverage pseudo-labels and multiple class-wise domain discriminators to enable fine-grained alignment between domains.
In contrast, \cite{cdan} conditions the adversarial domain alignment on discriminative information via the outer product between features and predictions.
\cite{rca, idda} introduce a classification-aware domain discriminator. Especially, \cite{rca} imposes the fine-grained adversarial loss with pseudo-labels and \cite{idda} performs the adversarial training asymmetrically between semantic labels and the domain labels. Different from \cite{idda}, \cite{dannca} replaces the original discriminator with a domain-aware classifier and achieves the asymmetrical adversarial training within the exact classifier. \cite{dada} improves \cite{dannca} by introducing discriminative interaction between domain predictions and semantic predictions.

Concerning the above works, the conditional domain adversarial training methods \cite{dann,madacls,madaseg,cdan,idda,rca,dannca} are the most related to ours. We thus compare NOUN and PRONOUN with these works in Table \ref{tab:cmp_adv} in terms of the implementation of the discriminator and the sensitivity to pseudo-labels. The comparison shows that NOUN and PRONOUN are compact, easy to implement, and reliable conditional domain adversarial training methods.

\section{Method}
\label{sec:method}

In this section, we begin with the basic settings of UDA as well as the introduction of domain adversarial training \cite{dann}. Then we revisit a naive conditional domain adversarial training method, i.e., the concatenation conditioning strategy mentioned in~\cite{cdan} and analyze its failure in domain alignment with multi-modal structures. We then propose NOUN, a simple yet effective concatenation conditioning method. Finally, we propose a novel prototype-based conditioning strategy to mitigate the misalignment induced by inaccurate predictions. 

\subsection{Preliminaries}
\label{sec:preliminaries}

In a vanilla UDA task, we are given label-rich source domain data $\{(\vx_{s}^{i}, \vy_{s}^{i})\}_{{i=1}}^{n_s}$ sampled from the joint distribution $P_s(\vx_s, \vy_s)$ and unlabeled target domain data $\{\vx_{t}^{i}\}_{{i=1}}^{n_t}$ sampled from the joint distribution $Q_t(\vx_t, \vy_t)$, where $\vx_{s}^{i} \in \mathcal{X}_S$ and $\vy_{s}^{i} \in \mathcal{Y}_S$ denote an image and its corresponding label from the source domain dataset and $\vx_{t}^{i} \in \mathcal{X}_T$ denotes an image from the target domain dataset and $P_s \ne Q_t$. The goal of UDA is to learn a discriminative model from $\mathcal{X}_S$, $\mathcal{Y}_S$, and $\mathcal{X}_T$ to predict labels for unlabeled target samples $\mathcal{X}_T$.

Domain adversarial training methods aim to reduce the domain discrepancy through learning domain-invariant features. Then the discriminative classifier trained by source labeled data can be freely applied to target unlabeled data. A prevailing domain adversarial training framework proposed in \cite{dann, dannjmlr} consists of a feature extractor network $G$, a classifier network $F$, and a discriminator network $D$. 
Given an image $\vx$, we denote the feature vector extracted by $G$ as $\vf = G(\vx) \in \R^{d}$ and the probability prediction obtained by $F$ as $\vp = F(\vf) \in \R^{c}$, where $d$ means the feature dimension and $c$ means the number of classes. 
The vanilla domain adversarial training method dubbed DANN in \cite{dann, dannjmlr} can be formulated as optimizing the following minimax problem:
\begin{equation} 
\label{eq:dann_obj}
\underset{G, F}{\min} \; \underset{D}{\max} \; \mathcal{L}_{y}(G,F) - \lambda_{adv} \mathcal{L}_{dann}(G,D),
\end{equation}

\begin{equation}
\label{eq:dann_adv}
{\mathcal{L}_{dann}(G,D)} = -\E_{{{\vx}_s^i} \sim P_s}\log[{D}({\vf_s^i})] -\E_{{{\vx}_t^j} \sim Q_t}\log[1 - {D}({\vf_t^j})],
\end{equation}

\begin{equation}
\label{eq:ly}
{\mathcal{L}_{y}(G,F)} = -\E_{{({\vx}_s^i}, {{\vy}_s^i}) \sim P_s} \ {{{\vy}_s^i}}^T\log({\vp_s^i}), \; \; {\vp_s^i}=F(G({{\vx}_s^i})),
\end{equation}
where the binary classifier $D: \R^{d} \to [0,1]$ predicts the domain assignment probability for the input features $f$, $\mathcal{L}_y(G,F)$ is the cross-entropy loss of labeled source data as for the classification task, and $\lambda_{adv}$ is the coefficient for the adversarial loss. 

\subsection{Why does the naive concatenation conditioning strategy fail?}

\label{sec:cada}
DANN does not explicitly consider discriminative information in cross-domain feature alignment. Therefore, it cannot guarantee the effective alignment of domains with multi-modal structures, even if the domain discriminator is fully confused. Motivated by conditional GANs \cite{isola2017image, odena2017conditional}, \cite{cdan} proposes to leverage the discriminative network prediction $\vp$ for conditional domain adversarial training. A simple baseline conditioning strategy named DANN-[f, p] in \cite{cdan} is to condition the cross-domain feature alignment on network predictions by directly concatenating features with predictions, i.e., $\vf \oplus \vp$. Therefore, the adversarial loss of this naive concatenation conditioning method, i.e., DANN-[f, p], can be formulated as the following:

\begin{equation}
\label{eq:fg_loss}
\begin{split}
{{\mathcal{L}_{dann-fp}}(G, D)} = &-\E_{{{\vx}_s^i} \sim P_s}\log[{D}({\vf_s^i} \oplus {\vp_s^i})]\\
&-\E_{{{\vx}_t^j} \sim Q_t}\log[1 - {D}({\vf_t^j} \oplus {\vp_t^j})].
\end{split}
\end{equation}

\begin{figure}[!htbp]
	\centering
	\footnotesize
	\setlength\tabcolsep{1mm}
	\renewcommand\arraystretch{0.1}
	\begin{tabular}{cc}
		\includegraphics[width=0.48\linewidth,trim={0.0cm 0.0cm 0.0cm 0.0cm}, clip]{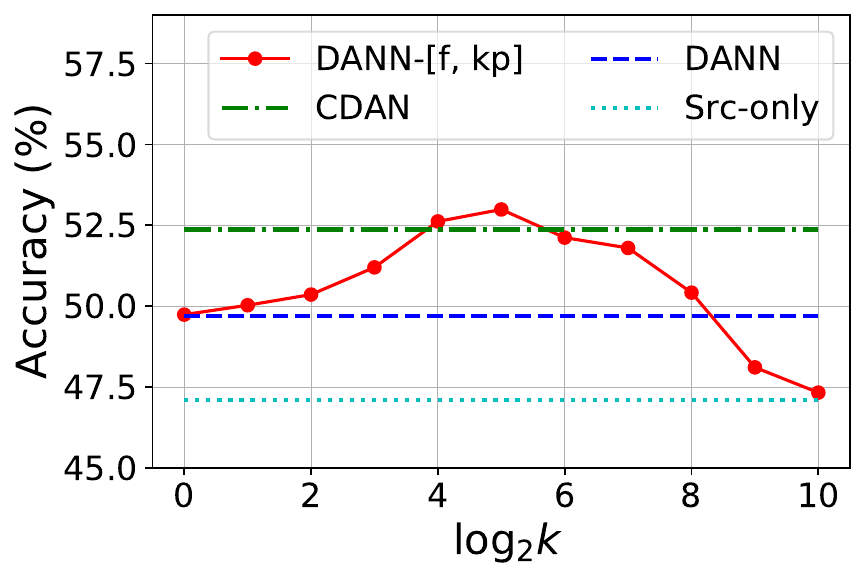} &
		\includegraphics[width=0.48\linewidth,trim={0.0cm 0.0cm 0.0cm 0.0cm}, clip]{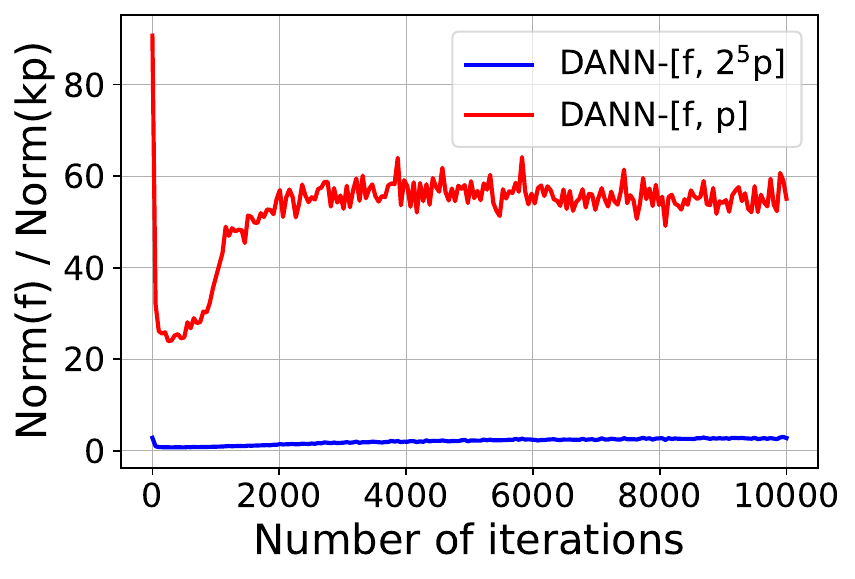} \\
		~\\
		(a) Norm control factor  & (b) Norm ratio
	\end{tabular}
	\caption{Results for a 65-way classification UDA task Ar$\to$Cl on \emph{Office-Home}.  `$k$' is the norm control factor to enlarge the norm of conditioned output. 
	`Src-only' means using the source-trained model without domain adaptation. (a) illustrates how the norm of the conditioned output affects the adaptation performance of the concatenation conditioning method. (b) shows the ratio between the Euclidean norm of two concatenated parts during the training.}
	\label{fig:intro_norm}
\end{figure} 

Compared with the adversarial loss of DANN in Eq.~(\ref{eq:dann_adv}), the concatenation operation $\vf \oplus \vp$ explicitly considers discriminative predictions for conditional feature alignment between domains. However, the naive concatenation conditioning strategy is demonstrated to be unsuccessful for cross-domain alignment with multi-modal structures in \cite{cdan}. 
Furthermore, \cite{cdan} accounts the failure of the naive concatenation conditioning for the independence between $\vf$ and $\vp$ in $\vf \oplus \vp$.
A similar observation for DANN-[f, p] in our experiments is shown in Fig. \ref{fig:intro_norm}(a) when the norm control factor $k$ is set as $1$, i.e., DANN-[f, p] does not outperform DANN. 
 
However, experiments in Fig.~\ref{fig:intro_norm}(a) show that if we simply introduce a norm control factor $k$ to re-weight the network prediction for the concatenation conditioning, the adaptation performance would vary a lot with different $k$. Since this re-weighting operation only changes the norm of the concatenated prediction $\vp$, it is natural to take a closer look at the norm of the two concatenated parts during the model training. We illustrate the norm ratio between $\vf$ and $\vp$ in Fig.~\ref{fig:intro_norm}(b) for the naive concatenation conditioning ($k$ equals to $1$) and the effective re-weighted concatenation conditioning case ($k$ equals to $32$). It is noteworthy that for the naive concatenation strategy, the norm ratio keeps more than $50$.  While for the re-weighted concatenation strategy, the norm ratio keeps small and close to $1$. In addition, as shown in Fig.~\ref{fig:intro_norm}(a), re-weighted with moderate norm factors, the concatenation conditioning can outperform the informative multi-linear conditioning strategy implemented via the outer product between $\vf$ and $\vp$~\cite{cdan}.

\begin{figure}[!htbp]
	\centering
	\footnotesize
	\setlength\tabcolsep{1mm}
	\renewcommand\arraystretch{0.1}
	\begin{tabular}{c}
	    \includegraphics[width=0.96\linewidth,trim={0.0cm 0.0cm 0.0cm 0.0cm}, clip]{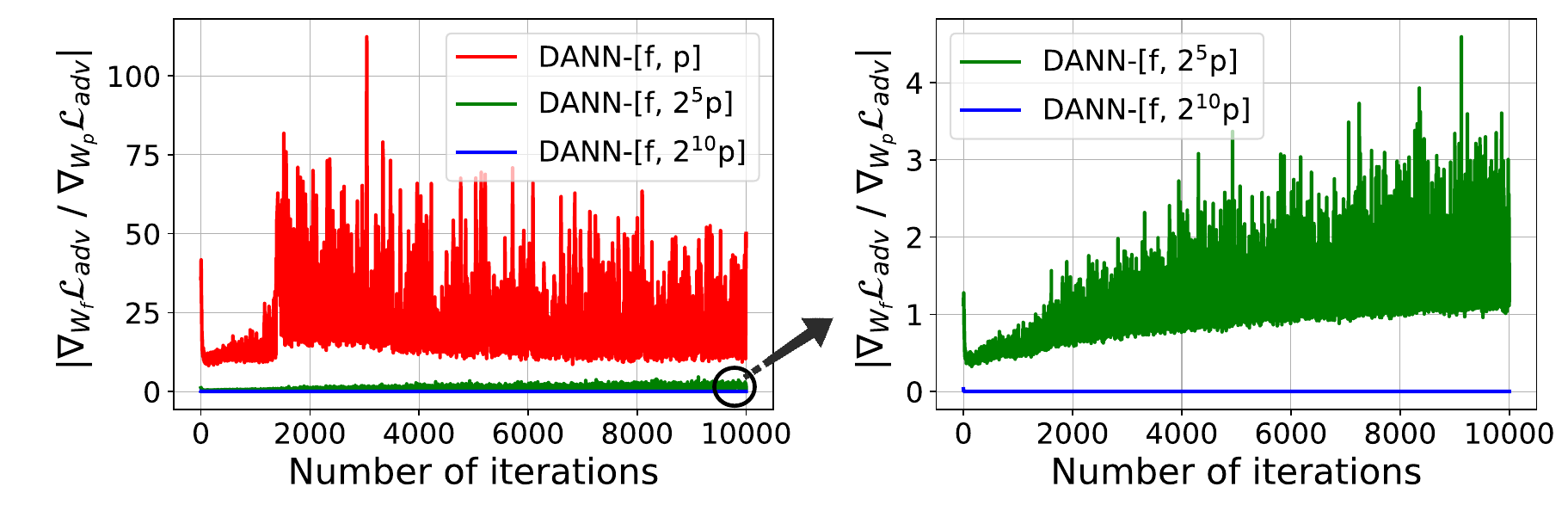}\\
		~\\
	\end{tabular}
	\caption{How does the norm of the conditioned output affect the optimization of $D$? $| {\nabla_{W_f} {\mathcal{L}_{adv}}} |$ denotes the absolute value of the mean of gradient matrix of $\mW_f$, i.e., ${\nabla_{\mW_f} {\mathcal{L}_{adv}}}$. $| {\nabla_{W_f} {\mathcal{L}_{adv}}}~/~{\nabla_{W_p} {\mathcal{L}_{adv}}} |$ thus measures the ratio between gradient of $\mW_f$ and gradient of $\mW_p$.} 
	\label{fig:grad_vis}
\end{figure}

The above observations inspire us to rethink the reason for the failure of the naive concatenation conditioning strategy. To figure out how the norm control factor `$k$' affects the conditional domain adversarial training, we take a closer look at the optimization process. For concatenation conditioning strategy, denote the output dimension of the first layer of $D$ by $d_e$ and the weight of the first layer of $D$ by $\mW_{1} = \mW_f \oplus \mW_g$, where $\mW_f \in \R^{d \times d_e}$ accepts the input feature $\vf$ and $\mW_p \in \R^{c \times d_e}$ accepts the conditioned output $\vp$. Taking $\vf \oplus k\vp$ as the input of $D$ and $\mathcal{L}_{adv}$ as the adversarial loss, we compare the gradients from domain confusion loss to $\mW_f$ and $\mW_g$ under different $k$ values in Fig.~\ref{fig:grad_vis}. We analyze the concatenation conditioning with different $k$ values according to the performance in Fig.~\ref{fig:intro_norm}(a) and the optimization process in Fig.~\ref{fig:grad_vis}. For naive concatenation conditioning ($k=1$), the gradient of $\mW_f$ is tens of times that of $\mW_p$. The marginal features $\vf$ dominate the optimization of $D$, and conditioned output predictions $\vp$ are ignored, making the conditional adversarial learning degenerate to the `DANN' case. 

While for conditioning with extremely large output norm ($k=2^{10}$), the conditioned output predictions $\vp$ dominate the discriminator optimization, and features $vf$ are ignored, making the conditional adversarial learning degenerate to the `Src-only' case. Conditioning is beneficial for the domain adaptation only when the norm of conditioned output is moderate, e.g., $k=2^5$.

To sum up, the success of re-weighted concatenation conditioning indicates that conditioning on the stop-gradient concatenated predictions, the discriminator can capture the multi-modal structures for effective conditional domain alignment. Therefore, it seems that the independence between $\vf$ and $\vp$ in the concatenation strategy is not the main obstacle blocking effective conditioning. 
Instead, the devil is in the norm of the concatenated prediction.
In fact, since $\vp$ means the probability predictions for c categories, the Euclidean norm of $\vp$ is between $\frac{1}{c}$ and $1$. In contrast, the feature $\vf$ is usually a high-dimension vector. The feature norm of $\vf$ is probably tens of times that of $\vp$. The effect of this overwhelming norm ratio between $f$ and $p$ is that the domain confusion objective neglects the concatenated prediction, 
leading to ineffective conditioning. We argue this is the main reason for the failure of the naive concatenation conditioning strategy.

\begin{figure*}[!htbp]
\centering
\begin{overpic}[scale=0.6]{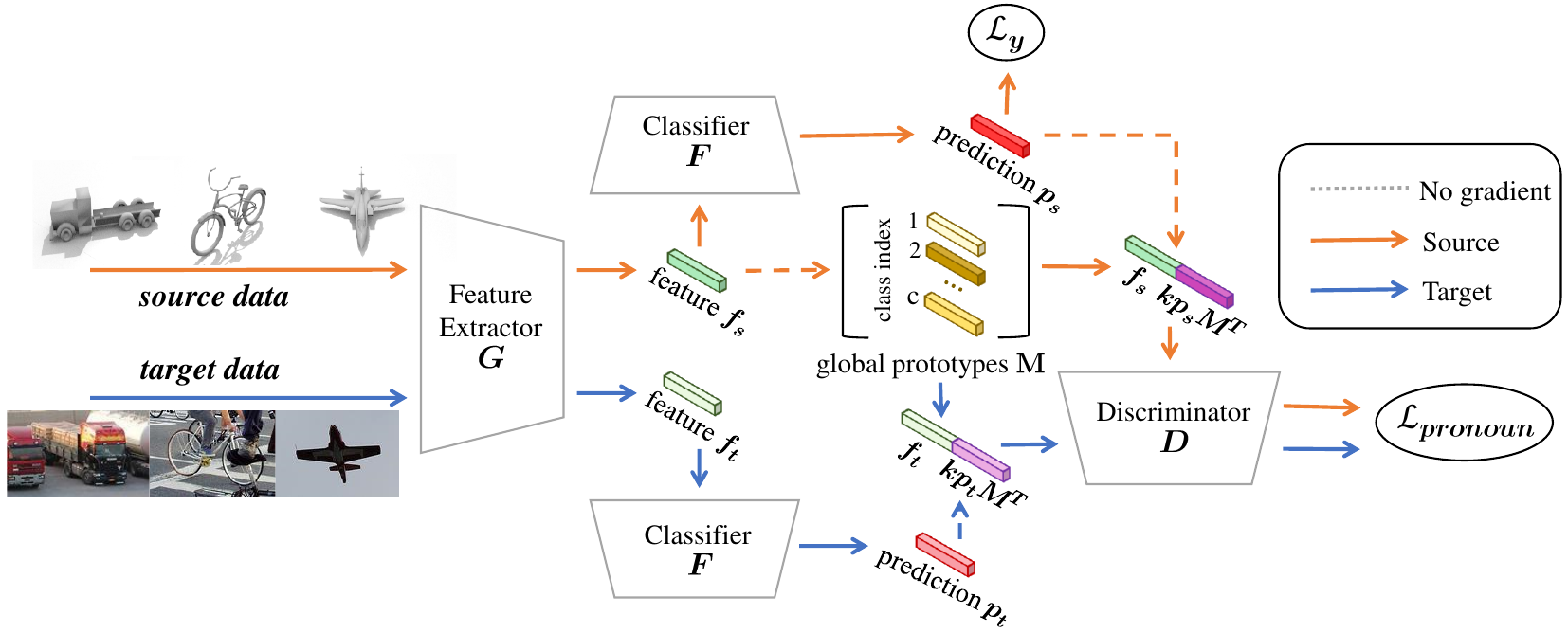}
	\end{overpic}
	\caption{The framework of PRONOUN. PRONOUN conditions the cross-domain feature alignment on semantic structure-aware predictions via the concatenation conditioning. Note that concatenated output predictions are detached without gradient from the binary discriminator $D$ and the global prototypes $\mM$ are collected from source domain by Eq.~(\ref{eq:batch_ctr}) and Eq.~(\ref{eq:ema_ctr}).}
    \label{fig:net}
\end{figure*}

\subsection{Normalized OutpUt coNditioner~(NOUN)}
\label{sec:SDAN}

Since the norm of features varies a lot for different tasks, it is non-trivial to find suitable $k$ for a given UDA task, as shown in Fig.~\ref{fig:intro_norm} (a). Besides, Fig.~\ref{fig:intro_norm} (b) shows that the norm ratio between two concatenated parts plays an important role in effective conditional feature alignment across domains. Therefore, we normalize $\vp$ to have the same Euclidean norm of $\vf$ using Eq.~(\ref{eq:norm_p}). To keep flexible, we retain $k$ as an extra norm control factor. Then $k$ would represent the norm ratio between $\vf$ and $\vp$. Therefore, it will be friendly to decide the value of $k$ within a normal range for different UDA tasks. 
In this way, we obtain an effective concatenation conditioning method called Normalized OutpUt coNditioner~(NOUN).
The adversarial loss of NOUN is formulated as:

\begin{equation} 
\label{eq:norm_p}
\widehat{\vp}_{a} = \frac{{\| \vf_a \|}_2}{{\| \vp_a \|}_2} \vp_a, ~a \in \{s,t\},
\end{equation}

\begin{equation}
\label{eq:sdan_loss}
\begin{split}
{{\mathcal{L}_{noun}}(G, D)} = &-\E_{{{\vx}_s^i} \sim P_s}\log[{D}({\vf_s^i} \oplus {k \widehat{\vp}_s^i})]\\
&-\E_{{{\vx}_t^j} \sim Q_t}\log[1 - {D}({\vf_t^j} \oplus {k \widehat{\vp}_t^i})],
\end{split}
\end{equation}
where $k$ is a norm control factor and the default value is $1$.

\subsection{PROtotype-based Normalized OutpUt coNditioner~(PRONOUN)}
\label{sec:ssdan}
Effective conditioning on network output is supposed to achieve more discriminative domain alignment than marginal domain feature alignment. However, output predictions of target data are probably inaccurate due to the domain shift. As a result, conditioning the cross-domain feature alignment directly on pseudo-labels may damage the domain adaptation performance. \cite{cdan} thus proposes the entropy conditioning strategy where entropy is leveraged to explicitly distinguish out easy-to-transfer samples towards safe adaptation. Differently, we propose to alleviate the misalignment of multi-modal structures between domains from the perspective of the conditioning space.

In UDA, a model is trained with labeled source data, making source predictions nearly one-hot vectors. 
For better illustration, we consider using one-hot target predictions for the conditional domain alignment. Though discriminative, the multiple modes embedded in the output space are generally independent and orthogonal to each other. If conditioning in the output space, concatenated output predictions would directly dominate the domain alignment. Guided by inaccurate pseudo-labels, the cross-domain alignment would be misleading. Note target predictions contain both uncertain predictions and certain but inaccurate ones. The entropy conditioning~\cite{cdan} thus cannot fully mitigate the misalignment. Specifically, imposing domain alignment with confident instances of two different categories would heavily damage the feature learning in the target domain.
Even if the leveraged target predictions are soft probabilities, the output space only encodes implicit and faint information of semantic structures, which helps little in mitigating the misalignment induced by inaccurate pseudo labels.

Therefore, we propose to condition the domain alignment on prototypes, which explicitly encode semantic structures. Semantic correlations embedded in the semantic structures are promising to alleviate the misalignment induced by inaccurate conditioned outputs.
We represent the domain-shared semantic structures with source global prototypes. Specifically, we update global prototypes in training by the exponential moving average strategy (ema) over the batch-level prototypes. We formulate the process of obtaining reliable prototypes as follows:

\begin{equation}
\label{eq:batch_ctr}
{\mM}_{\mathrm{batch}}=[\vm_1  \cdots \vm_e  \cdots \vm_c], \ {\vm_e} = \frac{\sum\nolimits_i^b {\mathds{1}({\vy_i^s}=e)}{\vf_s^i}}{\sum\nolimits_i^b {\mathds{1}({\vy_i^s}=e)}},
\end{equation}

\begin{equation}
\label{eq:ema_ctr}
{\mM} = {\lambda_{\mathrm{ema}}} {\mM}+(1-{\lambda_{\mathrm{ema}}}) {\mM}_{\mathrm{batch}},
\end{equation}
where $\mM_{batch} \in \R^{d \times c}$ denotes batch-level prototypes, $\vm_e \in \R^{d}$ denotes the prototype of class $e$, $\lambda_{ema} \in [0, 1)$ is a coefficient and $\mM \in \R^{d \times c}$ denotes global prototypes.

Taking source prototypes as the basis and projecting output predictions into the same prototype feature space, we would obtain discriminative yet semantic structure-aware predictions for conditioning. 
Seamlessly combining NOUN with prototype-based conditioning, we obtain the PROtotype-based Normalized OutpUt coNditioner~(PRONOUN). The domain adversarial training objective of PRONOUN is shown as below:

\begin{equation}
\label{eq:ssdan_loss}
\begin{split}
{{\mathcal{L}_{pronoun}}(G, D)} = &-\E_{{{\vx}_s^i} \sim P_s}\log[{D}({\vf_s^i} \oplus {k \widehat{\vp}_s^i} {\mM}^T )]\\
&-\E_{{{\vx}_t^j} \sim Q_t}\log[1 - {D}({\vf_t^j} \oplus {k \widehat{\vp}_t^i} {\mM}^T)].
\end{split}
\end{equation}

Fig.~\ref{fig:net} illustrates the framework of PRONOUN. Without complicated network designs or complex optimization objectives, both NOUN and PRONOUN are simple and compact solutions to cross-domain feature alignment with multi-modal structures.

\section{Experiments}
\label{sec:exp}

\setlength{\tabcolsep}{4.0pt}
\begin{table*}[h]
	\centering
	\small
	\caption{Recognition accuracies (\%) on \emph{Office-Home} via ResNet-50. \textbf{bold}: Best.}
	\label{tab:home}
	\centering
	\resizebox{0.95\textwidth}{!}{$
		\begin{tabular}{lcccccccccccca}
		\toprule
		Methods (Source$\to$Target)   & Ar$\to$Cl & Ar$\to$Pr & Ar$\to$Re & Cl$\to$Ar & Cl$\to$Pr & Cl$\to$Re & Pr$\to$Ar & Pr$\to$Cl & Pr$\to$Re & Re$\to$Ar & Re$\to$Cl & Re$\to$Pr & Avg. \\
		\midrule
		ResNet-50~\cite{he2016deep} & 34.9  & 50.0  & 58.0  & 37.4  & 41.9  & 46.2  & 38.5  & 31.2  & 60.4  & 53.9  & 41.2  & 59.9  & 46.1 \\
		DANN~\cite{dann}      & 45.6  & 59.3  & 70.1  & 47.0  & 58.5  & 60.9  & 46.1  & 43.7  & 68.5  & 63.2  & 51.8  & 76.8  & 57.6 \\
		CDAN~\cite{cdan}      & 49.0  & 69.3  & 74.5  & 54.4  & 66.0  & 68.4  & 55.6  & 48.3  & 75.9  & 68.4  & 55.4  & 80.5  & 63.8 \\
		CDAN+E~\cite{cdan}    & 50.7  & 70.6  & 76.0  & 57.6  & 70.0  & 70.0  & 57.4  & 50.9  & 77.3  & 70.9  & 56.7  & 81.6  & 65.8 \\
		DWT-MEC~\cite{dwt-mec}   & 50.3  & 72.1  & 77.0  & 59.6  & 69.3  & 70.2  & 58.3  & 48.1  & 77.3  & 69.3  & 53.6  & 82.0  & 65.6 \\
		SAFN~\cite{safn}     & 52.0  & 73.3  & 76.3  & 64.2  & 69.9  & 71.9  & 63.7  & 51.4  & 77.1  & 70.9  & 57.1  & 81.5  & 67.3 \\
		\midrule
		DANN~\cite{dann} & 48.1 & 62.3 & 74.9 & 54.2 & 64.1 & 65.2 & 53.6 & 47.7 & 74.6 & 66.9 & 54.6 & 79.0 & 62.1 \\
		DANN-[f, p]~\cite{cdan}   & 49.4  & 63.4  & 74.0 & 56.1 & 63.9 & 65.7 & 54.7 & 47.4 & 74.7 & 67.6 & 54.8 & 79.6  & 62.6 \\
		MADA~\cite{madacls}  & 47.2 & 69.7 & 75.1 & 54.3 & 68.5 & 69.8 & 53.0 & 42.6 & 75.0 & 70.1 &  51.2 & 80.8 & 63.1 \\
		CDAN~\cite{cdan} & 51.5 & 69.6 & 75.2 & 58.1 & 69.1 & 68.9 & 58.8 & 49.7 & 77.7 & 70.3 & 55.5 & 81.4 & 65.5 \\
		IDDA~\cite{idda}   & 49.9  & 66.2  & 74.2 & 57.6 & 66.5 & 66.9 & 55.5 & 49.6 & 74.9  & 67.0 & 53.2 & 78.6  & 63.4 \\
		RCA~\cite{rca} & 47.8 & 59.7 & 72.7 & 56.6 & 68.1 & 68.9 & 55.2 & 47.8 & 75.9 & 67.9 & 52.7 & 79.7 & 62.7 \\
		DANN-CA~\cite{dannca} & 45.7 & 68.4 & 75.4 & 58.4 & 67.1 & 68.8 & 55.4 & 43.7 & 74.6 & 66.5 & 49.2 & 78.3 & 62.6 \\
		CDAN+E~\cite{cdan} & 55.3 & 72.0 & 77.8 & 63.5 & 71.9 & 72.7 & 62.5 & 54.5 & 80.2 & 72.3 & 57.9 & 83.7 & 68.7 \\
		\midrule
		source-model-only & 46.9 & 67.5 & 75.3 & 56.2 & 63.8 & 66.6 & 55.1 & 42.3 & 74.3 & 66.9 & 49.3 & 78.2 & 61.9 \\
		& $\pm$0.2 & $\pm$0.7 & $\pm$0.1 & $\pm$0.9 & $\pm$1.1 & $\pm$0.2 & $\pm$0.3 & $\pm$1.1 & $\pm$0.8 & $\pm$0.5 & $\pm$0.2 & $\pm$0.1& $\pm$0.2\\
		\cmidrule{2-14}
		NOUN & 53.2 & 69.2 & 76.0 & 60.6 & 68.5 & 69.9 & 62.6 & 53.3 & 78.6 & 71.2 & 56.0 & 80.9 & 66.7 \\
		& $\pm$0.1 & $\pm$0.2 & $\pm$0.4 & $\pm$0.5 & $\pm$0.5 & $\pm$0.2 & $\pm$0.2 & $\pm$0.0 & $\pm$0.3 & $\pm$0.4 & $\pm$0.4 & $\pm$0.5& $\pm$0.0\\
		\cmidrule{2-14}
		PRONOUN  & \textbf{57.6} & \textbf{75.0} & \textbf{78.4} & \textbf{64.9} & \textbf{74.0} & \textbf{74.8} & \textbf{66.6} & \textbf{58.2} & \textbf{80.4} & \textbf{74.3} & \textbf{60.4} & \textbf{84.3} & \textbf{70.7}  \\
		& $\pm$0.7 & $\pm$0.5 & $\pm$0.1 & $\pm$0.6 & $\pm$0.8 & $\pm$0.0 & $\pm$0.3 & $\pm$0.4 & $\pm$0.0 & $\pm$0.3 & $\pm$1.1 & $\pm$0.1& $\pm$0.2\\
		\bottomrule 
		\end{tabular}
		$}
\end{table*}

\setlength{\tabcolsep}{6.0pt}
\begin{table*}[h]
	    \centering
	    \small
		\caption{Accuracies (\%) on \emph{VisDA2017} via ResNet-101.}
		\label{tab:visda}
		\centering
		\resizebox{0.95\textwidth}{!}{$
			\begin{tabular}{lcccccccccccca}
			\toprule
			Method & plane & bcycl & bus & car & horse & knife & mcycl & person & plant & sktbrd & train & truck & Per-class \\
			\midrule
			ResNet-101~\cite{he2016deep}  & 55.1 & 53.3 & 61.9 & 59.1 & 80.6 & 17.9 & 79.7 & 31.2  & 81.0 & 26.5  & 73.5 & 8.5  & 52.4   \\
			DANN~\cite{dann}   & 81.9 & 77.7 & 82.8 & 44.3 & 81.2 & 29.5 & 65.1 & 28.6  & 51.9 & 54.6  & 82.8 & 7.8  & 57.4   \\
			ADR~\cite{adr} & \textbf{94.2} & 48.5 & 84.0 & 72.9 & 90.1 & 74.2 & \textbf{92.6} & 72.5 & 80.8 & 61.8 & 82.2 & 28.8 & 73.5 \\
			CDAN~\cite{cdan}   & 85.2 & 66.9 & 83.0 & 50.8 & 84.2 & 74.9 & 88.1 & 74.5  & 83.4 & 76.0  & 81.9 & 38.0 & 73.7   \\
			CDAN+BSP \cite{bsp} & 92.4 & 61.0 & 81.0 & 57.5 & 89.0 & 80.6 & 90.1 & 77.0 & 84.2 & 77.9 & 82.1 & 38.4 & 75.9 \\
			SWD~\cite{swd} & 90.8 & 82.5 & 81.7 & 70.5 & 91.7 & 69.5 & 86.3 & 77.5 & 87.4 & 63.6 & 85.6 & 29.2 & 76.4 \\
			SAFN~\cite{safn} & 93.6 & 61.3 & \textbf{84.1} & 70.6 & \textbf{94.1} & 79.0 & 91.8 & 79.6 & 89.9 & 55.6 & 89.0 & 24.4 & 76.1 \\
			\midrule
			DANN~\cite{dann} & 90.0 & 58.9 & 76.9 & 56.1 & 80.3 & 60.9 & 89.1 & 72.5 & 84.3 & 73.8 & \textbf{89.3} & 35.8 & 72.3 \\
		    DANN-[f, p]~\cite{cdan} & 90.7 & 55.6 & 76.3 & 62.9 & 82.6 & 64.8 & 83.9 & 73.0 & 84.6 & 69.6 & 86.7 & 35.8 & 72.2 \\
		    MADA~\cite{madacls} & 84.8 & 53.3 & 78.1 & 66.6 & 87.4 & 80.3 & 85.9 & 72.4 & 87.2 & 69.7 & 83.3 & 25.8 & 72.9 \\
		    CDAN~\cite{cdan} & 91.9 & 62.4 & 74.5 & 72.3 & 89.0 & 83.0 & 88.6 & \textbf{80.9} & 85.3 & 72.7 & 86.6 & 32.6 & 76.3 \\
		    IDDA~\cite{idda} & 76.4 & 75.8 & 74.2 & 58.9 & 90.4 & \textbf{94.9} & 80.3 & 74.5 & 90.1 & 69.7 & 82.5 & 47.3 & 76.2 \\
		    RCA~\cite{rca} & 79.6 & 55.2 & 69.4 & \textbf{73.0} & 87.3 & 59.3 & 86.0 & 61.2 & 86.7 & 72.4 & 84.2 & 28.3 & 70.2 \\
		    DANN-CA~\cite{dannca} & 88.9 & 37.4 & 78.4 & 71.3 & 91.9 & 63.1 & 90.9 & 35.3 & 84.5 & 23.9 & 88.2 & 20.1 & 64.5 \\
		    CDAN+E~\cite{cdan} & 90.1 & 63.4 & 83.8 & 71.9 & 88.5 & 94.2 & 91.2 & \textbf{80.9} & 85.4 & 77.4 & 85.0 & 43.4 & 79.1 \\
			\midrule
			source-model-only & 77.5 & 14.5 & 46.7 & 69.3 & 67.3 & 4.7 & 78.3 & 10.9 & 66.1 & 22.1 & 81.5 & 5.8 & 45.4 \\
			& $\pm$0.2 & $\pm$0.7 & $\pm$0.1 & $\pm$0.9 & $\pm$1.1 & $\pm$0.2 & $\pm$0.3 & $\pm$1.1 & $\pm$0.8 & $\pm$0.5 & $\pm$0.2 & $\pm$0.1& \\
		    \cmidrule{2-14}
			NOUN & 92.5 & 64.3 & 78.3 & 62.9 & 89.5 & 88.2 & 90.8 & 79.1 & 87.8 & 81.2 & 83.8 & \textbf{48.1} & 78.9 \\
			& $\pm$1.3 & $\pm$6.6 & $\pm$5.2 & $\pm$7.3 & $\pm$1.2 & $\pm$2.8 & $\pm$1.8 & $\pm$0.5 & $\pm$2.6 & $\pm$4.5 & $\pm$1.4 & $\pm$11.5& \\
			\cmidrule{2-14}
			PRONOUN & 92.7 & \textbf{69.3} & 82.7 & 70.5 & 92.8 & 91.4 & 91.3 & 79.9 & \textbf{90.2} & \textbf{82.2} & 87.8 & 47.9 & \textbf{81.6}\\
			& $\pm$2.1 & $\pm$7.6 & $\pm$1.0 & $\pm$0.8 & $\pm$0.7 & $\pm$2.4 & $\pm$1.2 & $\pm$2.1 & $\pm$1.3 & $\pm$1.6 & $\pm$1.6 & $\pm$3.9 & \\
			\bottomrule
			\end{tabular}
			$}
\end{table*}

In this section, we conduct extensive experiments to evaluate NOUN and PRONOUN. We start with experimental settings, including benchmarks, implementation details, and compared methods. Then we compare our solutions with related works on various UDA benchmarks. At last, we provide detailed analyses on our solutions, including ablation study, compatibility with self-training, stability of training, the sensitivity of hyper-parameters, and qualitative visualizations in the feature space or output space.

\subsection{Experimental Setup}
\label{sec:exp_setup}
We conduct experiments to verify the effectiveness of NOUN and PRONOUN on cross-domain object recognition benchmarks including \emph{Office-Home}~\cite{venkateswara2017Deep}, \emph{VisDA2017}~\cite{peng2017visda}, \emph{Office31}~\cite{saenko2010adapting} and \emph{ImageCLEF-DA}\footnote{\url{https://www.imageclef.org/2014/adaptation}}. Besides, we validate the generalization of our methods on synthetic-to-real semantic segmentation benchmarks including \emph{GTA5}~\cite{richter2016playing}$\to$\emph{Cityscapes}~\cite{cordts2016cityscapes} and \emph{Synthia}~\cite{ros2016synthia}$\to$ \emph{Cityscapes}.

\subsubsection{Datasets} 
\emph{Office-Home} is a new challenging dataset that consists of 65 different object categories found typically in 4 different Office and Home settings, i.e., Artistic (\textbf{Ar}) images, Clip Art (\textbf{Ca}), Product images (\textbf{Pr}) and Real-World (\textbf{Re}) images.
\emph{VisDA2017} is a challenging large-scale benchmark aimed at the synthetic-to-real object recognition task across 12 categories. The source domain consists of 152k synthetic images generated by rendering 3D models. The target domain contains 55k realistic images gathered from \emph{Microsoft COCO}~\cite{lin2014microsoft}.
\emph{Office31} is a popular dataset that includes 31 object categories taken from 3 domains, i.e., Amazon (\textbf{A}), DSLR (\textbf{D}) and Webcam (\textbf{W}).
\emph{ImageCLEF-DA} is a dataset built for the `ImageCLEF2014:domain-adaptation' competition. We follow \cite{long2015learning} to select 3 subsets, i.e., \textbf{C}, \textbf{I} and \textbf{P}, which share 12 common object classes.

\emph{Cityscapes} is a realistic dataset of pixel-level annotated urban street scenes. We use its original training split and validation split as training target data and testing target data, respectively. 
\emph{GTA5} consists of 24,966 densely labeled synthetic road scenes annotated with the same 19 classes as \emph{Cityscapes}. 
For \emph{Synthia}, we take the SYNTHIA-RAND-CITYSCAPES set as the source domain, which contains 9,400 synthetic images compatible with 16 annotated classes of \emph{Cityscapes}~\cite{zhang2017curriculum}. 

\subsubsection{Implementation details} 
For object recognition, we follow the standard protocol~\cite{dann}, i.e., using all labeled source instances and all unlabeled target instances for UDA, and report the average accuracy based on results of three random trials for fair comparisons.
Following \cite{cdan}, we experiment with ResNet-50 model pre-trained on ImageNet for all datasets excluding \emph{VisDA2017} where we use ResNet-101 model to compare with other works.
We follow \cite{cdan} to choose the network parameters. The whole model is trained through back-propagation, and $\lambda_{adv}$ increases from 0 to 1 with the same strategy as \cite{dann}.
Regarding the domain discriminator, we design a simple classifier with only one ReLU \cite{nair2010rectified} layer and two linear layers (256$\to$1024$\to$1). Empirically, we fix the batch size to 36 with the initial learning rate as 1e-4.

For semantic segmentation, we adopt DeepLab-V2~\cite{chen2017deeplab} based on ResNet-101~\cite{he2016deep} as done in \cite{adaptseg, advent, clan, adaptpatch}.
Following DCGAN~\cite{radford2015unsupervised}, the discriminator network 
consists of one LeakyReLU layer~\cite{maas2013rectifier} with the slope of 0.2 and two $4 \times 4$ convolutional layers with stride 2 (512$\to$512$\to$1).
In training, we use SGD~\cite{bottou2010large} to optimize the network with
momentum (0.9), weight decay (5e-4) and initial learning rate (2.5e-4).
We use the same learning rate policy as in~\cite{chen2017deeplab}. 
Discriminators are optimized by Adam~\cite{kingma2014adam} with momentum ($\beta_1=0.9$, $\beta_2=0.99$), initial learning rate (1e-4) along with the same decreasing strategy as above. $\lambda_{adv}$ is set to 1e-3 following~\cite{adaptseg}.

All experiments are implemented via PyTorch on a single 12GB GPU. The total iteration number is set as 10k for object recognition and 100k for semantic segmentation. The momentum value $\lambda_{\mathrm{ema}}$ is set to 0.5 for all tasks without parameter selection. The norm control factor $k$ is determined by reverse validation \cite{dann} and is set to 3 in PRONOUN for all tasks. Similar to the entropy-based criterion in~\cite{morerio2017minimal}, the adapted model producing target predictions with the minimum mean entropy is selected for testing in the target domain.
Data augmentation skills like random scale or ten-crop ensemble evaluation are not adopted.

\subsubsection{Baseline methods}
The general comparisons between our NOUN as well as PRONOUN and other conditional domain adversarial training methods are shown in Table \ref{tab:cmp_adv}.

For cross-domain object recognition tasks, we compare our methods with DANN \cite{dann}, DANN-[f, p] \cite{cdan}, MADA \cite{madacls}, CDAN \cite{cdan}, IDDA \cite{idda}, RCA \cite{rca}, DANN-CA \cite{dannca} and CDAN+E \cite{cdan}. Note that we re-implement these most related methods using the same training protocol including batch size and training iterations. As for other hyper-parameters including learning rate and optimizer setting, we strictly follow the released code by authors or the implementation details in original papers. For fair comparisons, we report both results in original papers if any, and the accuracy of our re-implementation. Besides, we also compare with the state-of-the-art works including SAFN \cite{safn}, SWD \cite{swd}, ADR \cite{adr}, DWC-MEC \cite{dwt-mec}, BSP \cite{bsp}, CAT \cite{cat} and iCAN \cite{ican}. Our reported `source-model-only' means training a model on source data with only the supervised loss in Eq.~(\ref{eq:ly}) while using the same training protocol and architecture. In this way, we can make fair comparisons among different methods.

For synthetic-to-real semantic segmentation tasks, we compare our methods with domain adversarial training methods including AdaptSeg \cite{adaptseg}, SIBAN \cite{siban}, AdvEnt \cite{advent}, CLAN \cite{clan}, AdaptPatch \cite{adaptpatch} and another generic UDA method SWD \cite{swd}. `NonAdapt' shares the similar meaning with `source-model-only' in object recognition benchmarks. 

By default, NOUN introduces no hyper-parameter by setting the $k$ in Eq.~(\ref{eq:sdan_loss}) as 1.
For results reported in Table~\ref{tab:home}--Table~\ref{tab:synthia}, we report results from original papers in the top rows, the re-implementation results of related methods using the same protocol in the middle rows, and results of our methods in the bottom rows. 

\subsection{Results on object recognition benchmarks.} 
\label{exp_cls}
For cross-domain object recognition, we evaluate our methods on benchmarks including \emph{Office-Home}, \emph{VisDA2017}, \emph{Office31}, and \emph{ImageCLEF-DA}.

\subsubsection{\emph{Office-Home}}
The results of the challenging 65-way classification benchmark \emph{Office-Home} are reported in Table~\ref{tab:home}. All of the compared conditional domain adversarial training methods beat DANN in terms of the average accuracy, which verifies the effectiveness of conditional feature alignment for domains with multi-modal structures. Among these re-implemented methods, CDAN~\cite{cdan} models perform the best and achieve markedly higher average accuracy than results in original papers, i.e., 63.8\% $\to$ 65.5\% for CDAN and 65.8\% $\to$ 68.7\% for CDAN+E. The naive concatenation conditioning strategy DANN-[f, p] brings minor improvement of the average accuracy (0.6\%) over DANN. 
While NOUN brings a 4.1\% absolute increase of the average accuracy over DANN[f, p] and even outperforms the more informative multi-linear conditioning method CDAN. Suffering from inaccurate pseudo-labels, NOUN still falls behind CDAN+E. While conditioning on the semantic structure-aware predictions, PRONOUN outperforms CDAN+E for all of the 12 transfer tasks and improves the average accuracy over NOUN by 4\%.

\subsubsection{\emph{VisDA2017}}
The results of the large-scale benchmark \emph{VisDA2017} are reported in Table~\ref{tab:visda}. DANN-[f, p] achieves almost the same per-class accuracy as DANN, while CDAN models still outperform other compared domain adversarial training methods. NOUN significantly improves the per-class accuracy by 6.7\% over DANN-[f, p], comparable with CDAN+E. Both NOUN and CDANE outperform all other comparison methods. PRONOUN notably improves NOUN and achieves the best per-class accuracy among all domain adversarial training methods.

\setlength{\tabcolsep}{1.0pt}
\begin{table}[h]
        \centering
        \small
		\caption{Accuracies (\%) on \emph{Office31} via ResNet-50.}
		\label{tab:office}
		\centering
		\resizebox{0.48\textwidth}{!}{$
			\begin{tabular}{lcccccca}
			\toprule
			Method (Source$\to$Target) & A$\to$D & A$\to$W & D$\to$A & D$\to$W & W$\to$A & W$\to$D  & Avg. \\
			\midrule
			ResNet-50~\cite{he2016deep} & 68.9 & 68.4 & 62.5 & 96.7 & 60.7 & 99.3 & 76.1 \\
			DANN~\cite{dann} & 79.7 & 82.0 & 68.2 & 96.9 & 67.4 & 99.1 & 82.2\\
			CDAN~\cite{cdan}  & 89.8 & 93.1 & 70.1 & 98.2 & 68.0 & \textbf{100.} & 86.6\\
			CDAN+E~\cite{cdan}  & 92.9 & 94.1 & 71.0 & 98.6 & 69.3 & \textbf{100.} & 87.7 \\
			DANN-CA~\cite{dannca}  & 89.9 & 91.4 & 69.6 & 98.2 & 68.8 & 99.5 & 86.2\\
			MADA~\cite{madacls}  & 87.8 & 90.0 & 70.3 & 97.4 & 66.4 & 99.6 & 85.2\\
			iCAN~\cite{ican} & 90.1 & 92.5 & 72.1 & \textbf{98.8} & 69.9 & \textbf{100.} & 87.2\\
			CAT~\cite{cat} & 90.8 & \textbf{94.4} & 72.2 & 98.0 & 70.2 & \textbf{100.}  & 87.6\\
			SAFN~\cite{safn} & 87.7 & 88.8 & 69.8 & 98.4 & 69.7 & 99.8 & 85.7\\
			SAFN+Ent~\cite{safn} & 90.7 & 90.1 & 73.0 & 98.6 & 70.2 & 99.8 & 87.1\\
			\midrule
			DANN~\cite{dann}  & 79.1 & 85.3 & 66.0 & 96.2 & 69.4 & 99.8 & 82.6\\
			DANN-[f, p]~\cite{cdan}  & 80.3 & 86.0 & 67.5 & 96.3 & 69.2 & 99.6 & 83.1\\
			MADA~\cite{madacls}  & 91.0 & 91.5 & 68.2 & 98.0 & 69.3 & 99.9 & 86.3\\
			CDAN~\cite{cdan}  & 89.8 & 88.6 & 69.6 & 98.5 & 70.4 & \textbf{100.} & 86.2\\
			IDDA~\cite{idda}  & 85.7 & 88.3 & 66.0 & 96.5 & 67.7 & 99.8 & 84.0\\
			RCA~\cite{rca}  & 83.3 & 86.7 & 67.3 & 97.9 & 66.8 & \textbf{100.} & 83.7\\
			DANN-CA~\cite{dannca}  & 84.4 & 84.7 & 66.1 & 97.7 & 65.9 & 99.8 & 83.1\\
			CDAN+E~\cite{cdan}  & 90.2 & 91.0 & 73.8 & \textbf{98.8} & 72.3 & 99.7 & 87.6\\
			\midrule
			source-model-only & 80.9 & 79.5 & 64.3 & 97.9 & 64.2 & 99.9 & 81.1\\
			& $\pm$1.6 & $\pm$0.9 & $\pm$0.7 & $\pm$0.4 & $\pm$0.5 & $\pm$0.1 & $\pm$0.4\\
		    \cmidrule{2-8}
			NOUN  & 88.5 & 90.9 & 73.4 & 98.5 & 72.7 & \textbf{100.} & 87.3\\
			& $\pm$2.2 & $\pm$1.5 & $\pm$0.1 & $\pm$0.7 & $\pm$1.0 & $\pm$0.0 & $\pm$0.5\\
		    \cmidrule{2-8}
			PRONOUN & \textbf{93.5} & 93.4 & \textbf{74.6} & 98.2 & \textbf{73.4} & \textbf{100.} & \textbf{88.8}\\
			& $\pm$2.2 & $\pm$0.8 & $\pm$0.5 & $\pm$0.8 & $\pm$0.8 & $\pm$0.0 & $\pm$0.4\\
			\bottomrule
			\end{tabular}
			$}
\end{table}

\subsubsection{\emph{Office31}}
The results of the standard benchmark \emph{Office31} are reported in Table~\ref{tab:office}. DANN-[f, p] improves marginally over DANN. Though CDAN+E outperforms comparison methods markedly, CDAN proves slightly inferior to MADA. The observation is reasonable because tasks in \emph{Office31} such as A$\to$D and A$\to$W are easy and can provide more accurate pseudo-labels, which prioritizes methods involving explicit class-wise cross-domain feature alignment like MADA. Comparable with CDAN+E, NOUN improves the average accuracy by 4.2\% over DANN-[f, p]. PRONOUN brings handsome improvement over NOUN on four main transfer tasks, including A$\to$D, A$\to$W, D$\to$A, and W$\to$A, and outperforms CDAN+E.

\setlength{\tabcolsep}{1.0pt}
\begin{table}[h]
        \centering
        \small
		\caption{Accuracies (\%) on \emph{ImageCLEF-DA} via ResNet-50.}
		\label{tab:clef}
		\centering
		\resizebox{0.48\textwidth}{!}{$
			\begin{tabular}{lcccccca}
			\toprule
			Method (Source$\to$Target) & C$\to$I & C$\to$P & I$\to$C & I$\to$P & P$\to$C & P$\to$I & Avg. \\
			\midrule
			ResNet-50~\cite{he2016deep}    & 78.0 & 65.5 & 91.5 & 74.8 & 91.2 & 83.9 & 80.7 \\
			DANN~\cite{dann}       & 87.0 & 74.3 & 96.2 & 75.0 & 91.5 & 86.0 & 85.0 \\
			CDAN~\cite{cdan}      & 90.5 & 74.5 & 97.0 & 76.7 & 93.5 & 90.6 & 87.1 \\
			CDAN+E~\cite{cdan}       & 91.3 & 74.2 & \textbf{97.7} & 77.7 & 94.3 & 90.7 & 87.7 \\
			MADA~\cite{madacls}  & 88.8 & 75.2 & 96.0 & 75.0 & 92.2 & 87.9 & 85.8\\
			iCAN~\cite{ican} & 89.9 & \textbf{78.5} & 94.7 & \textbf{79.5} & 92.0 & 89.7 & 87.4 \\
			CAT~\cite{cat} & 91.3 & 75.3 & 95.5 & 77.2 & 93.6 & 91.0 & 87.3\\
			SAFN~\cite{safn}     & 91.1 & 77.0 & 96.2 & 78.0 & 94.7 & 91.7 & 88.1 \\
			SAFN+Ent~\cite{safn}     & 91.7 & 77.6 & 96.3 & 79.3 & \textbf{95.3} & \textbf{93.3} & 88.9\\
			\midrule
			DANN~\cite{dann}  & 91.6 & 76.5 & 95.3 & 76.8 & 92.3 & 90.1 & 87.1\\
			DANN-[f, p]~\cite{cdan}  & 91.3 & 77.3 & 95.0 & 76.6 & 92.9 & 89.9 & 87.2\\
			MADA~\cite{madacls}  & 91.6 & 77.1 & 97.0 & 77.1 & 94.6 & 90.7 & 88.0\\
			CDAN~\cite{cdan}  & 91.2 & 76.5 & 95.9 & 77.0 & 93.6 & 91.0 & 87.5\\
			IDDA~\cite{idda}  & 90.7 & 75.8 & 95.6 & 74.1 & 94.2 & 88.6 & 86.5\\
			RCA~\cite{rca}  & 90.1 & 75.5 & 95.5 & 76.1 & 89.4 & 88.9 & 85.9\\
			DANN-CA~\cite{dannca}  & 90.2 & 75.3 & 96.2 & 76.4 & 94.1 & 90.7 & 87.1\\
			CDAN+E~\cite{cdan}  & 91.4 & 77.1 & 95.9 & 77.7 & 94.3 & 91.9 & 88.1\\
			\midrule
			source-model-only & 85.3 & 72.6 & 93.2 & 75.7 & 91.8 & 88.2 & 84.5\\
			& $\pm$0.3 & $\pm$0.8 & $\pm$0.3 & $\pm$0.4 & $\pm$0.9 & $\pm$0.6 & $\pm$0.1\\
		    \cmidrule{2-8}
			NOUN & \textbf{92.3} & 78.1 & 96.8 & 77.6 & 94.6 & 91.8 & 88.5\\
			& $\pm$1.1 & $\pm$0.5 & $\pm$0.2 & $\pm$0.3 & $\pm$0.2 & $\pm$1.0 & $\pm$0.3\\
		    \cmidrule{2-8}
			PRONOUN & \textbf{92.3} & 78.3 & 97.4 & 78.0 & 95.2 & 92.6 & \textbf{89.0}\\
			& $\pm$0.8 & $\pm$0.9 & $\pm$0.3 & $\pm$0.4 & $\pm$0.8 & $\pm$0.9 & $\pm$0.4\\
			\bottomrule
			\end{tabular}
			$}
\end{table}

\setlength{\tabcolsep}{2.0pt}
\begin{table*}[t]
   \caption{Comparison results of \emph{GTA5}~\cite{richter2016playing}$\to$\emph{Cityscapes}~\cite{cordts2016cityscapes} semantic segmentation using ResNet-101 as the backbone.} 
   \label{tab:gta}
   \centering
   \resizebox{1.0\textwidth}{!}{$
        \begin{tabular}{lccccccccccccccccccca}
        \toprule
        Methods        & \rotatebox{90}{road}  & \rotatebox{90}{sdwk}  & \rotatebox{90}{bldng}  & \rotatebox{90}{wall} & \rotatebox{90}{fence} & \rotatebox{90}{pole}  & \rotatebox{90}{light} & \rotatebox{90}{sign} & \rotatebox{90}{veg.}  & \rotatebox{90}{ter.} & \rotatebox{90}{sky}   & \rotatebox{90}{per.}  & \rotatebox{90}{rider} & \rotatebox{90}{car}   & \rotatebox{90}{truck} & \rotatebox{90}{bus}  & \rotatebox{90}{train} & \rotatebox{90}{mbike} & \rotatebox{90}{bike}  & \rotatebox{90}{mIoU} \\
        \midrule
        NonAdapt~\cite{adaptseg}  & 75.8  & 16.8  & 77.2  & 12.5 & 21.0    & 25.5  & 30.1  & 20.1 & 81.3  & 24.6 & 70.3  & 53.8  & 26.4  & 49.9  & 17.2  & 25.9 & 6.5   & 25.3  & 36.0    & 36.6 \\
        AdaptSeg~\cite{adaptseg}       & 86.5  & 25.9  & 79.8  & 22.1 & 20.0    & 23.6  & 33.1  & 21.8 & 81.8  & 25.9 & 75.9  & 57.3  & 26.2  & 76.3  & 29.8  & 32.1 & \textbf{7.2}   & 29.5  & 32.5  & 41.4  \\
        AdvEnt~\cite{advent}         & 89.9  & 36.5  & 81.6  & \textbf{29.2} & \textbf{25.2}  & 28.5  & 32.3  & 22.4 & 83.9  & 34.0   & 77.1  & 57.4  & 27.9  & 83.7  & 29.4  & 39.1 & 1.5   & 28.4  & 23.3  & 43.8  \\
        SIBAN~\cite{siban}           & 88.5    & 35.4  & 79.5  & 26.3 & 24.3  & 28.5  & 32.5  & 18.3 & 81.2  & \textbf{40.0} & 76.5  & 58.1  & 25.8    & 82.6  & 30.3  & 34.4 & 3.4  & 21.6  & 21.5  & 42.6 \\
        CLAN~\cite{clan}           & 87.0    & 27.1  & 79.6  & 27.3 & 23.3  & 28.3  & 35.5  & 24.2 & 83.6  & 27.4 & 74.2  & 58.6  & 28.0    & 76.2  & 33.1  & 36.7 & 6.7   & \textbf{31.9}  & 31.4  & 43.2 \\
        AdaptPatch~\cite{adaptpatch}     & 89.2  & 38.4  & 80.4  & 24.4 & 21.0    & 27.7  & 32.9  & 16.1 & 83.1  & 34.1 & 77.8  & 57.4  & 27.6  & 78.6  & 31.2  & 40.2 & 4.7   & 27.6  & 27.6  & 43.2 \\
        SWD (PSP)~\cite{swd} & \textbf{92.0} & \textbf{46.4} & \textbf{82.4} & 24.8 & 24.0 & \textbf{35.1} & 33.4 & \textbf{34.2} & 83.6 & 30.4 & \textbf{80.9} & 56.9 & 21.9 & 82.0 & 24.4 & 28.7 & 6.1 & 25.0 & 33.6 & 44.5 \\
        \midrule
        NonAdapt  & 83.7 & 10.2 & 73.9 & 20.7 & 15.2 & 23.4 & 32.9 & 22.4 & 76.7 & 29.6 & 66.7 & 55.0 & 28.5 & 69.4 & 29.8 & 40.1 & 0.9 & 16.2 & 33.6 & 38.4 \\
        NOUN   & 88.5 & 20.9 & 81.6 & 26.7 & 23.6 & 31.1 & \textbf{37.5} & 29.2 & 84.0 & 37.2 & 76.4 & \textbf{59.8} & 26.6 & \textbf{85.4} & \textbf{36.8} & \textbf{43.9} & 3.4 & 26.5 & \textbf{42.2} & \textbf{45.3} \\
        PRONOUN   & 88.7 & 25.8 & 81.7 & 25.8 & 23.1 & 30.5 & 36.3 & 25.7 & \textbf{84.1} & 34.8 & 76.6 & 59.4 & \textbf{28.9} & 84.1 & 34.6 & 42.7 & 6.4 & 25.3 & 34.1 & 44.7 \\
        \bottomrule
        \end{tabular}
        $}
\end{table*}

\setlength{\tabcolsep}{3pt}
\begin{table*}[t]
   \caption{Comparison results of \emph{Synthia}~\cite{ros2016synthia}$\to$\emph{Cityscapes} semantic segmentation using ResNet-101 as the backbone. mIoU{*} denotes the mean IoU of 16 classes, including the classes with {*}.} 
   \label{tab:synthia}
   \centering
   \resizebox{1.0\textwidth}{!}{$
        \begin{tabular}{lccccccccccccccccaa}
        \toprule
         Methods        & \rotatebox{90}{road}  & \rotatebox{90}{sdwk}  & \rotatebox{90}{bldng}  & \rotatebox{90}{wall*} & \rotatebox{90}{fence*} & \rotatebox{90}{pole*}  & \rotatebox{90}{light} & \rotatebox{90}{sign} & \rotatebox{90}{veg.}  & \rotatebox{90}{sky}   & \rotatebox{90}{per.}  & \rotatebox{90}{rider} & \rotatebox{90}{car} & \rotatebox{90}{bus}  & \rotatebox{90}{mbike} & \rotatebox{90}{bike}  & \rotatebox{90}{mIoU} & \rotatebox{90}{mIoU*} \\
        \midrule
        NonAdapt~\cite{adaptseg}   & 55.6  & 23.8  & 74.6  & -    & -     & -     & 6.1   & 12.1 & 74.8  & 79.0    & 55.3  & 19.1  & 39.6  &  23.3  & 13.7  & 25.0    & 38.6 & -  \\
        AdaptSeg~\cite{adaptseg}       & 79.2  & 37.2  & 78.8  & -    & -     & -     & 9.9  & 10.5 & 78.2  & 80.5  & 53.5  & 19.6  & 67.0  & 29.5  & 21.6  & 31.3  & 45.9 & - \\
        AdvEnt~\cite{advent}         & 87.0    & 44.1  & 79.7  & \textbf{9.6} & \textbf{0.6} & 24.3  & 4.8   & 7.2  & 80.1   & 83.6  & 56.4  & \textbf{23.7}  & 72.7 & 32.6  & 12.8  & 33.7  & 47.6 & 40.8 \\
        SIBAN~\cite{siban}          & 82.5  & 24.0    & 79.4  & -    & -     & -     & 16.5  & 12.7 & 79.2  & 82.8  & \textbf{58.3}  & 18.0  & 79.3  & 25.3 & 17.6  & 25.9  & 46.3 & - \\
        CLAN~\cite{clan}           & 81.3  & 37.0    & 80.1  & -    & -     & -     & 16.1  & 13.7 & 78.2  & 81.5  & 53.4  & 21.2  & 73.0  & \textbf{32.9} & \textbf{22.6}  & 30.7  & 47.8 & -  \\
        AdaptPatch~\cite{adaptpatch}     & 82.4  & 38.0  & 78.6  & 8.7  &  \textbf{0.6}    &  26.0    & 3.9   & 11.1 & 75.5  & \textbf{84.6}  & 53.5  & 21.6  & 71.4  & 32.6 & 19.3  & 31.7  & 46.5 & 40.0 \\
        SWD (PSP)~\cite{swd} & 82.4 & 33.2 & \textbf{82.5} & - & - & - & \textbf{22.6} & \textbf{19.7} & \textbf{83.7} & 78.8 & 44.0 & 17.9 & 75.4 & 30.2 & 14.4 & \textbf{39.9} & 48.1 & - \\
        \midrule
        NonAdapt  & 38.2 & 17.5 & 73.8 & 4.6 & 0.0 & 25.0 & 5.8 & 10.9 & 73.3 & 80.7 & 55.7 & 17.3 & 40.3 & 16.4 & 12.3 & 21.7 & 35.7 & 30.9 \\
        NOUN  & \textbf{89.3} & \textbf{45.2} & 79.4 & 4.0 & 0.4 & \textbf{27.0} & 8.0 & 13.2 & 78.9 & 82.3 & 
        57.0 & 20.4 & 80.9 & 30.8 & 15.5 & 38.0 & \textbf{49.2} & \textbf{41.9}  \\
        PRONOUN  & 89.2 & 44.4 & 78.8 & 4.5 & 0.2 & \textbf{27.0} & 6.5  & 11.0 & 78.0 & 81.3 & 57.0 & 22.0 & \textbf{81.5}  & 30.3 & 18.4 & 39.4 & 49.1 & 41.8 \\
        \bottomrule
        \end{tabular}
        $}
\end{table*}

\subsubsection{\emph{ImageCLEF-DA}}
We report results of the easy benchmark \emph{ImageCLEF-DA} with equal domain size and balanced classes in Table~\ref{tab:clef}. Domains of \emph{ImageCLEF-DA} are more visually similar than those of the above benchmarks. Thus the average accuracy of the source-trained model is higher, and various comparison methods exhibit generally comparable performance. Specifically, DANN-[f, p] shows few improvements over DANN. Beyond the observation on \emph{Office31}, MADA is shown to be comparable with CDAN+E and outperforms CDAN. While NOUN slightly beats CDAN+E and PRONOUN further improves the average accuracy over NOUN.

Across the above results on four different cross-domain object recognition benchmarks, there are two consistent observations about our methods. The first one is that NOUN always brings a significant improvement over DANN-[f, p] and NOUN beats all of the other domain adversarial training methods compared in Table~\ref{tab:cmp_adv}. This observation strongly demonstrates the effectiveness of NOUN. It also indicates that the failure of the naive concatenation conditioning method mainly lies in the weak conditioning strength, since NOUN only normalizes the norm of concatenated output in DANN-[f, p] with the norm of the feature. The second consistent observation is that PRONOUN always further improves NOUN and outperforms all of the compared baselines, which verifies the superiority of our prototype-based conditioning strategy. 

\subsection{Results on semantic segmentation benchmarks.}
\label{exp_seg}
NOUN can be seamlessly extended to tasks with dense prediction. As for PRONOUN, we obtain prototypes by simply averaging features of corresponding pixels. We evaluate both methods on synthetic-to-real semantic segmentation benchmarks. Following the VisDA challenge~\cite{peng2017visda}, we adopt the mean intersection-over-union (mIoU) as the evaluation metric. The results of \emph{GTA5}$\to$\emph{Cityscapes} are reported in Table~\ref{tab:gta} and the results of \emph{Synthia}$\to$\emph{Cityscapes} are reported in Table~\ref{tab:synthia}. 
For both tasks, NOUN outperforms all of the compared domain adversarial training methods, including those specially designed for semantic segmentation, which verifies the generalization and effectiveness of NOUN.
Inconsistent with observations in cross-domain object recognition tasks, PRONOUN does not further improve NOUN on both synthetic-to-real semantic segmentation tasks. A possible reason is that the estimated prototypes in semantic segmentation are inaccurate. Even so, PRONOUN achieves the second-best results on both segmentation benchmarks.

\setlength{\tabcolsep}{4.0pt}
\begin{table}[!htbp]
	\small
	\centering
	\caption{Ablation study: the average accuracy (\%) of different variants of NOUN on object recognition benchmarks.}
	\resizebox{0.45\textwidth}{!}{$
		\begin{tabular}{lcccc}
			\toprule
			Ablation & \emph{Home} & \emph{VisDA} & \emph{Office31} & \emph{CLEF} \\
			\midrule
			NOUN ($k=1$) &66.7& 78.9& 87.3& 88.5 \\
			\midrule
			NOUN+E ($k=1$) &68.9({\textcolor{red}{$\uparrow$}})& 79.8({\textcolor{red}{$\uparrow$}})& 88.1({\textcolor{red}{$\uparrow$}})& 88.5(-) \\
			PRONOUN ($k=1$)  & 68.1({\textcolor{red}{$\uparrow$}}) & 80.0({\textcolor{red}{$\uparrow$}}) & 88.0({\textcolor{red}{$\uparrow$}}) &88.6({\textcolor{red}{$\uparrow$}}) \\
			NOUN ($k=3$) &66.2({\textcolor{blue}{$\downarrow$}})& 79.5({\textcolor{red}{$\uparrow$}})& 86.9({\textcolor{blue}{$\downarrow$}})& 88.5(-) \\
			NOUN+E ($k=3$) &69.1({\textcolor{red}{$\uparrow$}})& 80.6({\textcolor{red}{$\uparrow$}}) & 88.2({\textcolor{red}{$\uparrow$}})& 88.6({\textcolor{red}{$\uparrow$}}) \\
			PRONOUN ($k=3$)  & \textbf{70.7}({\textcolor{red}{$\uparrow$}}) & \textbf{81.6}({\textcolor{red}{$\uparrow$}}) & \textbf{88.8}({\textcolor{red}{$\uparrow$}}) & \textbf{89.0}({\textcolor{red}{$\uparrow$}}) \\
			\bottomrule
		\end{tabular}
		$}
	\label{tab:ablatn}
\end{table}

\begin{figure*}[!htbp]
	\centering
	\footnotesize
	\setlength\tabcolsep{1.0mm}
	\renewcommand\arraystretch{0.1}
	\begin{tabular}{cccc}
		\includegraphics[width=0.24\linewidth,trim={0.0cm 0.0cm 0.0cm 0.0cm}, clip]{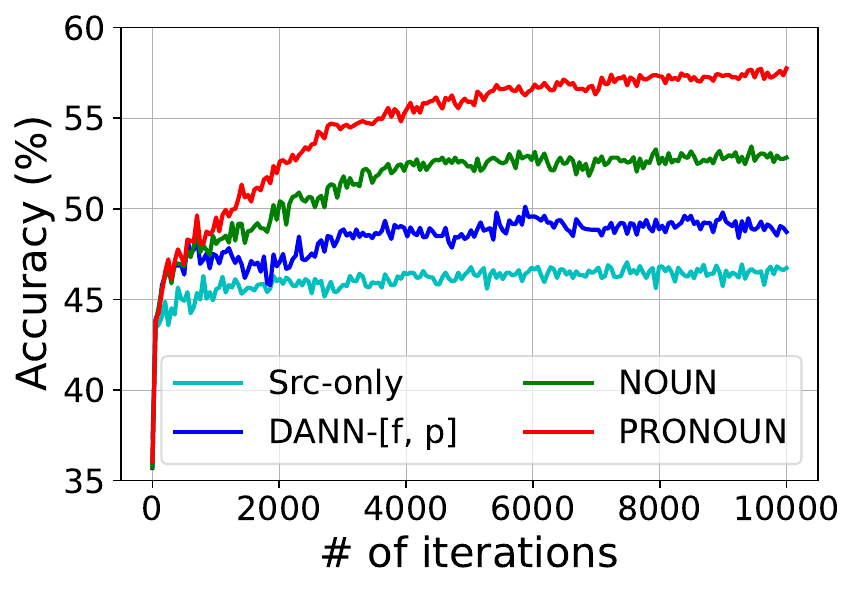} &
		\includegraphics[width=0.24\linewidth,trim={0.0cm 0.0cm 0.0cm 0.0cm}, clip]{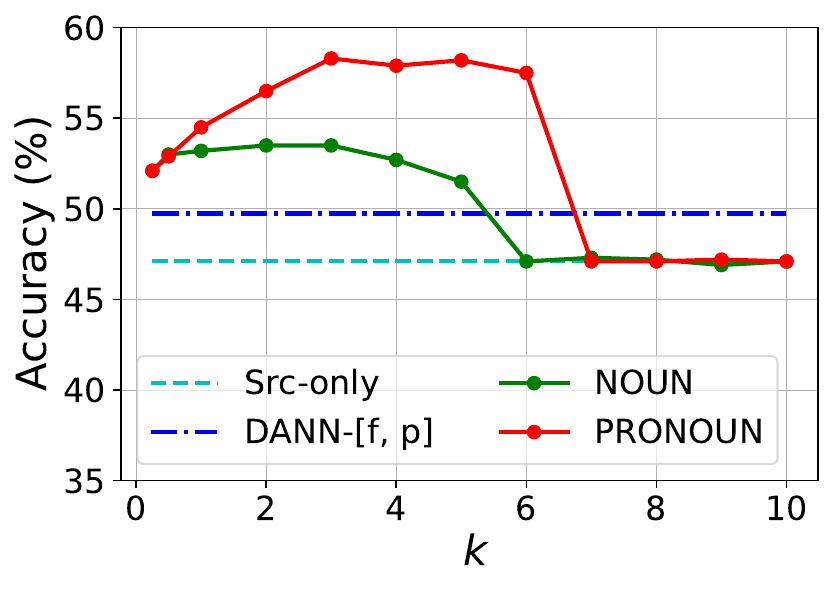} &
		\includegraphics[width=0.24\linewidth,trim={0.0cm 0.0cm 0.0cm 0.0cm}, clip]{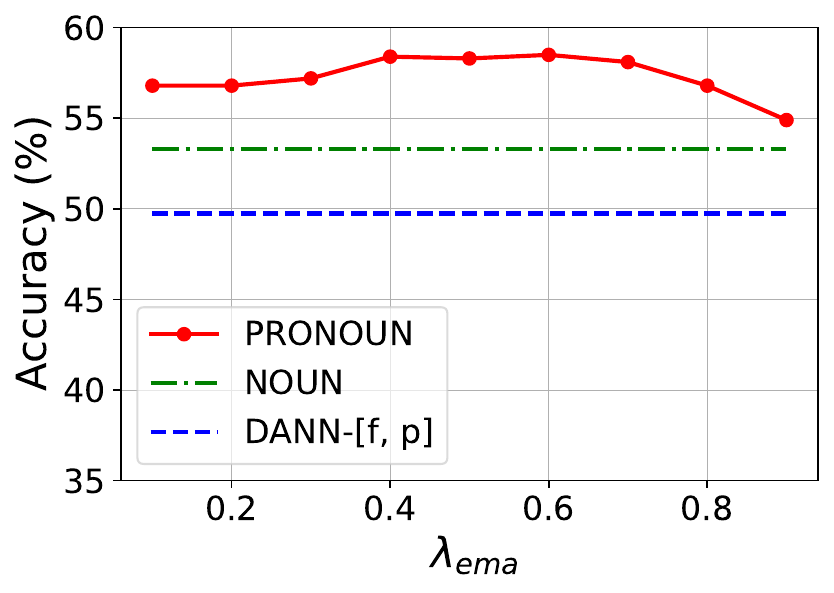} &
		\includegraphics[width=0.24\linewidth,trim={0.0cm 0.0cm 0.0cm 0.0cm}, clip]{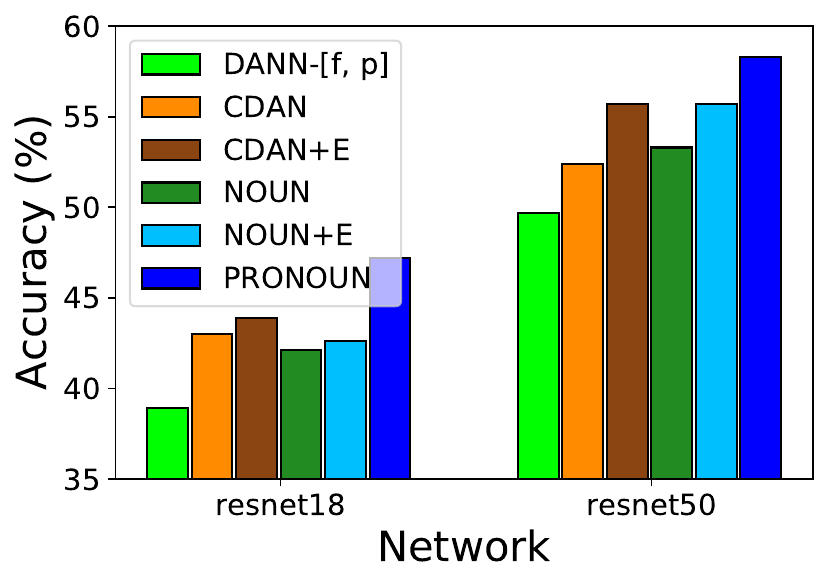} \\
		~\\
		(a) convergence & (b) sensitivity of $k$ & (c) sensitivity of $\lambda_{ema}$ & (d) backbone network
	\end{tabular}
	\caption{Further analysis of NOUN and PRONOUN on the task Ar$\to$Cl on \emph{Office-Home}.}
	\label{fig:ablatn}
\end{figure*} 

\subsection{Model analysis and discussions.}

\subsubsection{Ablation study}
The benefit of enlarging the norm of concatenated predictions introduced in NOUN has been demonstrated by comparing DANN-[f, p] and NOUN. To validate the effectiveness of the prototype-based conditioning in PRONOUN, we further conduct ablation on the four classification benchmarks and show the respective average accuracy in Table~\ref{tab:ablatn}. `NOUN+E' means combining the entropy conditioning~\cite{cdan} with NOUN. Increasing the norm control factor $k$ from 1 to 3, NOUN models show notable performance drop on \emph{Office-Home} and \emph{Office31}, while both NOUN+E and PRONOUN models achieve performance improvement on all benchmarks. This comparison indicates that NOUN suffers from inaccurate predictions with too strong conditioning. Both the entropy conditioning~\cite{cdan} and our prototype-based conditioning are able to alleviate the misalignment induced by inaccurate predictions. In addition, the consistent improvement of NOUN+E and PRONOUN by simply increasing the norm control factor $k$, in turn, demonstrates that a larger norm of concatenated predictions can contribute to more effective conditional domain alignment. With stronger conditioning, i.e., $k=3$, PRONOUN evidently surpasses the competitive counterpart of NOUN+E on all benchmarks, which explicitly verifies the superiority of our prototype-based conditioning.

\subsubsection{Training stability}
We investigate the convergence of our methods by testing the model every 50 iterations during the training and illustrate the empirical accuracy curves of Ar$\to$Cl on \emph{Office-Home} in Fig.~\ref{fig:ablatn}(a). It is obvious that for both NOUN and PRONOUN, the model training is stable and finally converges.

\subsubsection{Parameter sensitivity}
The introduced hyper-parameters are $k$ for both NOUN and PRONOUN and $\lambda_{ema}$ only for PRONOUN. We find NOUN generally performs well with $k$ of 1, i.e., NOUN can work without extra hyper-parameters. Thus we mainly investigate the sensitivity of $k$ and $\lambda_{ema}$ for PRONOUN on the UDA task Ar$\to$Cl on \emph{Office-Home} and show results in Fig.~\ref{fig:ablatn}(b) and Fig.~\ref{fig:ablatn}(c), respectively. For $k$, we observe that within a moderate range, i.e., when $k$ is less than 3, increasing the norm of concatenated output brings improvement.
However, an especially large norm of output leads to performance degradation.
The moderate range is 7 for PRONOUN and 6 for NOUN, larger values of $k$ lead to the worst performance close to the non-adapted source model. Because with an extreme value of $k$, the detached concatenation part will dominate the discriminator optimization. Then the contribution of features to the adversarial loss would be ignored, making domain adversarial training ineffective. 
For $\lambda_{ema}$, we set it as 0.5 for all tasks without selection. As shown in Fig.~\ref{fig:ablatn}(c), 0.4--0.7 is a suitable range, and an extremely small or large value of $\lambda_{ema}$ would weaken the performance. The reason is that moderate $\lambda_{ema}$ can provide a better representation of semantic structures by keeping the prototypes both consistent and up-to-date, which is beneficial to PRONOUN.
Generally, the newly involved hyper-parameters in PRONOUN are not sensitive within the normal range.

\subsubsection{Network sensitivity}
Pseudo-labels play an important role in conditional domain adversarial training.  The network architecture has a great influence on the quality of pseudo-labels. Thus we evaluate the sensitivity of our methods to the network architecture. In Fig.~\ref{fig:ablatn}(d) we report the respective accuracy of DANN-[f, p], DANN, CDAN, CDAN+E, NOUN, NOUN+E and PRONOUN for Ar$\to$Cl on \emph{Office-Home} with two backbones of different capacity, i.e., ResNet-18 and ResNet-50, where `NOUN+E' denotes only replacing our prototype-based conditioning with the entropy conditioning \cite{cdan}. For the stronger model ResNet-50, NOUN slightly beats CDAN and PRONOUN outperforms both CDAN+E and NOUN+E. For ResNet-18 with worse pseudo-labels, NOUN slightly falls behind CDAN but still significantly exceeds DANN-[f, p], and PRONOUN outperforms other methods by a large margin. Generally, both of our methods show desirable robustness to the network change. NOUN always significantly outperforms DANN-[f, p], and PRONOUN always performs the best.

\setlength{\tabcolsep}{4.0pt}
\begin{table}[!htbp]
	\small
	\centering
	\caption{Compatibility with the self-training method: the average accuracy (\%) on object recognition benchmarks.}
	\resizebox{0.4\textwidth}{!}{$
		\begin{tabular}{lcccc}
			\toprule
			Methods & \emph{Home} & \emph{VisDA} & \emph{Office31}\\
			\midrule
			source-model-only~\cite{atdoc} & 59.4 & 49.1 & 76.5 \\
			ATDOC~\cite{atdoc} & 72.2 & 80.3 & 89.7 \\
			\midrule
			ATDOC+NOUN & 73.0 & 80.6& 90.0\\
			ATDOC+PRONOUN  & \textbf{74.1} & \textbf{81.9} & \textbf{90.4} \\
			\bottomrule
		\end{tabular}
		$}
	\label{tab:atdoc}
\end{table}

\subsubsection{Compatibility with self-training}
To evaluate the compatibility of NOUN and PRONOUN with other types of domain adaptation methods, we choose a state-of-the-art self-training method ATDOC~\cite{atdoc} for experiments. We report results on three object recognition benchmarks in Table~\ref{tab:atdoc}. We find both NOUN and PRONOUN consistently bring evident performance improvement when combined with ATDOC~\cite{atdoc}. Especially, PRONOUN helps ATDOC~\cite{atdoc} achieve the new-state-of-the-art average accuracy of $74.1\%$ on \emph{Office-Home}.

\subsubsection{Visualization}
\begin{figure*}[!htbp]
	\centering
	\footnotesize
	\setlength\tabcolsep{1mm}
	\renewcommand\arraystretch{0.1}
	\begin{tabular}{ccc}
		\includegraphics[width=0.32\linewidth,trim={1.0cm 1.0cm 1.0cm 1.0cm}, clip]{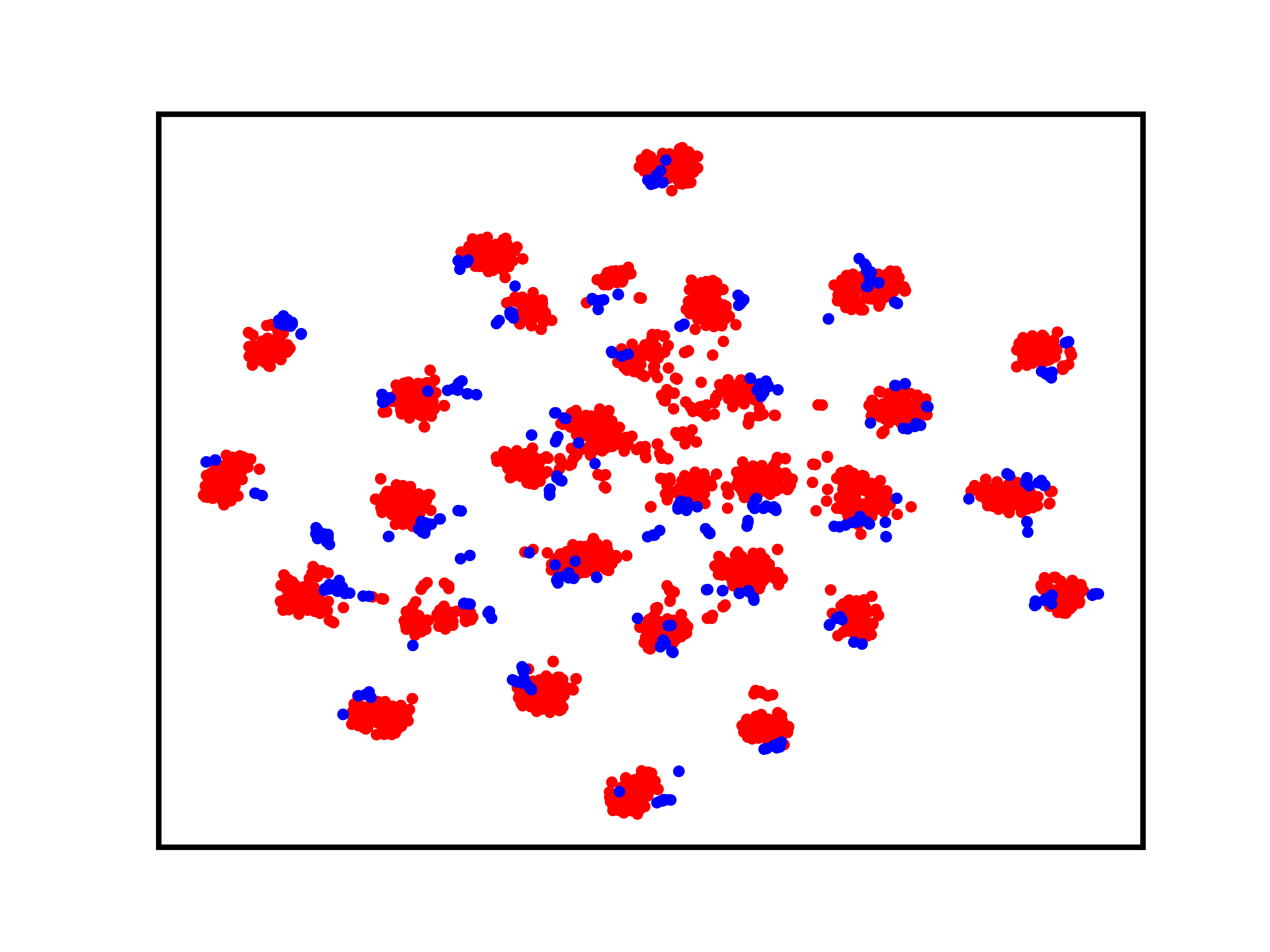} &
		\includegraphics[width=0.32\linewidth,trim={1.0cm 1.0cm 1.0cm 1.0cm}, clip]{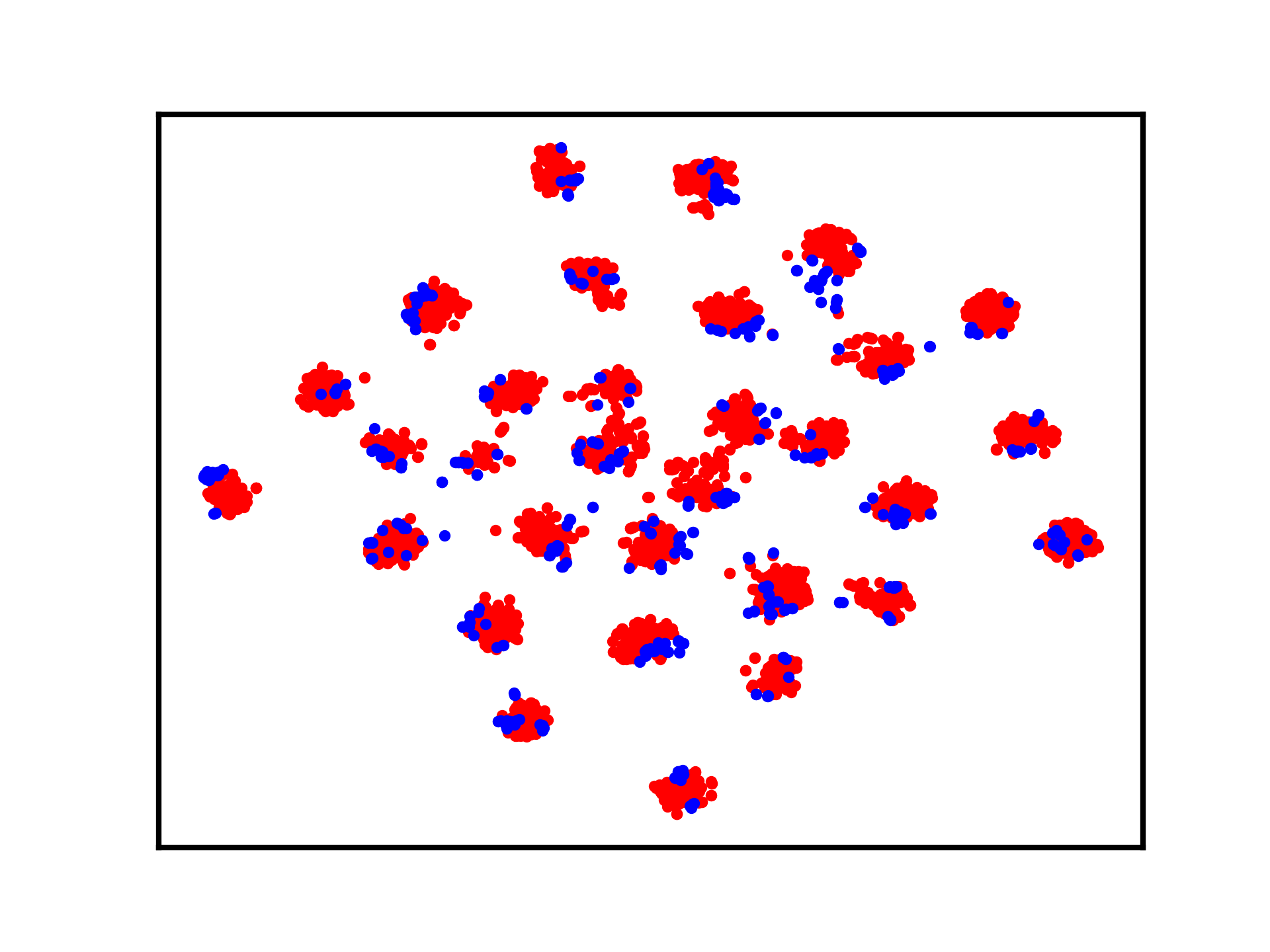} &
		\includegraphics[width=0.32\linewidth,trim={1.0cm 1.0cm 1.0cm 1.0cm}, clip]{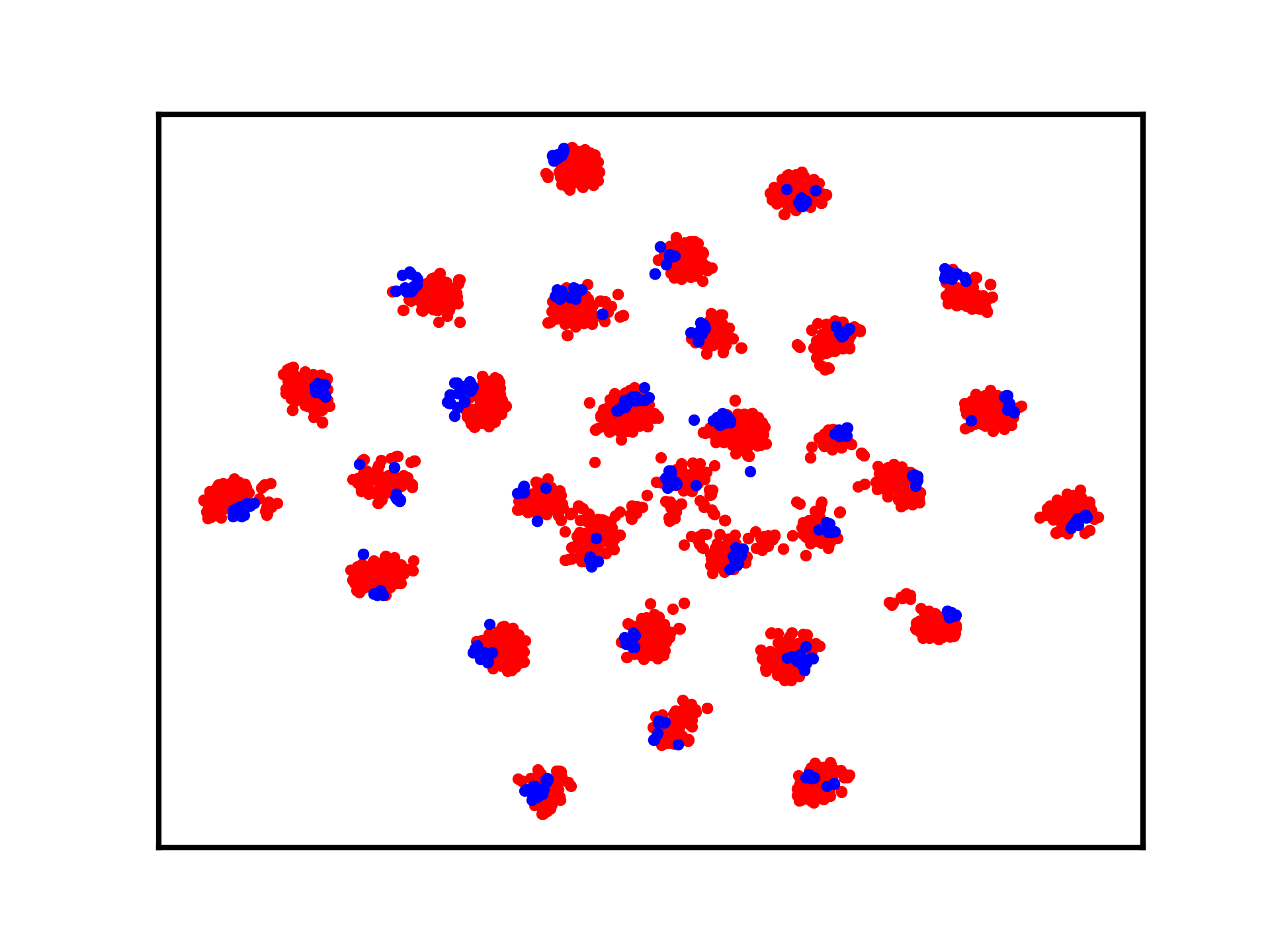} \\
		\includegraphics[width=0.32\linewidth,trim={1.0cm 1.0cm 1.0cm 1.0cm}, clip]{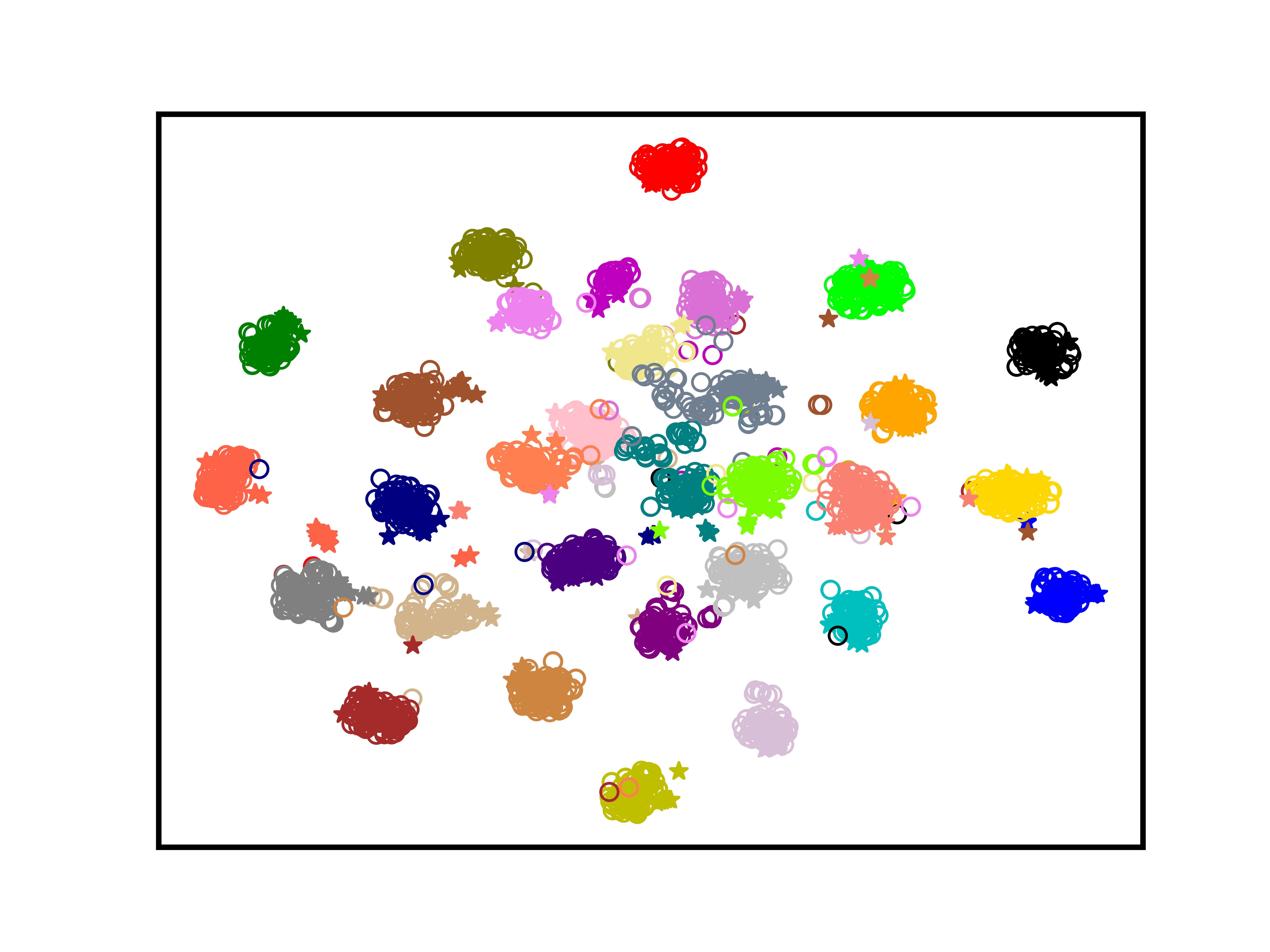} &
		\includegraphics[width=0.32\linewidth,trim={1.0cm 1.0cm 1.0cm 1.0cm}, clip]{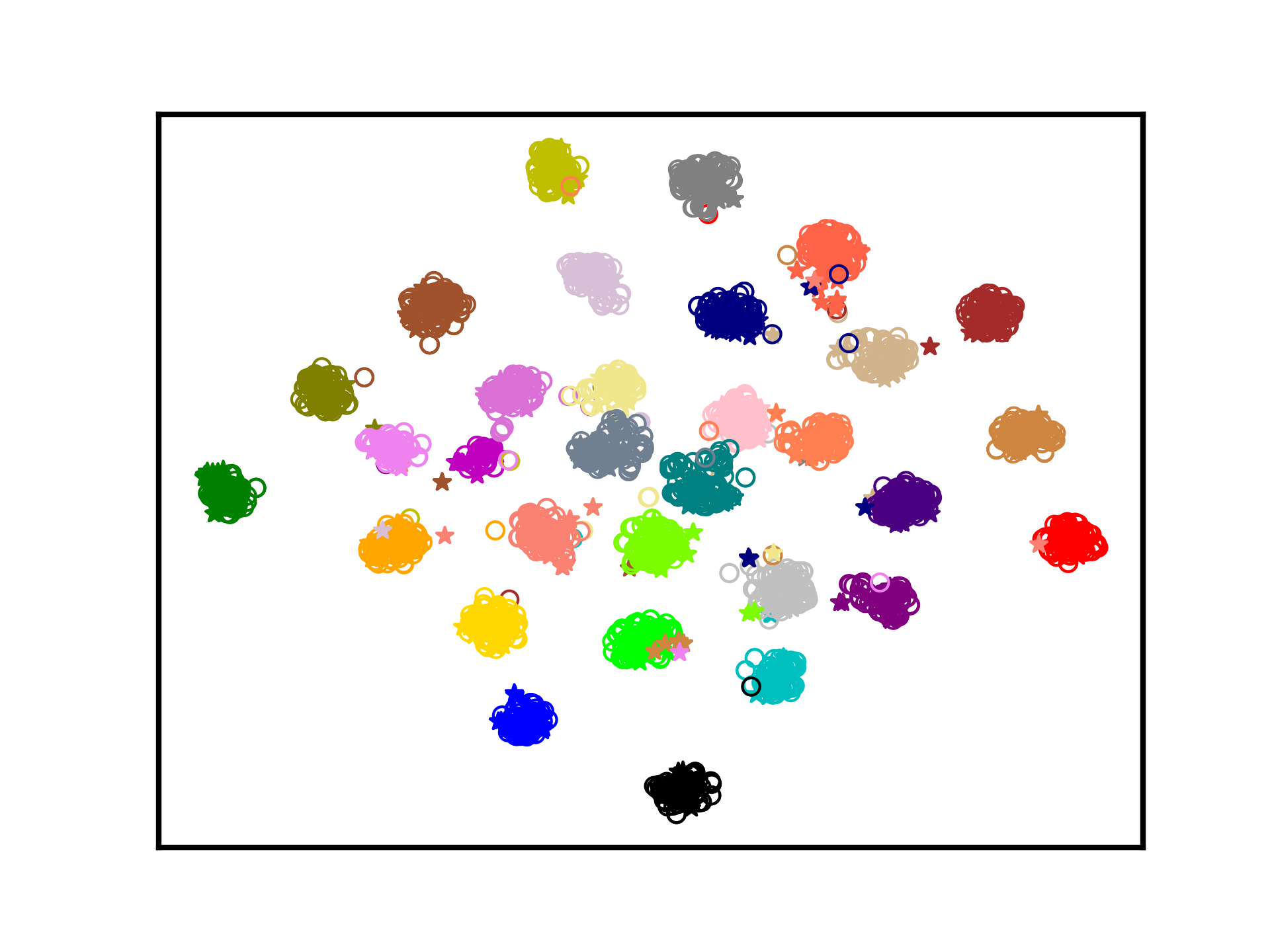} &
		\includegraphics[width=0.32\linewidth,trim={1.0cm 1.0cm 1.0cm 1.0cm}, clip]{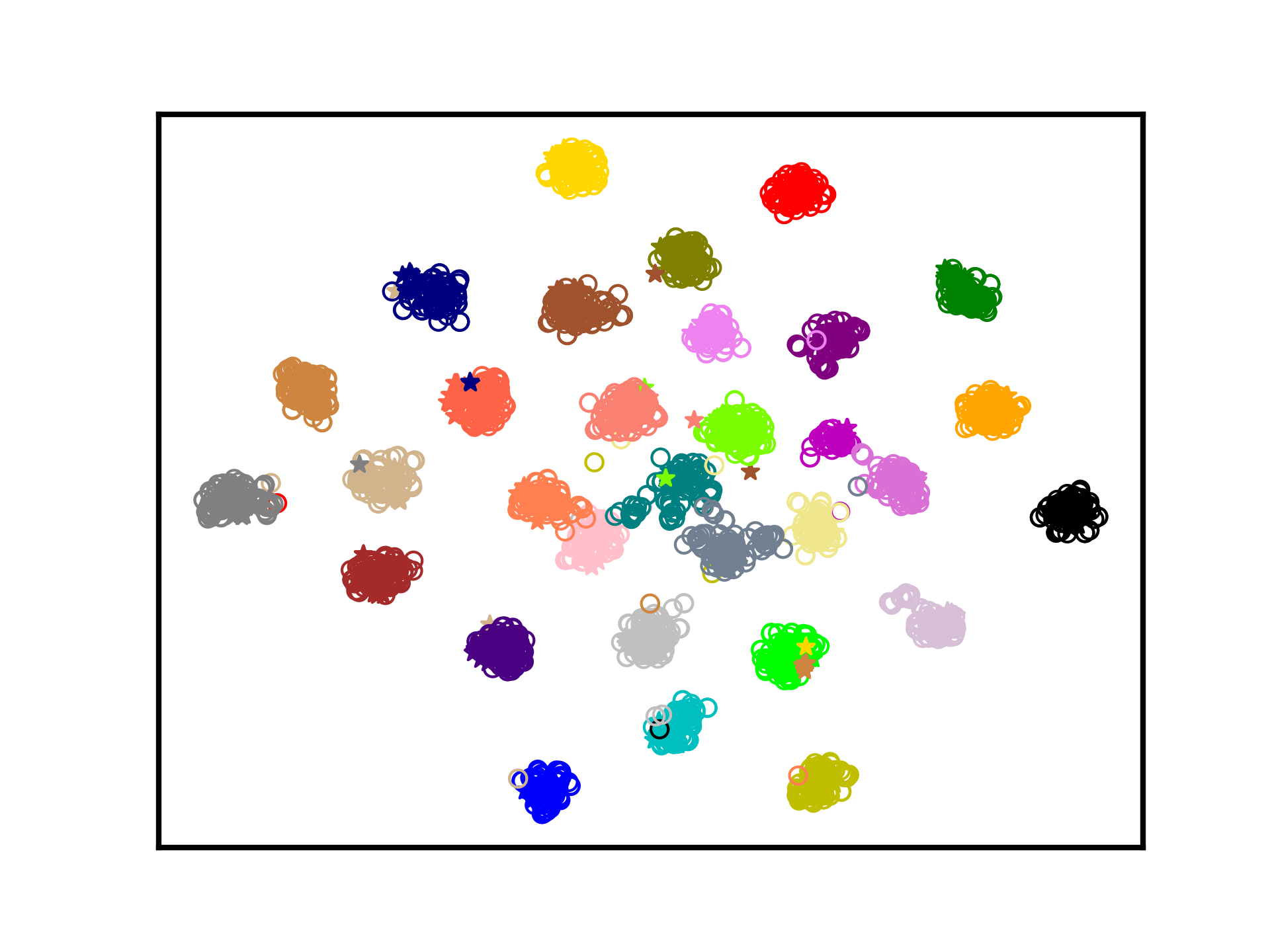} \\
		~\\
		(a) DANN-[f, p] & (b) NOUN & (c) PRONOUN
	\end{tabular}
	\caption{t-SNE~\cite{maaten2008visualizing} embedding visualizations of UDA methods for the A$\to$D task on \emph{Office31}. In the upper row, colors denote different domains (red: source, blue: target). In the bottom row, colors denote different classes, and shapes denote the domain information (source in $\circ$ and target in $\star$).}
	\label{fig:tsne}
\end{figure*}

\begin{figure*}[!htbp]
    \centering
    \footnotesize
    \setlength\tabcolsep{0.3mm}
    \renewcommand\arraystretch{1.0}
    \begin{tabular}{ccccc}
    \rotatebox{90}{\ \ \ Image} &
    \includegraphics[width=0.23\linewidth]{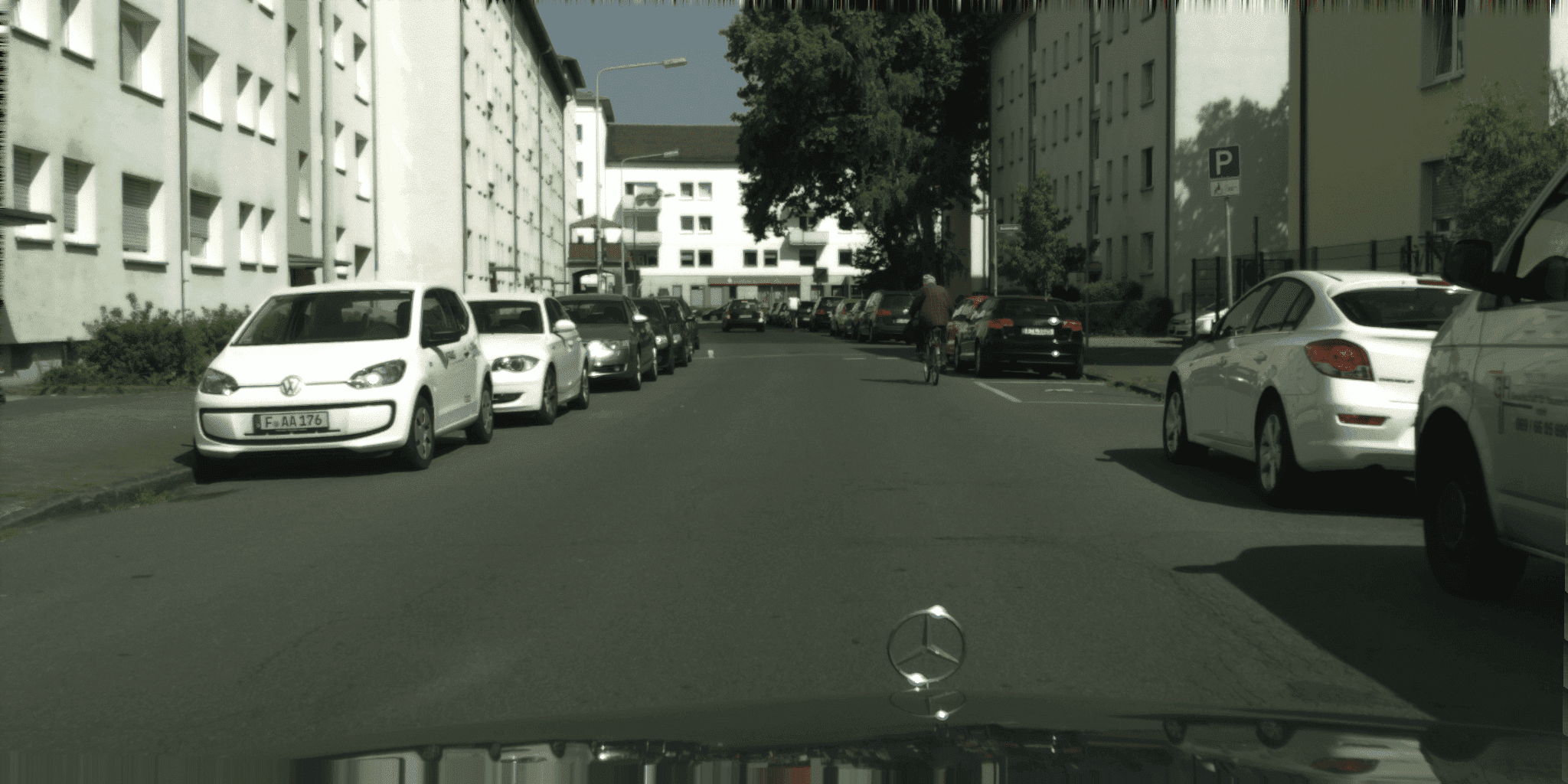} & 
    \includegraphics[width=0.23\linewidth]{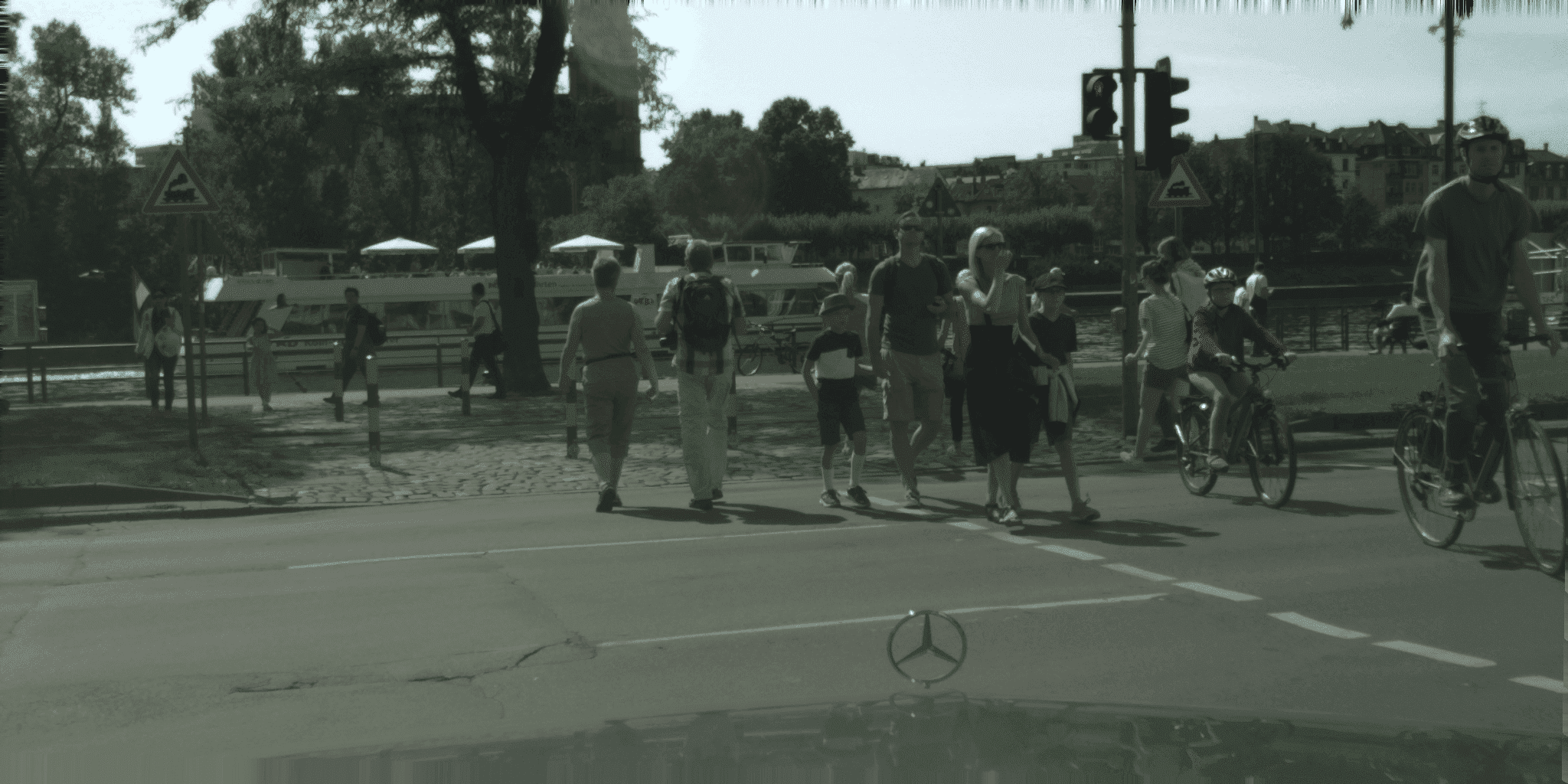} & 
    \includegraphics[width=0.23\linewidth]{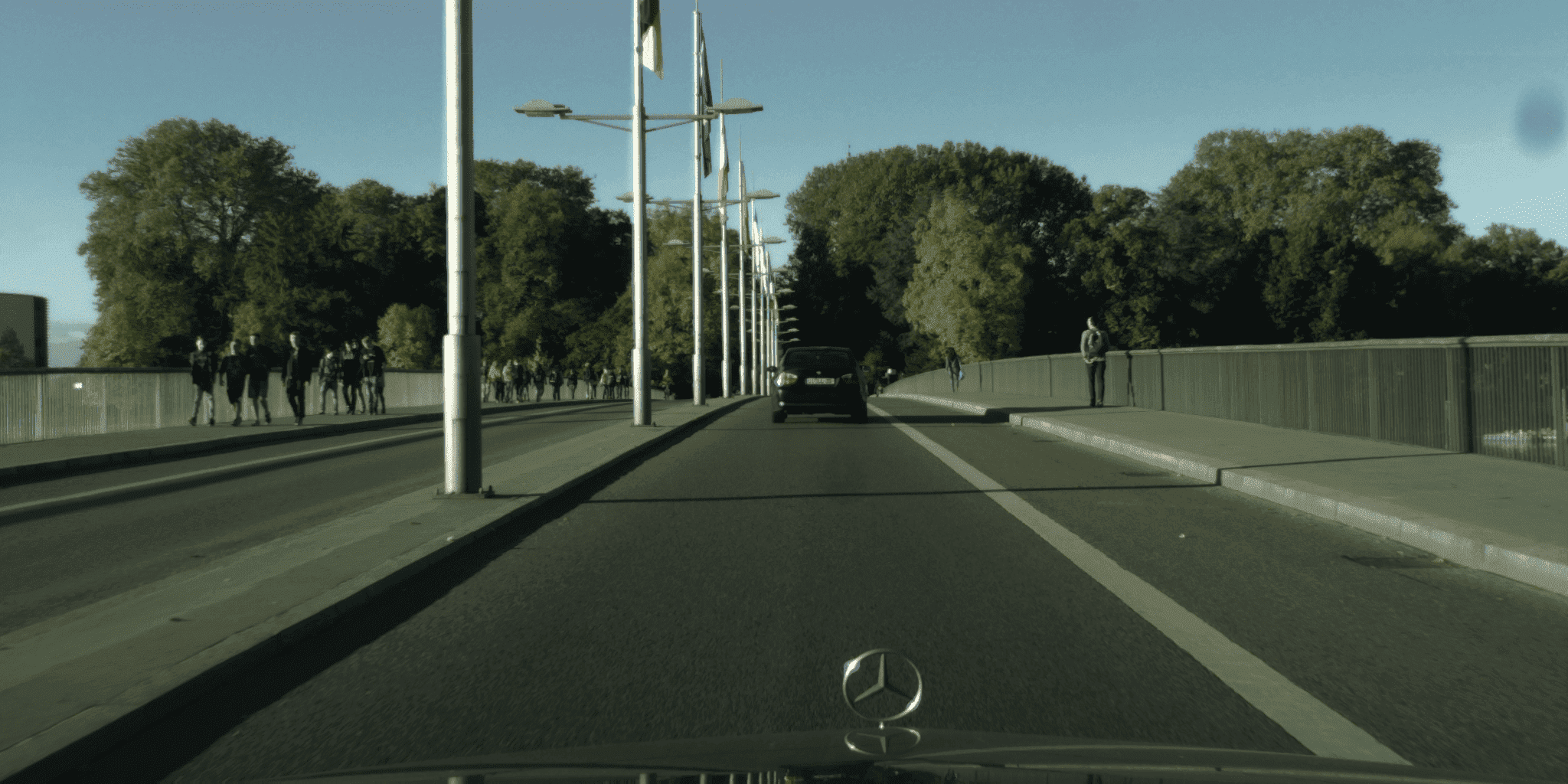} & 
    \includegraphics[width=0.23\linewidth]{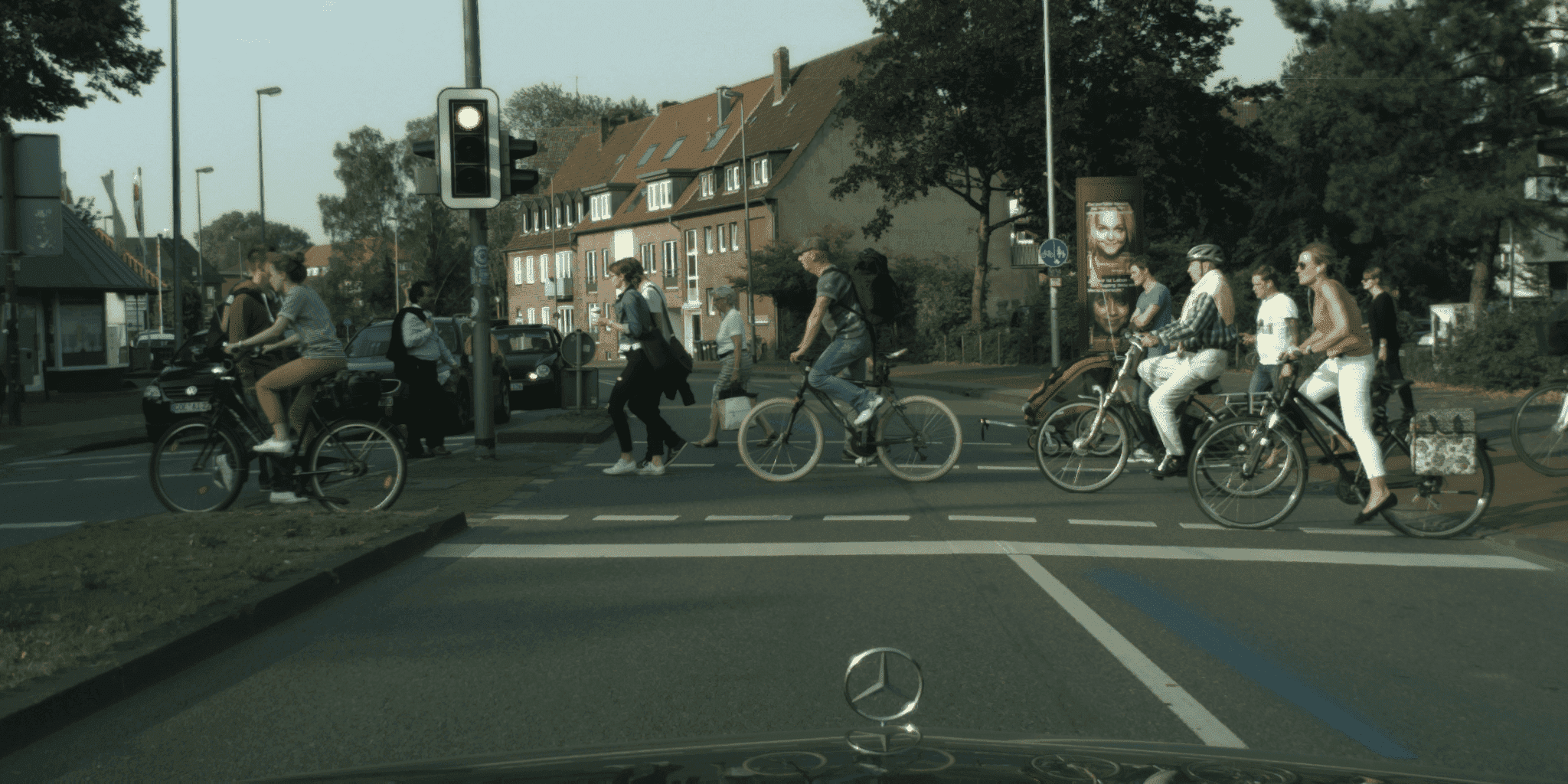} \\
    
    \rotatebox{90}{\ \ \ GT} &
    \includegraphics[width=0.23\linewidth]{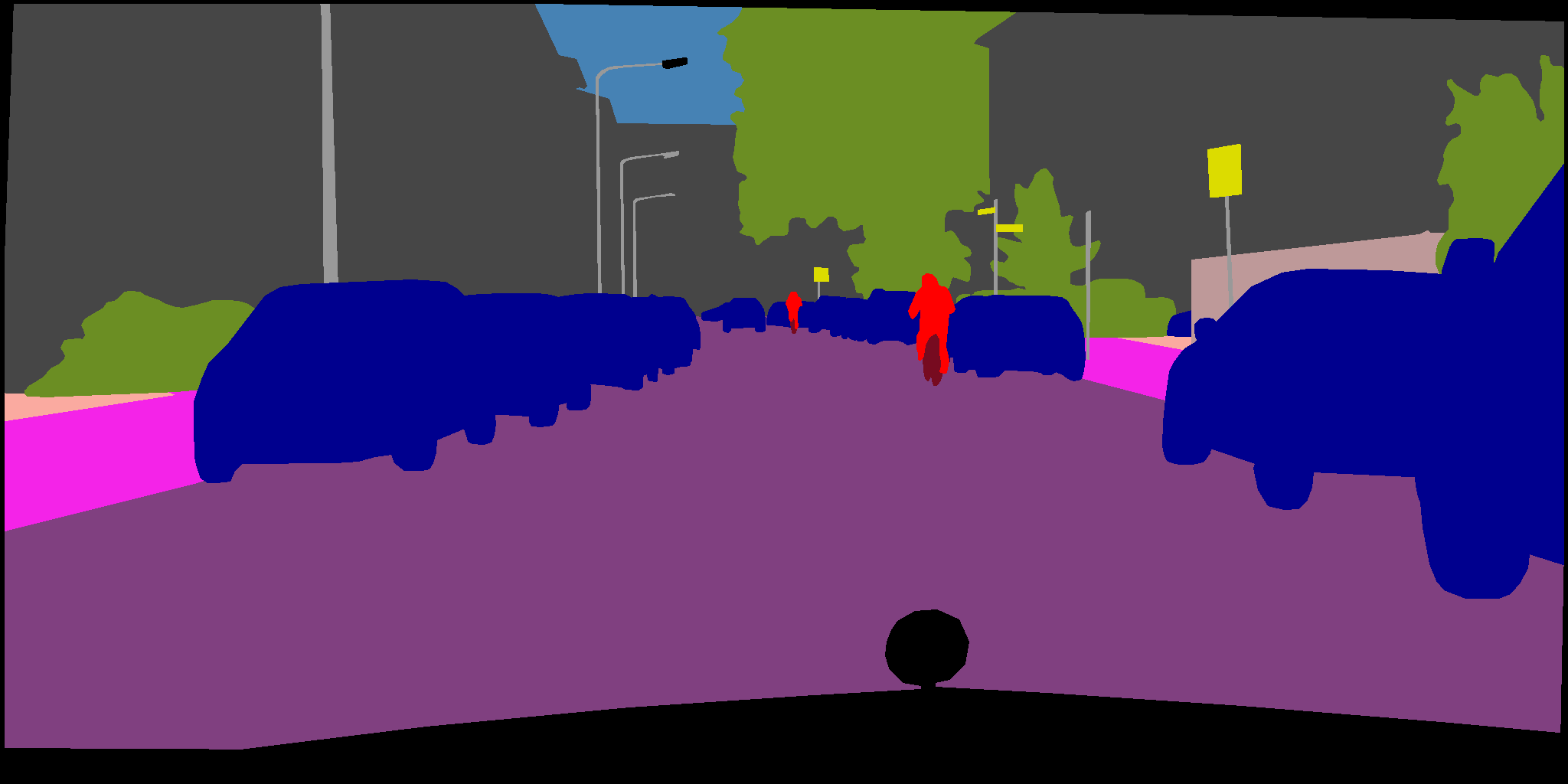} & 
    \includegraphics[width=0.23\linewidth]{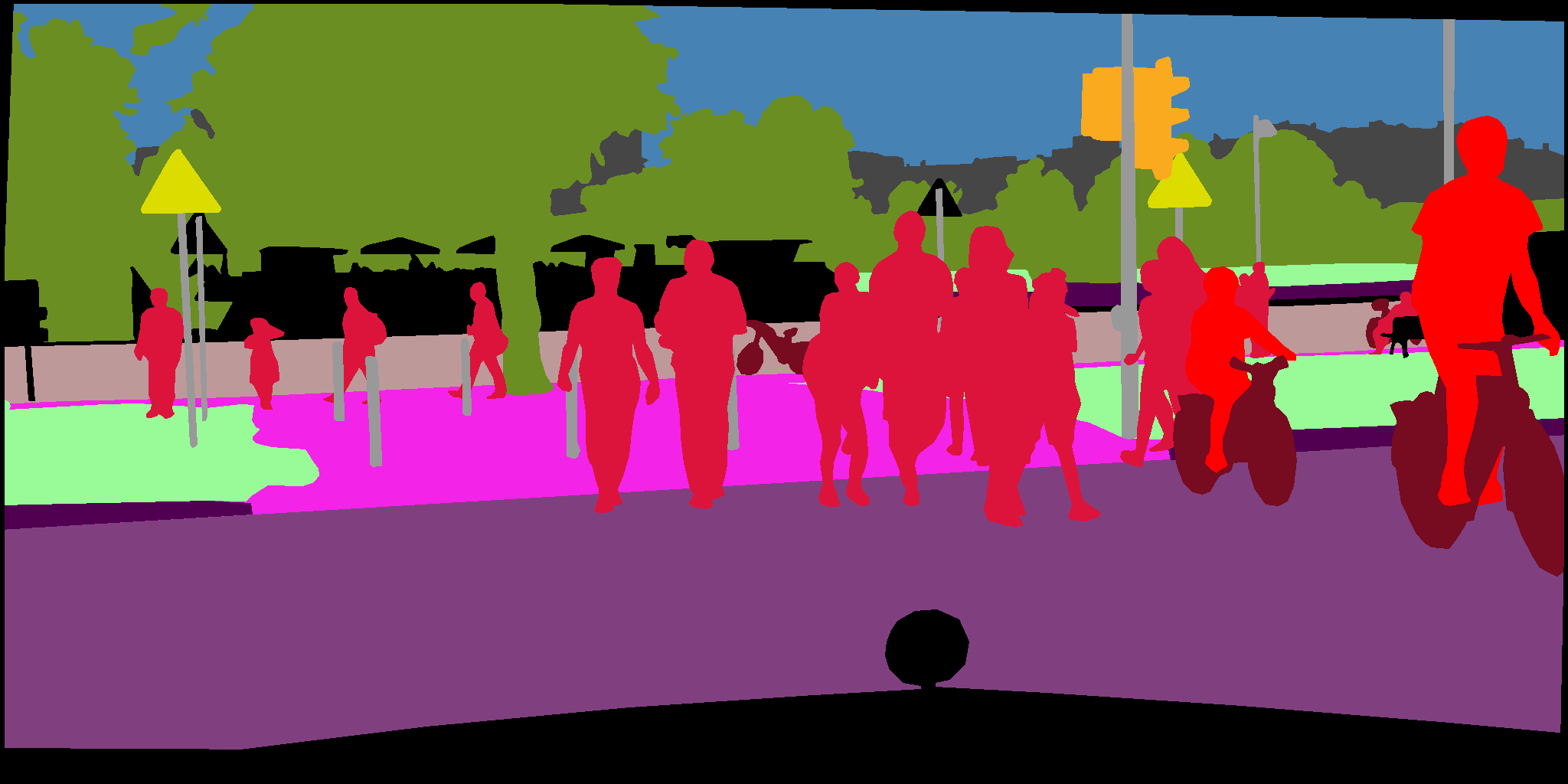} & 
    \includegraphics[width=0.23\linewidth]{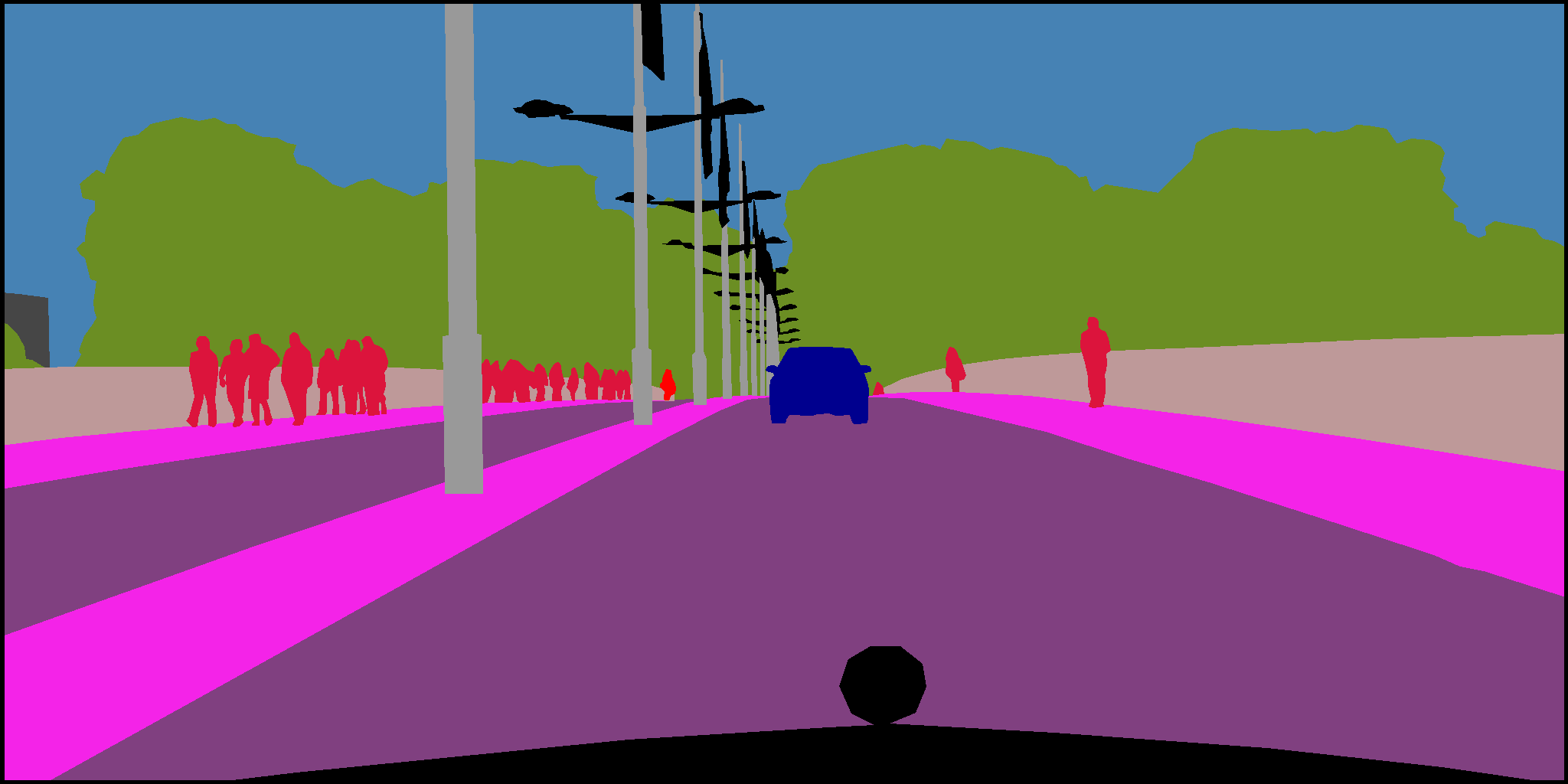} & 
    \includegraphics[width=0.23\linewidth]{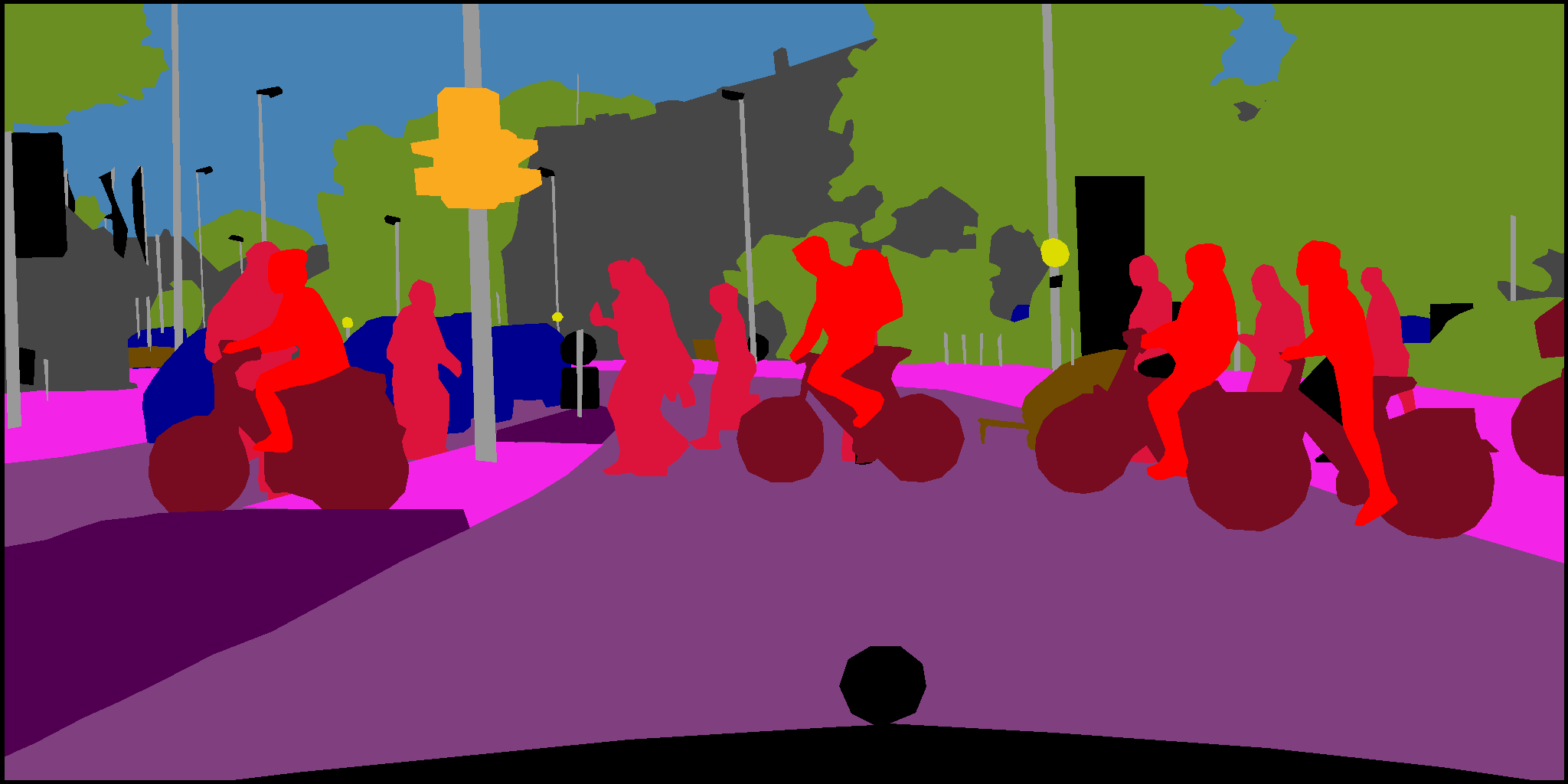} \\
    
    \rotatebox{90}{\ \ \ NonAdapt} &
    \includegraphics[width=0.23\linewidth]{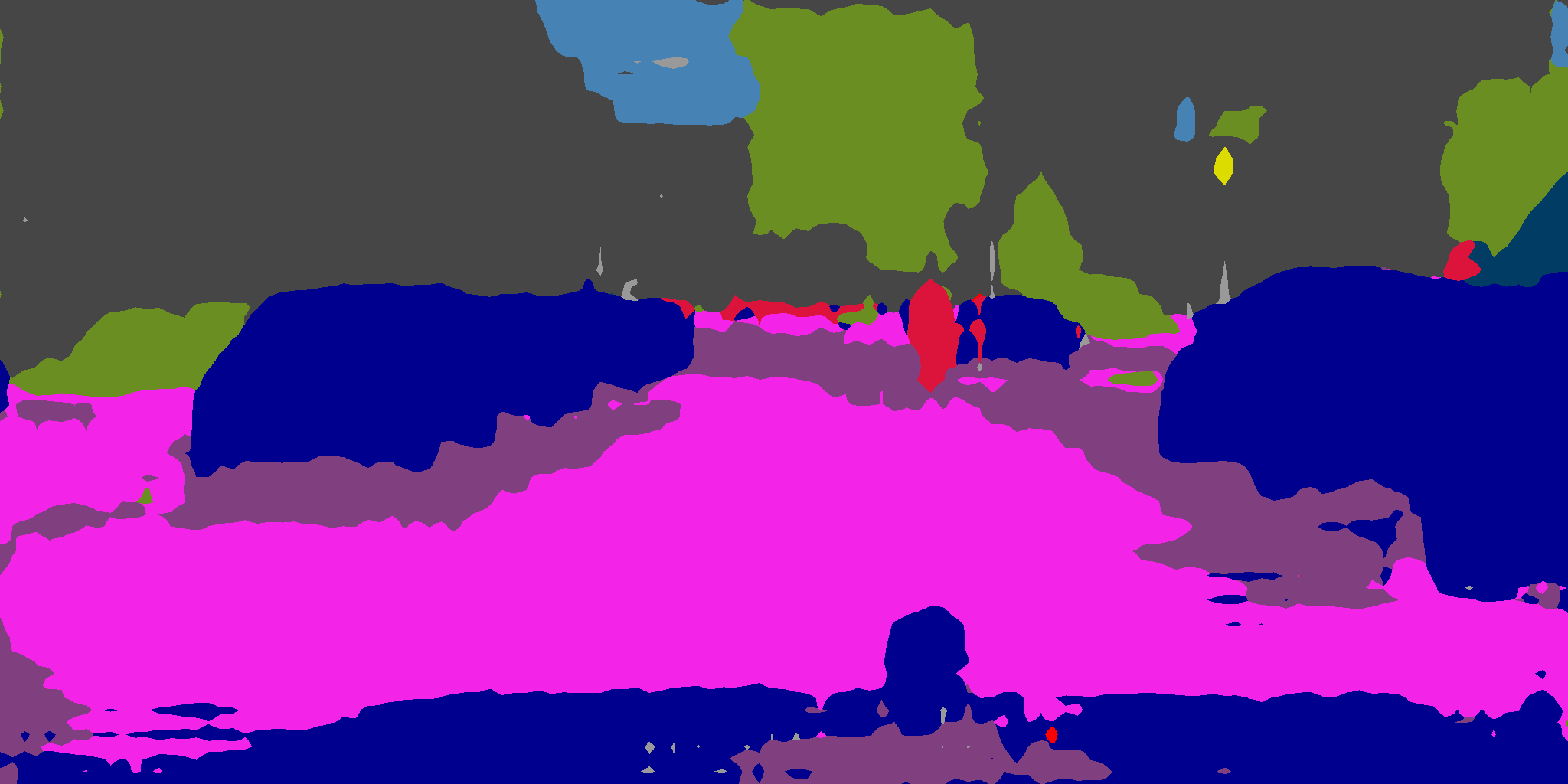} & 
    \includegraphics[width=0.23\linewidth]{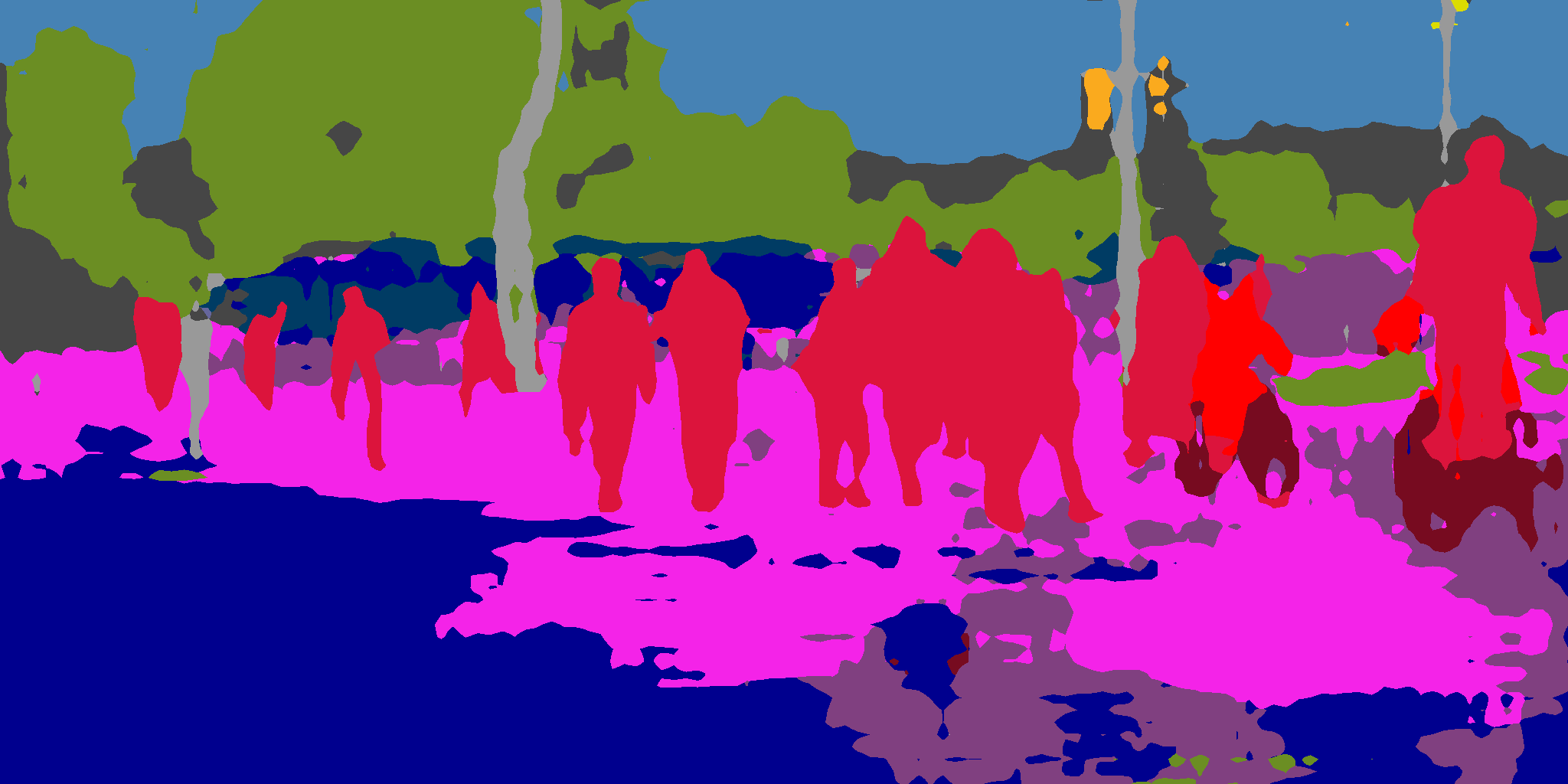} & 
    \includegraphics[width=0.23\linewidth]{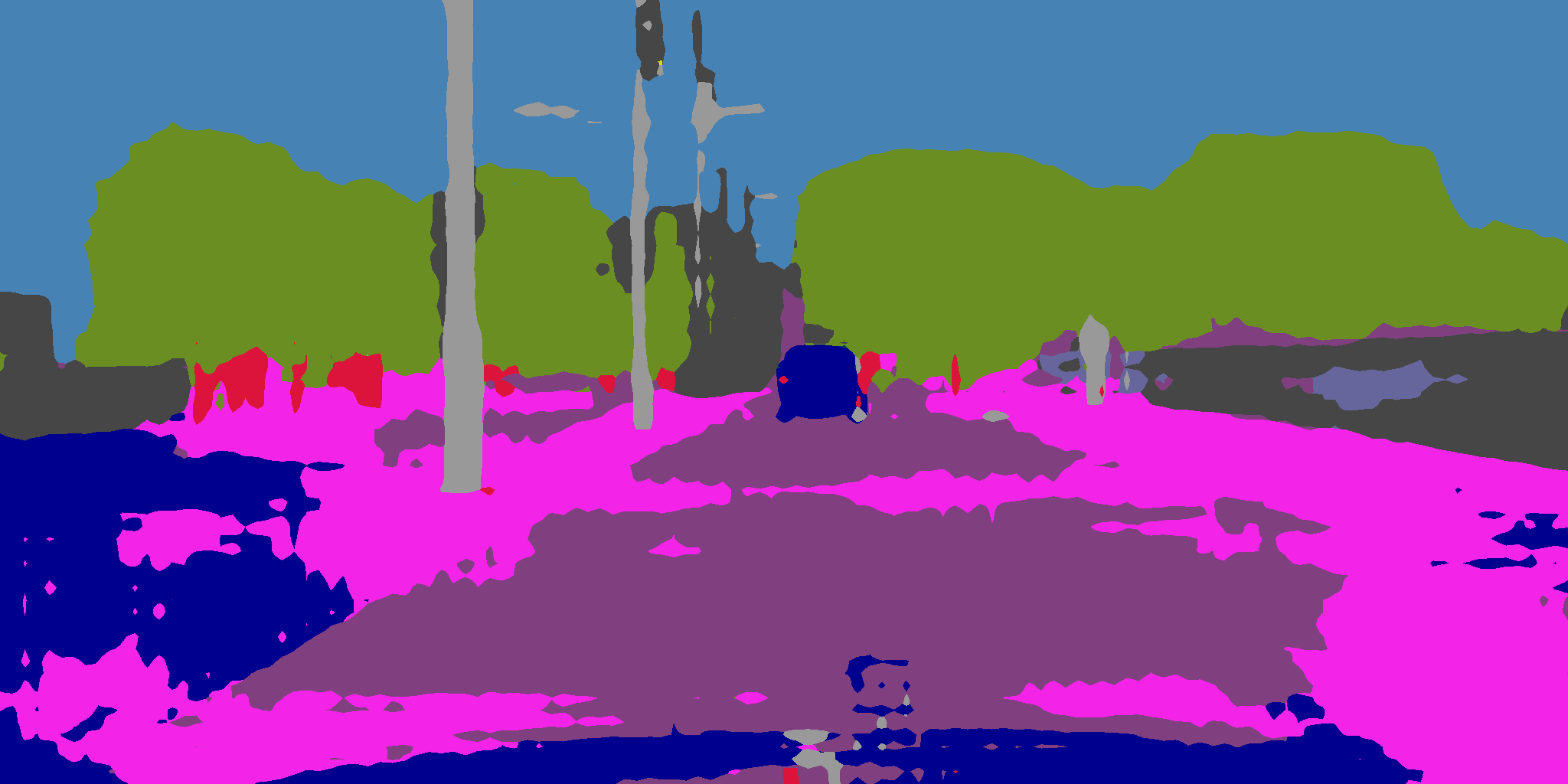} &
    \includegraphics[width=0.23\linewidth]{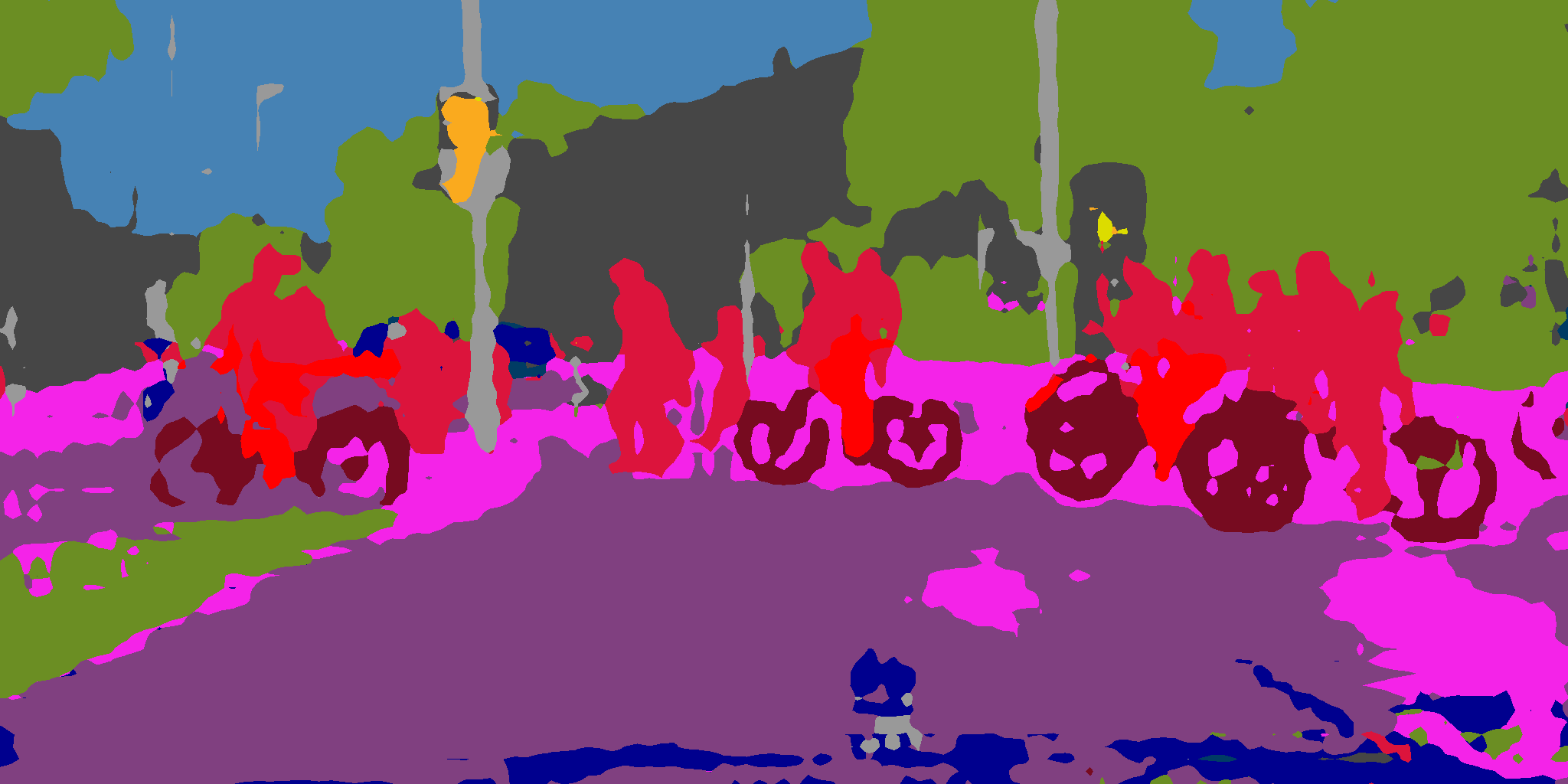} \\
    
    \rotatebox{90}{\ \ \ FeatAdapt} &
    \includegraphics[width=0.23\linewidth]{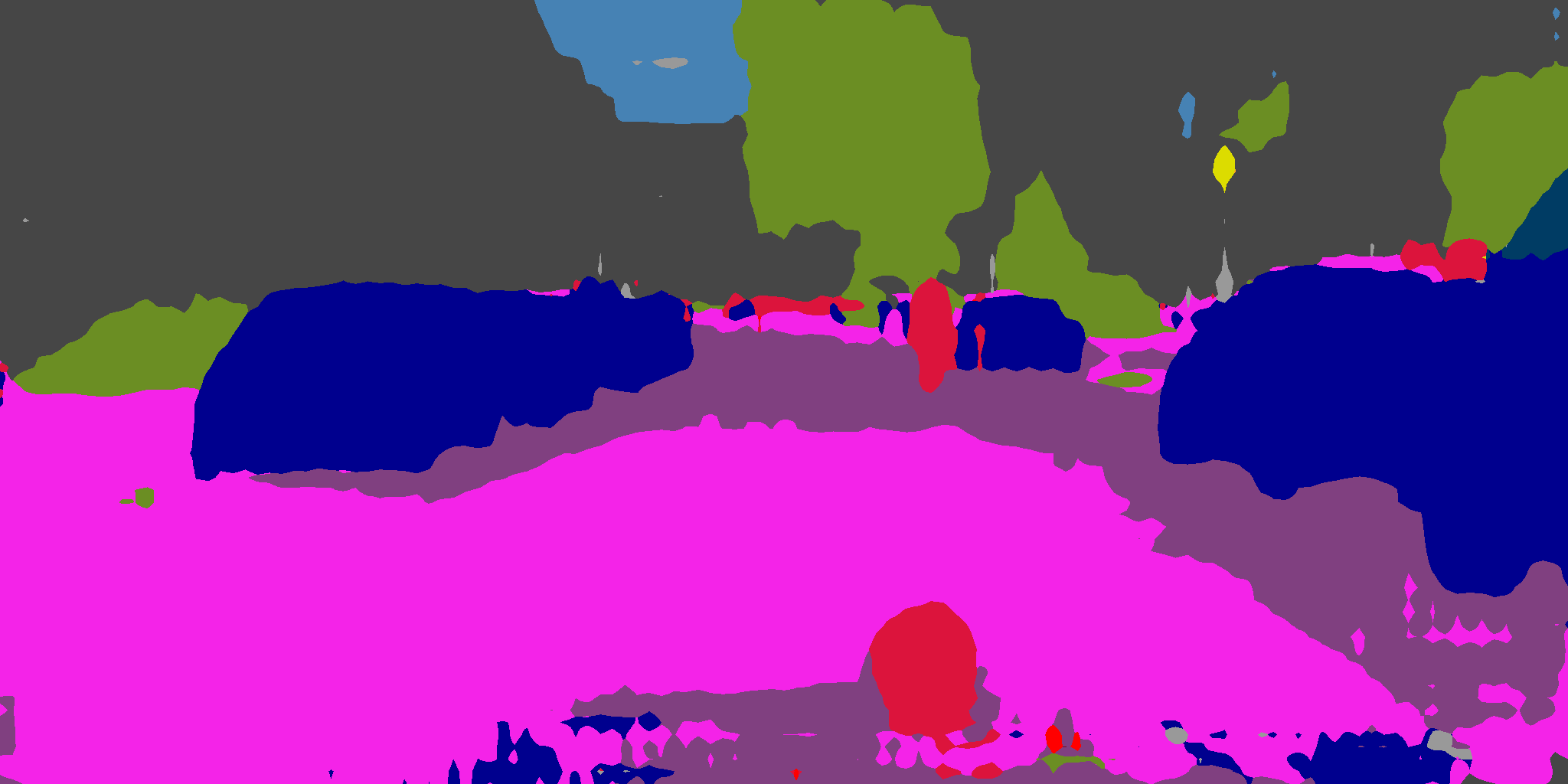} &
    \includegraphics[width=0.23\linewidth]{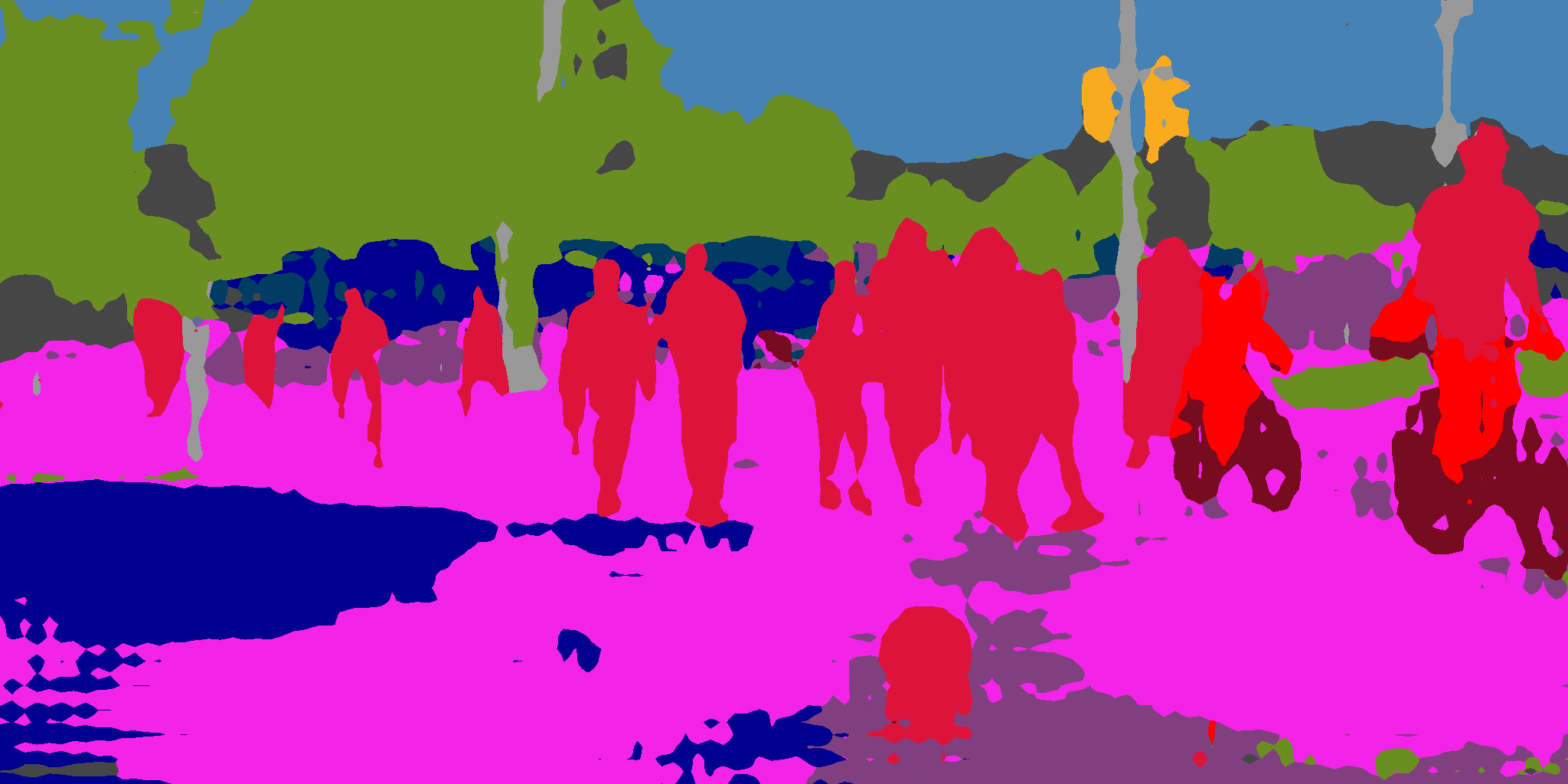} &
    \includegraphics[width=0.23\linewidth]{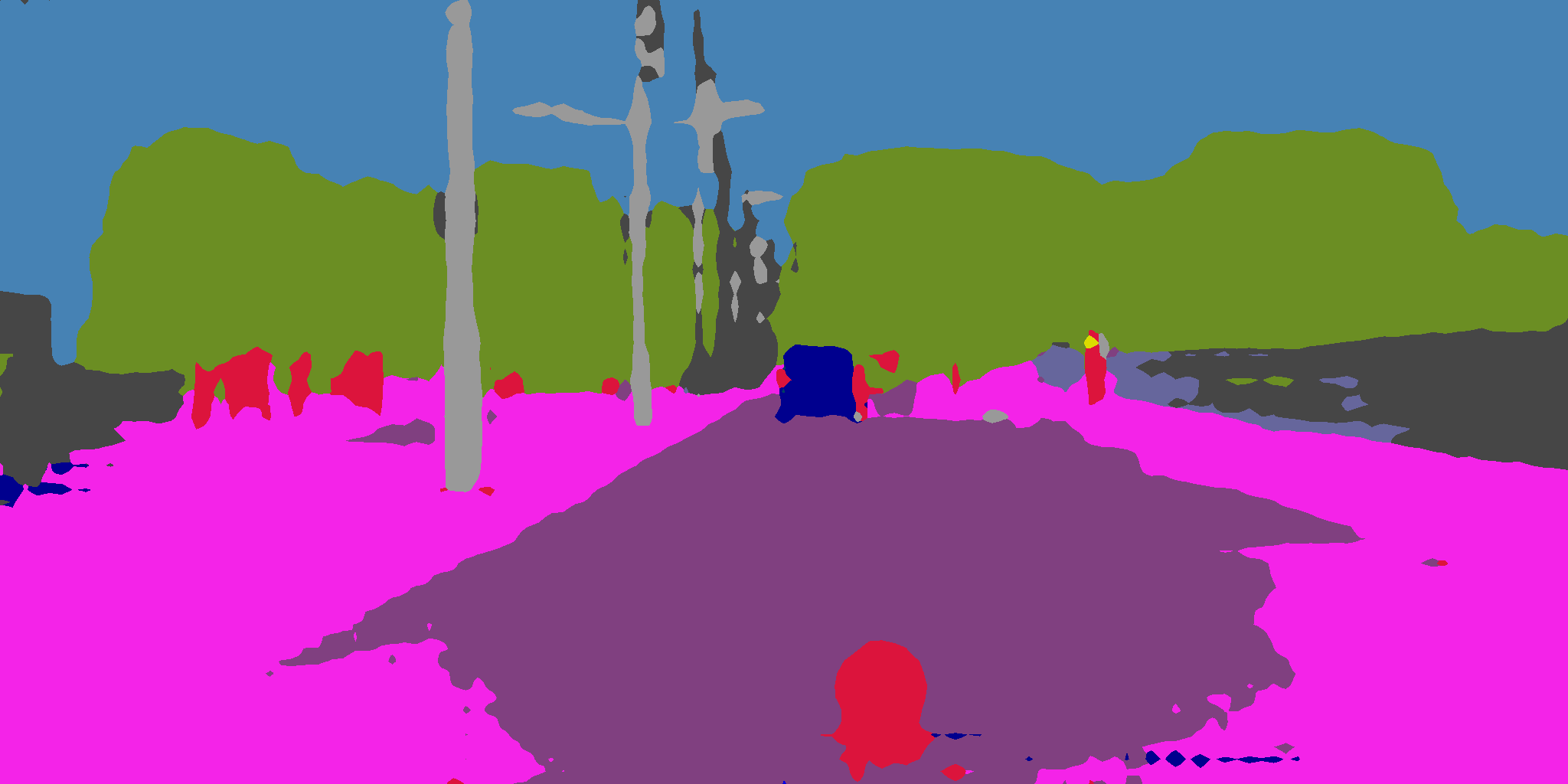} &
    \includegraphics[width=0.23\linewidth]{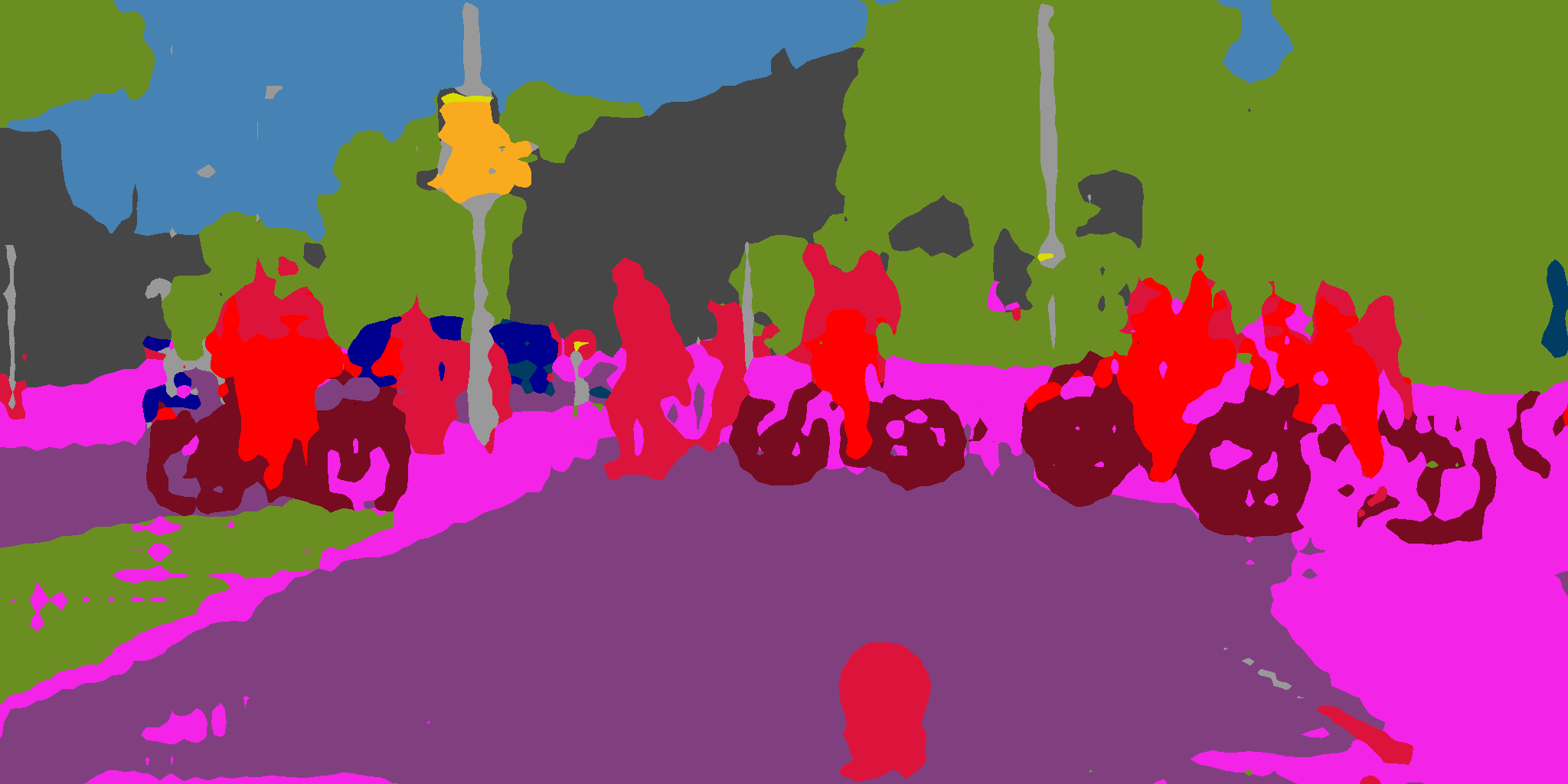} \\
    
    \rotatebox{90}{\ \ \ NOUN}  &
    \includegraphics[width=0.23\linewidth]{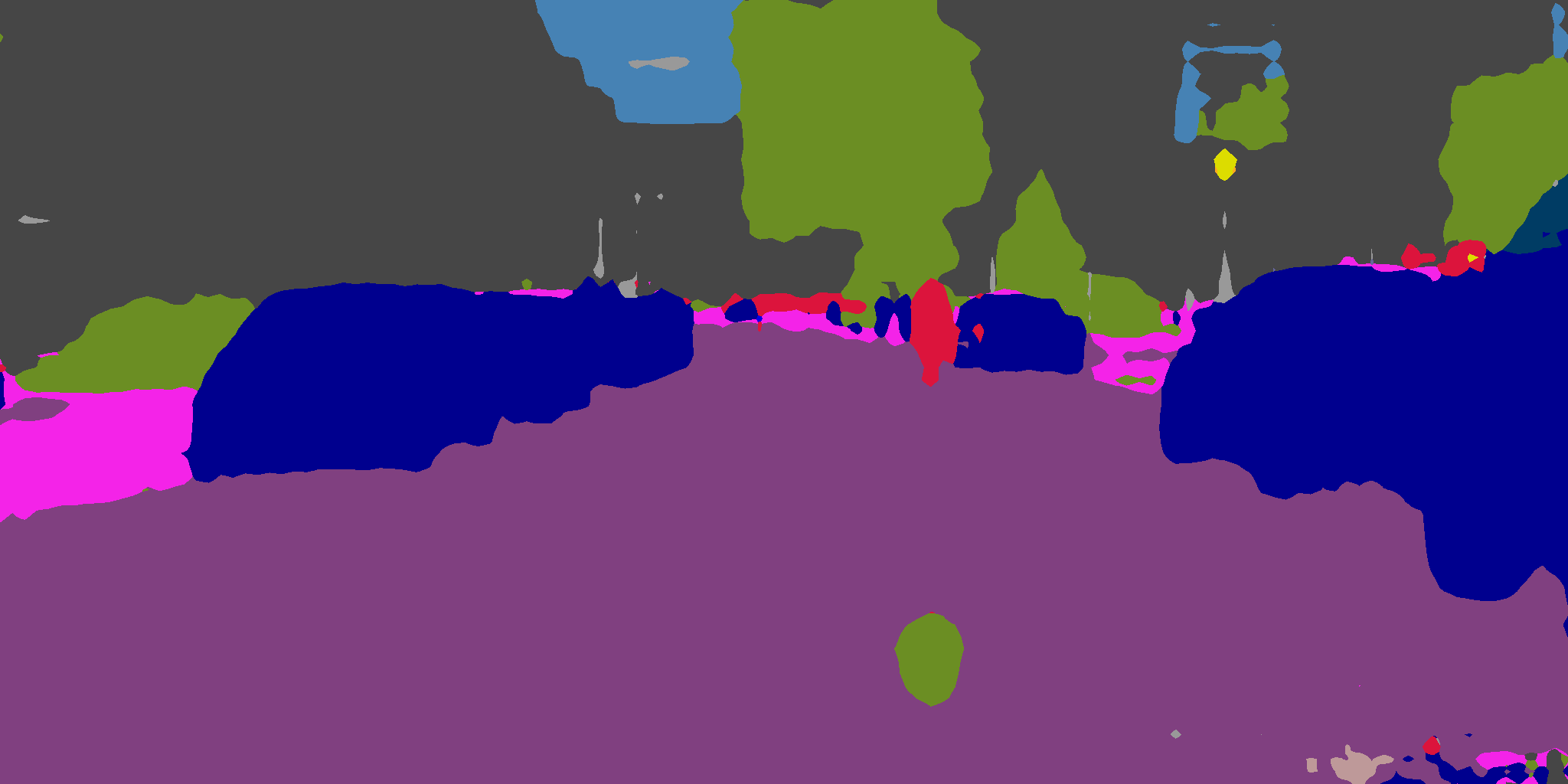} &
    \includegraphics[width=0.23\linewidth]{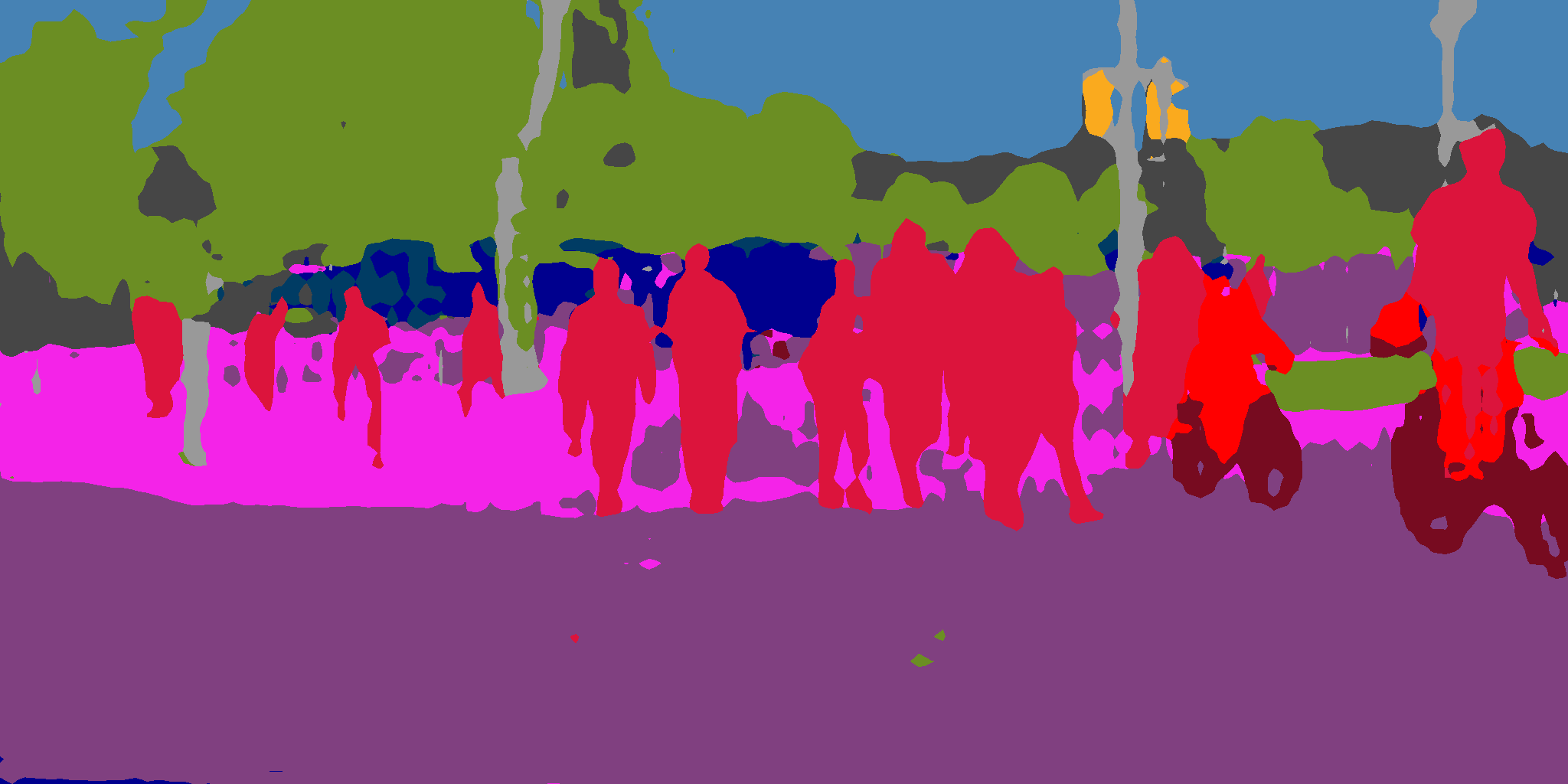} &
    \includegraphics[width=0.23\linewidth]{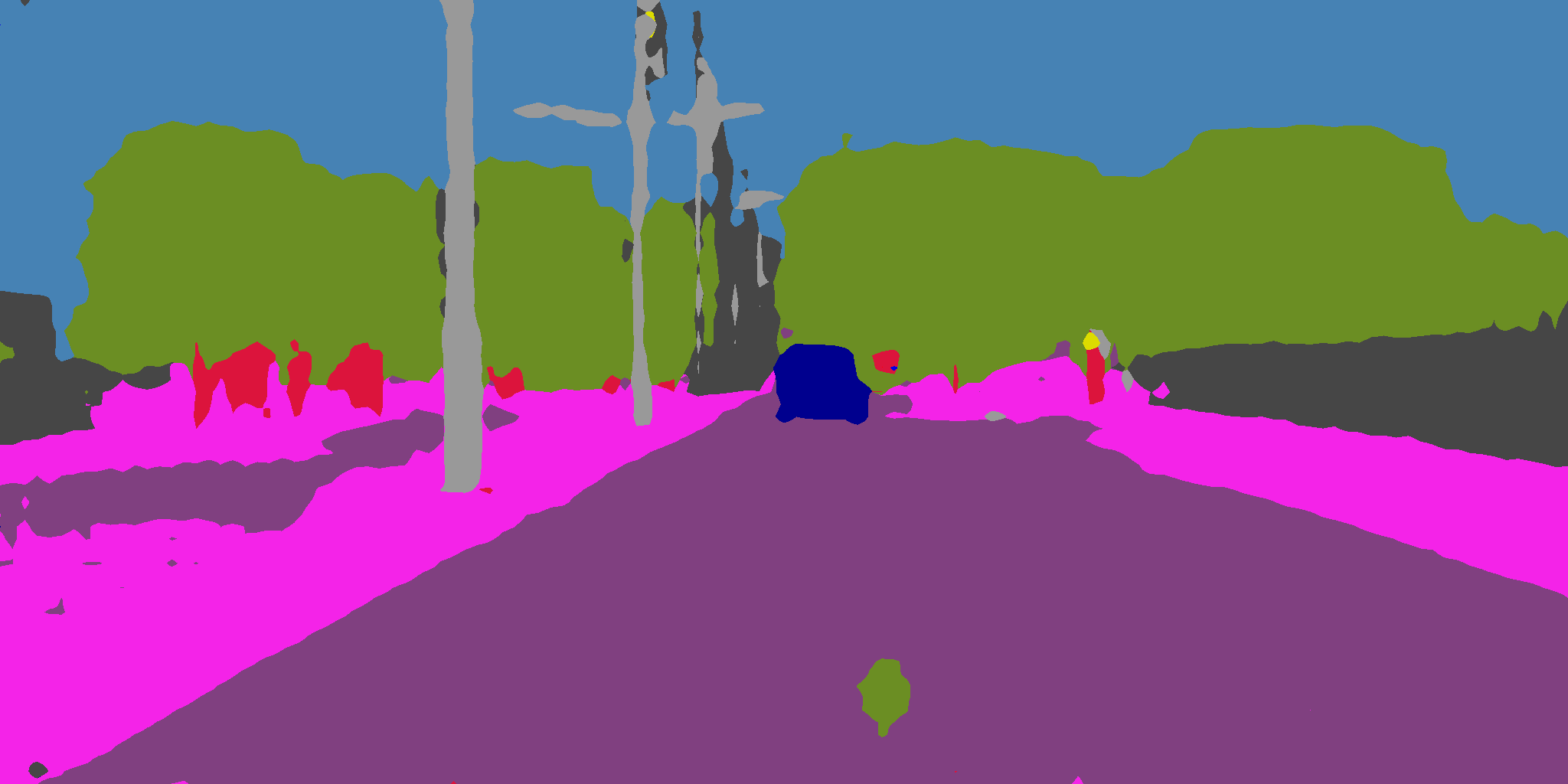} &
    \includegraphics[width=0.23\linewidth]{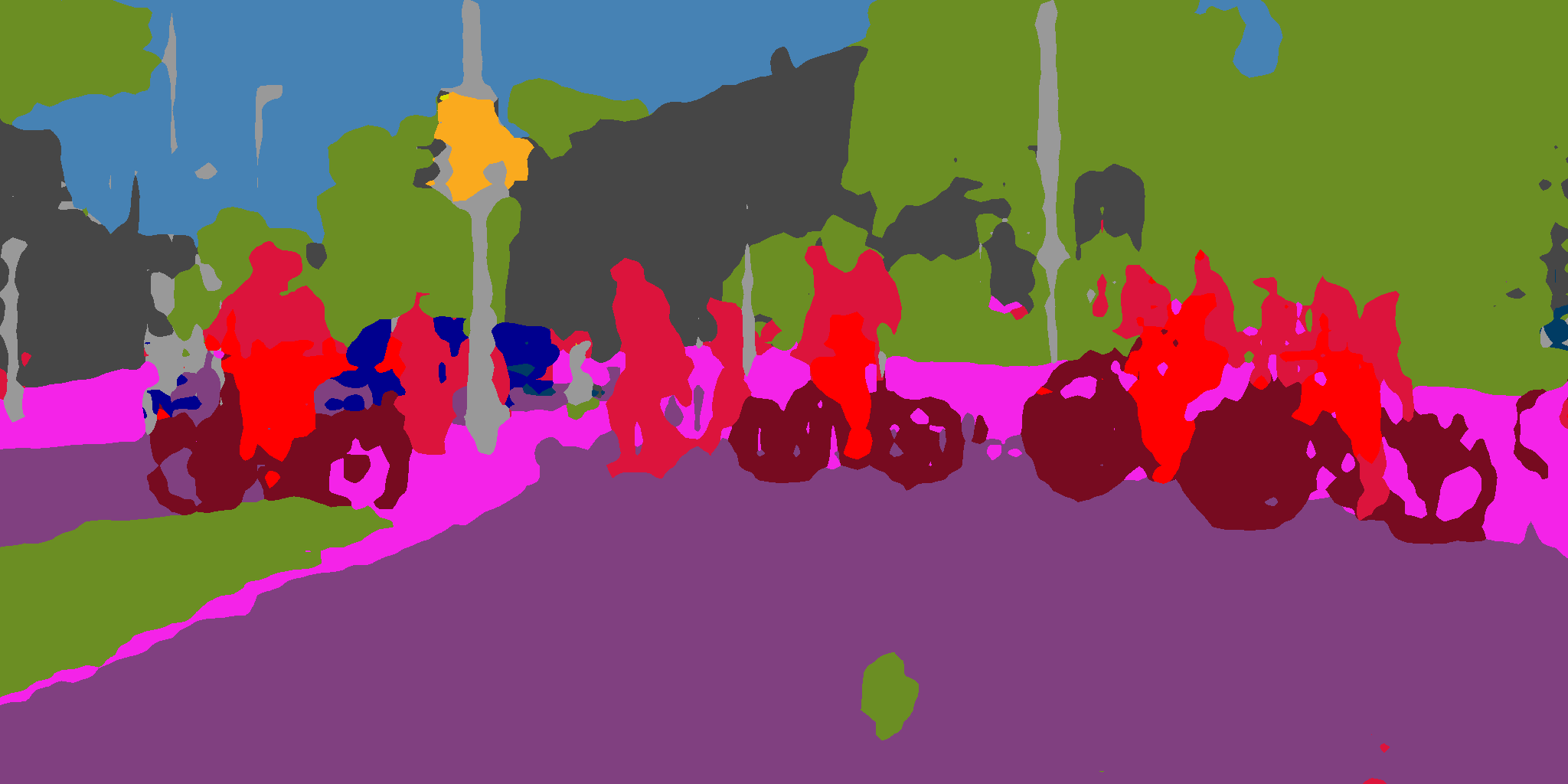} \\
    
    \rotatebox{90}{\ \ \ PRONOUN}  &
    \includegraphics[width=0.23\linewidth]{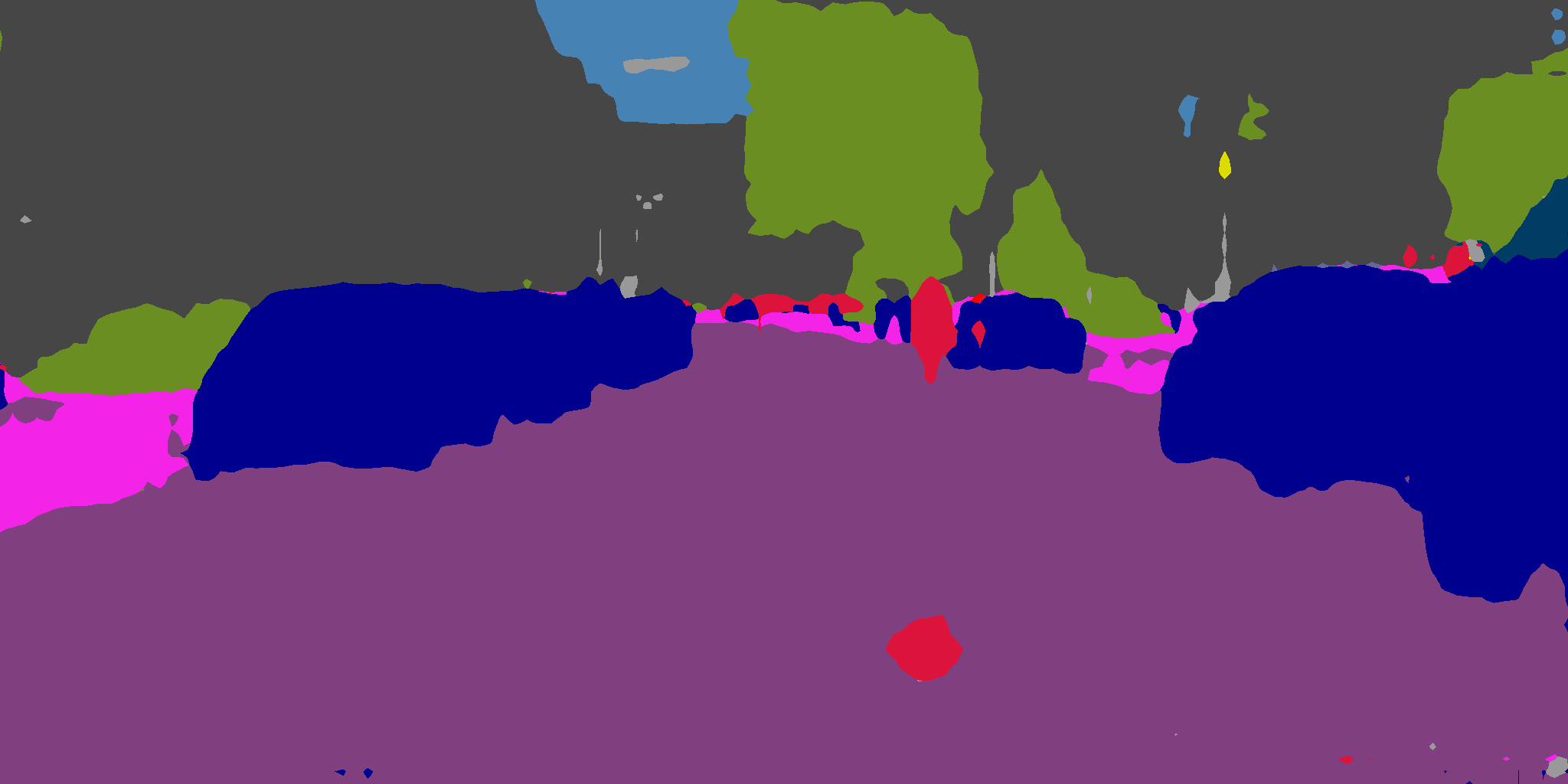} &
    \includegraphics[width=0.23\linewidth]{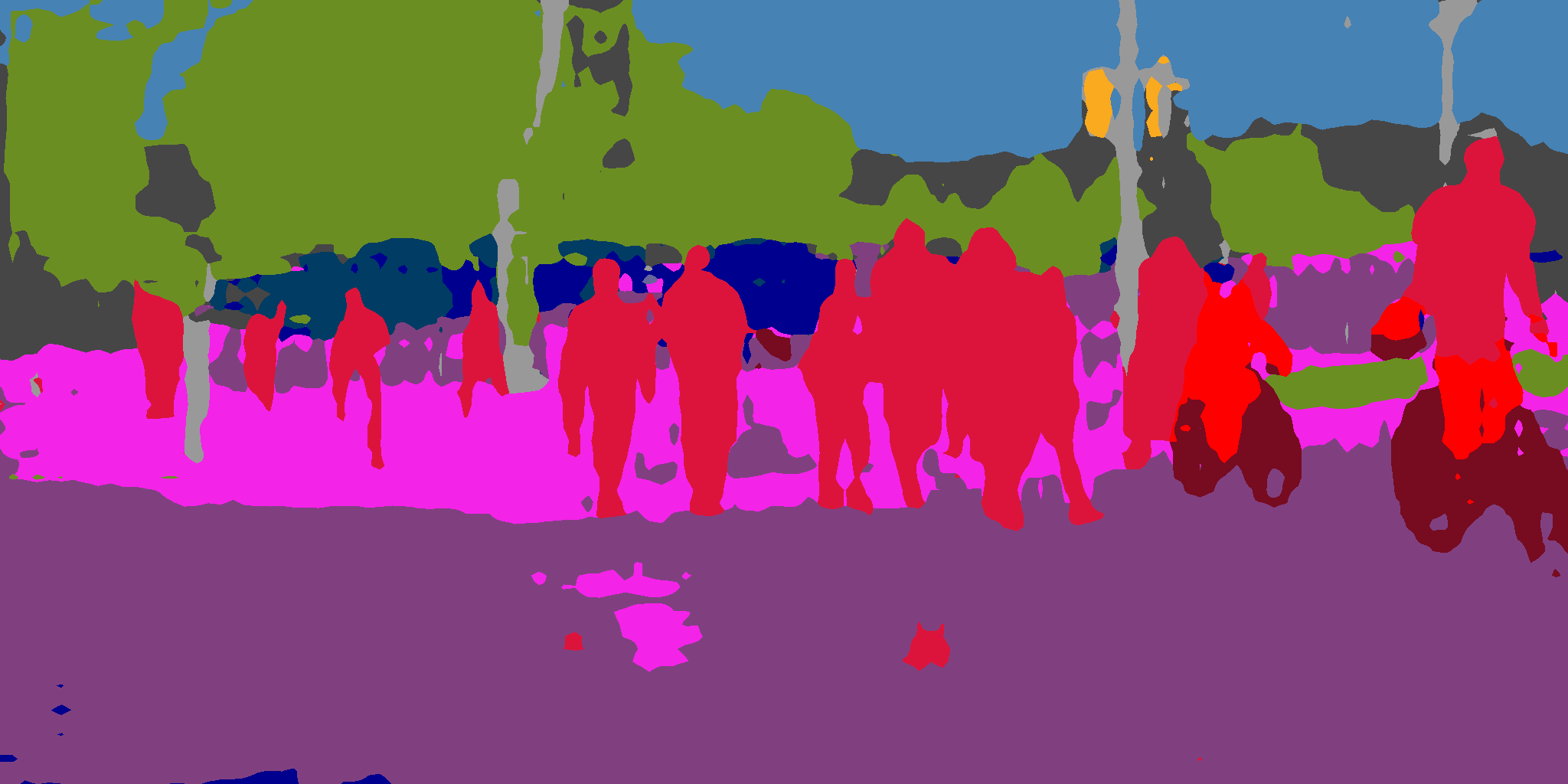} &
    \includegraphics[width=0.23\linewidth]{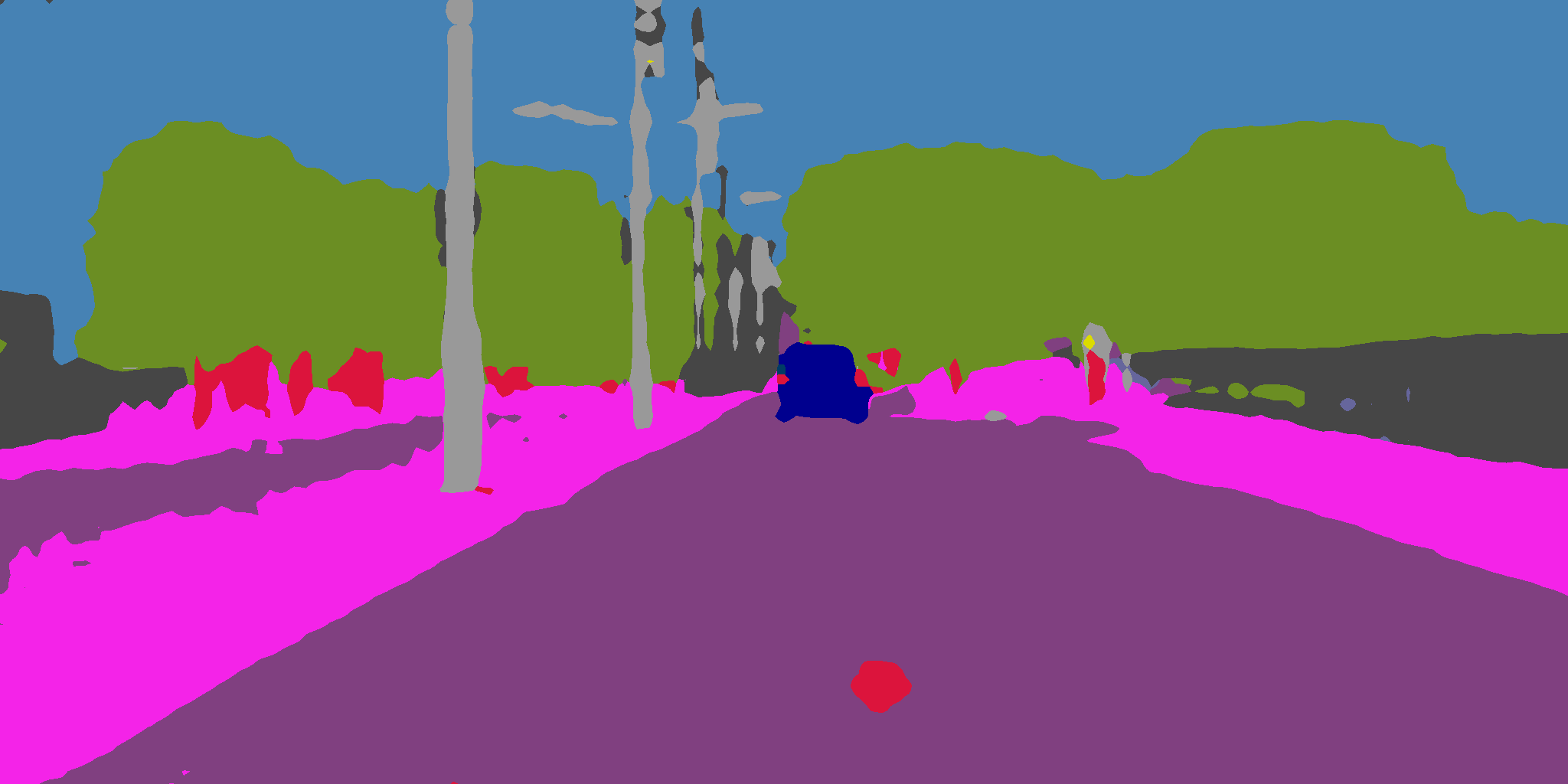} &
    \includegraphics[width=0.23\linewidth]{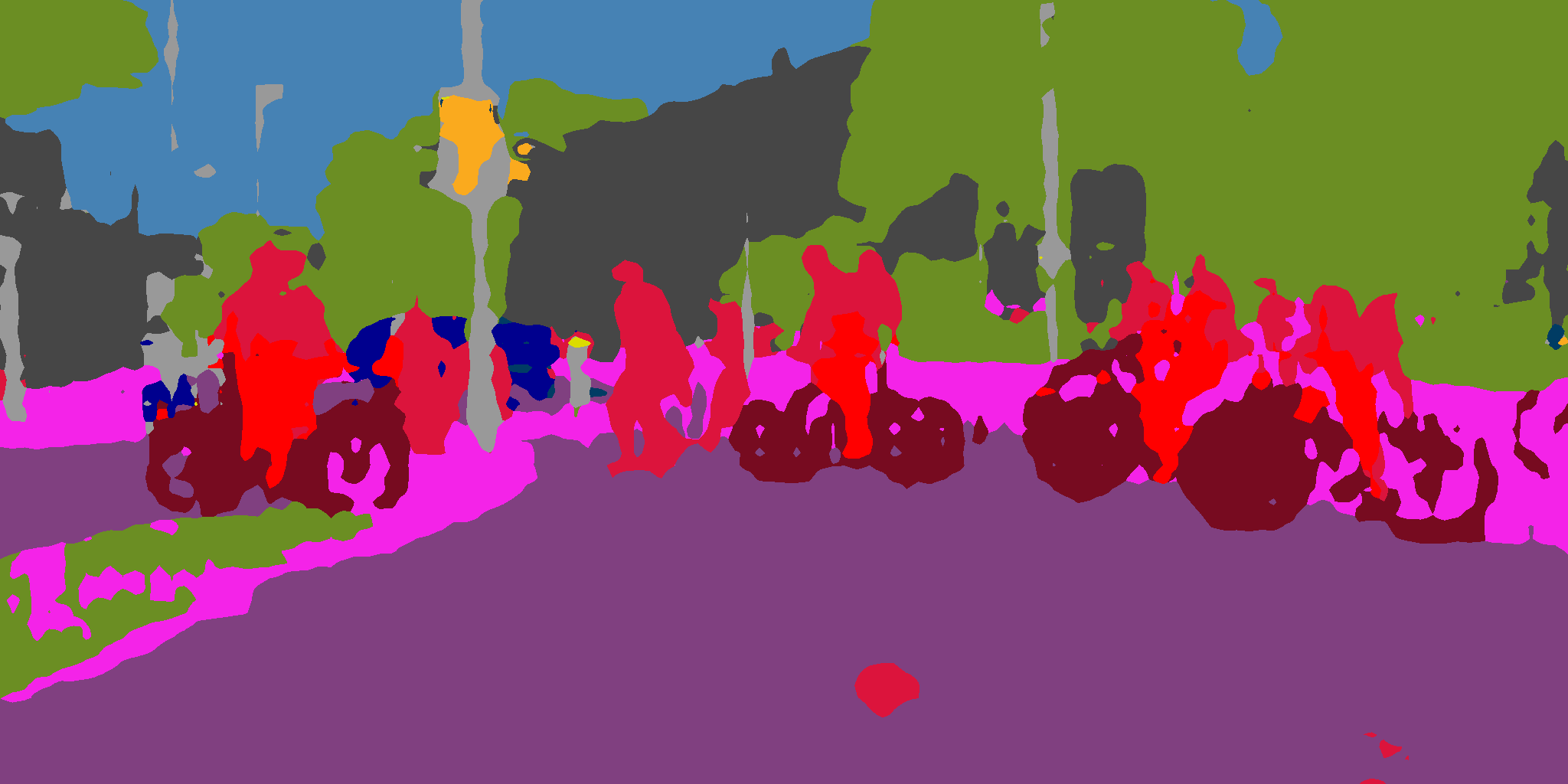} \\
    \end{tabular}
    \caption{Qualitative results of UDA methods on the synthetic-to-real semantic segmentation task of \emph{Synthia}$\to$\emph{Cityscapes}.}
	\label{fig:seg_vis}
\end{figure*}

For object recognition, we follow the \emph{de facto} practice to study the t-SNE~\cite{maaten2008visualizing} visualization of aligned features generated by different UDA methods in Fig.~\ref{fig:tsne}.
As expected, DANN-[f, p] fails to align multi-class distributions between domains. With effective conditioning, NOUN better aligns target samples to source clusters and learns more discriminative target features. With the prototype-based conditioning strategy, PRONOUN alleviates the misalignment induced by inaccurate pseudo-labels. 
For semantic segmentation, we present some qualitative results in Fig.~\ref{fig:seg_vis}. Compared with the domain adversarial training involving only marginal features (FeatAdapt), our NOUN produces cleaner and more discriminative adapted segmentation results, which verifies the benefit of the effective conditioning for domain adaptation. 

\section{Conclusion} \label{sec:conclusion}
In this work, we target the UDA problem with multi-modal structures. We aim to develop simple also compact domain adversarial training solutions to address UDA.
We first rethink the previous failure of the naive concatenation conditioning strategy. We find that the small norm of the concatenated output prediction makes the conditioning ineffective. We thus propose NOUN to ensure effective conditional domain alignment by enlarging the vector norm of the concatenated prediction. We provide a strong domain adversarial training method PRONOUN, which involves the novel prototype-based conditioning strategy to protect NOUN from inaccurate pseudo-labels.  
Extensive evaluations on various object recognition benchmarks justify the effectiveness of NOUN and PRONOUN over well-established UDA baselines. NOUN is also verified to be genric and competitive for semantic segmentation UDA tasks. In the future, we aim to explore applications of PRONOUN to more challenging UDA tasks like semantic segmentation and object detection.

\section*{Acknowledgements}
The authors would like to thank the reviewers and the associate editor for their valuable comments. The authors also would like to appreciate Jiashi Feng and Shuicheng Yan for their insightful discussions throughout this project and comments on the writing of this paper.

\bibliographystyle{IEEEtran.bst}
\bibliography{noun.bib}

\begin{thebibliography}{10}
\providecommand{\url}[1]{#1}
\csname url@samestyle\endcsname
\providecommand{\newblock}{\relax}
\providecommand{\bibinfo}[2]{#2}
\providecommand{\BIBentrySTDinterwordspacing}{\spaceskip=0pt\relax}
\providecommand{\BIBentryALTinterwordstretchfactor}{4}
\providecommand{\BIBentryALTinterwordspacing}{\spaceskip=\fontdimen2\font plus
\BIBentryALTinterwordstretchfactor\fontdimen3\font minus
  \fontdimen4\font\relax}
\providecommand{\BIBforeignlanguage}[2]{{%
\expandafter\ifx\csname l@#1\endcsname\relax
\typeout{** WARNING: IEEEtran.bst: No hyphenation pattern has been}%
\typeout{** loaded for the language `#1'. Using the pattern for}%
\typeout{** the default language instead.}%
\else
\language=\csname l@#1\endcsname
\fi
#2}}
\providecommand{\BIBdecl}{\relax}
\BIBdecl

\bibitem{long2015learning}
M.~Long, Y.~Cao, J.~Wang, and M.~I. Jordan, ``Learning transferable features
  with deep adaptation networks,'' in \emph{Proc. ICML}, 2015, pp. 97--105.

\bibitem{saito2017asymmetric}
K.~Saito, Y.~Ushiku, and T.~Harada, ``Asymmetric tri-training for unsupervised
  domain adaptation,'' in \emph{Proc. ICML}, 2017, pp. 2988--2997.

\bibitem{hoffman2016fcns}
J.~Hoffman, D.~Wang, F.~Yu, and T.~Darrell, ``Fcns in the wild: Pixel-level
  adversarial and constraint-based adaptation,'' \emph{arXiv preprint
  arXiv:1612.02649}, 2016.

\bibitem{adaptseg}
Y.-H. Tsai, W.-C. Hung, S.~Schulter, K.~Sohn, M.-H. Yang, and M.~Chandraker,
  ``Learning to adapt structured output space for semantic segmentation,'' in
  \emph{Proc. CVPR}, 2018, pp. 7472--7481.

\bibitem{siban}
Y.~Luo, P.~Liu, T.~Guan, J.~Yu, and Y.~Yang, ``Significance-aware information
  bottleneck for domain adaptive semantic segmentation,'' in \emph{Proc. ICCV},
  2019, pp. 6778--6787.

\bibitem{clan}
Y.~Luo, L.~Zheng, T.~Guan, J.~Yu, and Y.~Yang, ``Taking a closer look at domain
  shift: Category-level adversaries for semantics consistent domain
  adaptation,'' in \emph{Proc. CVPR}, 2019, pp. 2507--2516.

\bibitem{asm}
Y.~Luo, P.~Liu, T.~Guan, J.~Yu, and Y.~Yang, ``Adversarial style mining for
  one-shot unsupervised domain adaptation,'' in \emph{Proc. NeurIPS}, 2020, pp.
  20\,612--20\,623.

\bibitem{gretton2008kernel}
A.~Gretton, K.~M. Borgwardt, M.~J. Rasch, B.~Sch{\"o}lkopf, and A.~Smola, ``A
  kernel method for the two-sample problem,'' \emph{Journal of Machine Learning
  Research}, vol.~1, pp. 1--10, 2008.

\bibitem{sun2016deep}
B.~Sun and K.~Saenko, ``Deep coral: Correlation alignment for deep domain
  adaptation,'' in \emph{Proc. ECCV}, 2016, pp. 443--450.

\bibitem{zellinger2017central}
W.~Zellinger, T.~Grubinger, E.~Lughofer, T.~Natschl{\"a}ger, and
  S.~Saminger-Platz, ``Central moment discrepancy (cmd) for domain-invariant
  representation learning,'' in \emph{Proc. ICLR}, 2017.

\bibitem{koniusz2017domain}
P.~Koniusz, Y.~Tas, and F.~Porikli, ``Domain adaptation by mixture of
  alignments of second-or higher-order scatter tensors,'' in \emph{Proc. CVPR},
  2017, pp. 7139--7148.

\bibitem{goodfellow2014generative}
I.~Goodfellow, J.~Pouget-Abadie, M.~Mirza, B.~Xu, D.~Warde-Farley, S.~Ozair,
  A.~Courville, and Y.~Bengio, ``Generative adversarial nets,'' in \emph{Proc.
  NeurIPS}, 2014, pp. 2672--2680.

\bibitem{madacls}
Z.~Pei, Z.~Cao, M.~Long, and J.~Wang, ``Multi-adversarial domain adaptation,''
  in \emph{Proc. AAAI}, 2018.

\bibitem{cdan}
M.~Long, Z.~Cao, J.~Wang, and M.~I. Jordan, ``Conditional adversarial domain
  adaptation,'' in \emph{Proc. NeurIPS}, 2018, pp. 1647--1657.

\bibitem{idda}
V.~K. Kurmi and V.~P. Namboodiri, ``Looking back at labels: A class based
  domain adaptation technique,'' in \emph{Proc. IJCNN}, 2019, pp. 1--8.

\bibitem{rca}
S.~Cicek and S.~Soatto, ``Unsupervised domain adaptation via regularized
  conditional alignment,'' in \emph{Proc. ICCV}, 2019, pp. 1416--1425.

\bibitem{dannca}
L.~Tran, K.~Sohn, X.~Yu, X.~Liu, and M.~Chandraker, ``Gotta adapt'em all: Joint
  pixel and feature-level domain adaptation for recognition in the wild,'' in
  \emph{Proc. CVPR}, 2019, pp. 2672--2681.

\bibitem{madaseg}
Y.-H. Chen, W.-Y. Chen, Y.-T. Chen, B.-C. Tsai, Y.-C. Frank~Wang, and M.~Sun,
  ``No more discrimination: Cross city adaptation of road scene segmenters,''
  in \emph{Proc. ICCV}, 2017, pp. 1992--2001.

\bibitem{adaptpatch}
Y.-H. Tsai, K.~Sohn, S.~Schulter, and M.~Chandraker, ``Domain adaptation for
  structured output via discriminative patch representations,'' \emph{arXiv
  preprint arXiv:1901.05427v1}, 2019.

\bibitem{dann}
Y.~Ganin and V.~Lempitsky, ``Unsupervised domain adaptation by
  backpropagation,'' in \emph{Proc. ICML}, 2015, pp. 1180--1189.

\bibitem{dannjmlr}
Y.~Ganin, E.~Ustinova, H.~Ajakan, P.~Germain, H.~Larochelle, F.~Laviolette,
  M.~Marchand, and V.~Lempitsky, ``Domain-adversarial training of neural
  networks,'' \emph{Journal of Machine Learning Research (JMLR)}, vol.~17,
  no.~1, pp. 2096--2030, 2016.

\bibitem{shimodaira2000improving}
H.~Shimodaira, ``Improving predictive inference under covariate shift by
  weighting the log-likelihood function,'' \emph{Journal of Statistical
  Planning and Inference}, vol.~90, no.~2, pp. 227--244, 2000.

\bibitem{dudik2006correcting}
M.~Dud{\'\i}k, S.~J. Phillips, and R.~E. Schapire, ``Correcting sample
  selection bias in maximum entropy density estimation,'' in \emph{Proc.
  NeurIPS}, 2006, pp. 323--330.

\bibitem{huang2007correcting}
J.~Huang, A.~Gretton, K.~Borgwardt, B.~Sch{\"o}lkopf, and A.~J. Smola,
  ``Correcting sample selection bias by unlabeled data,'' in \emph{Proc.
  NeurIPS}, 2007, pp. 601--608.

\bibitem{gong2012geodesic}
B.~Gong, Y.~Shi, F.~Sha, and K.~Grauman, ``Geodesic flow kernel for
  unsupervised domain adaptation,'' in \emph{Proc. CVPR}, 2012, pp. 2066--2073.

\bibitem{fernando2013unsupervised}
B.~Fernando, A.~Habrard, M.~Sebban, and T.~Tuytelaars, ``Unsupervised visual
  domain adaptation using subspace alignment,'' in \emph{Proc. ICCV}, 2013, pp.
  2960--2967.

\bibitem{sun2016return}
B.~Sun, J.~Feng, and K.~Saenko, ``Return of frustratingly easy domain
  adaptation,'' in \emph{Proc. AAAI}, 2016, p. 2058–2065.

\bibitem{pan2010domain}
S.~J. Pan, I.~W. Tsang, J.~T. Kwok, and Q.~Yang, ``Domain adaptation via
  transfer component analysis,'' \emph{IEEE Transactions on Neural Networks},
  vol.~22, no.~2, pp. 199--210, 2010.

\bibitem{long2013transfer}
M.~Long, J.~Wang, G.~Ding, J.~Sun, and P.~S. Yu, ``Transfer feature learning
  with joint distribution adaptation,'' in \emph{Proc. ICCV}, 2013, pp.
  2200--2207.

\bibitem{herath2017learning}
S.~Herath, M.~Harandi, and F.~Porikli, ``Learning an invariant hilbert space
  for domain adaptation,'' in \emph{Proc. CVPR}, 2017, pp. 3956--3965.

\bibitem{adda}
E.~Tzeng, J.~Hoffman, K.~Saenko, and T.~Darrell, ``Adversarial discriminative
  domain adaptation,'' in \emph{Proc. CVPR}, 2017, pp. 2962--2971.

\bibitem{saito2018maximum}
K.~Saito, K.~Watanabe, Y.~Ushiku, and T.~Harada, ``Maximum classifier
  discrepancy for unsupervised domain adaptation,'' in \emph{Proc. CVPR}, 2018,
  pp. 3723--3732.

\bibitem{tzeng2014deep}
E.~Tzeng, J.~Hoffman, N.~Zhang, K.~Saenko, and T.~Darrell, ``Deep domain
  confusion: Maximizing for domain invariance,'' \emph{arXiv preprint
  arXiv:1412.3474}, 2014.

\bibitem{long2017deep}
M.~Long, H.~Zhu, J.~Wang, and M.~I. Jordan, ``Deep transfer learning with joint
  adaptation networks,'' in \emph{Proc. ICML}, 2017, pp. 2208--2217.

\bibitem{lee2013pseudo}
D.-H. Lee \emph{et~al.}, ``Pseudo-label: The simple and efficient
  semi-supervised learning method for deep neural networks,'' in \emph{Workshop
  on challenges in representation learning, ICML}, 2013, p. 896.

\bibitem{li2019bidirectional}
Y.~Li, L.~Yuan, and N.~Vasconcelos, ``Bidirectional learning for domain
  adaptation of semantic segmentation,'' in \emph{Proc. CVPR}, 2019, pp.
  6936--6945.

\bibitem{atdoc}
J.~Liang, D.~Hu, and J.~Feng, ``Domain adaptation with auxiliary target
  domain-oriented classifier,'' in \emph{Proc. CVPR}, 2021, pp.
  16\,632--16\,642.

\bibitem{choi2019pseudo}
J.~Choi, M.~Jeong, T.~Kim, and C.~Kim, ``Pseudo-labeling curriculum for
  unsupervised domain adaptation,'' in \emph{Proc. BMVC}, 2019, p.~67.

\bibitem{zou2018unsupervised}
Y.~Zou, Z.~Yu, B.~Vijaya~Kumar, and J.~Wang, ``Unsupervised domain adaptation
  for semantic segmentation via class-balanced self-training,'' in \emph{Proc.
  ECCV}, 2018, pp. 297--313.

\bibitem{zhang2017joint}
J.~Zhang, W.~Li, and P.~Ogunbona, ``Joint geometrical and statistical alignment
  for visual domain adaptation,'' in \emph{Proc. CVPR}, 2017, pp. 1859--1867.

\bibitem{dada}
H.~Tang and K.~Jia, ``Discriminative adversarial domain adaptation,'' in
  \emph{Proc. AAAI}, 2020, pp. 5940--5947.

\bibitem{isola2017image}
P.~Isola, J.-Y. Zhu, T.~Zhou, and A.~A. Efros, ``Image-to-image translation
  with conditional adversarial networks,'' in \emph{Proc. CVPR}, 2017, pp.
  1125--1134.

\bibitem{odena2017conditional}
A.~Odena, C.~Olah, and J.~Shlens, ``Conditional image synthesis with auxiliary
  classifier gans,'' in \emph{Proc. ICML}, 2017, pp. 2642--2651.

\bibitem{he2016deep}
K.~He, X.~Zhang, S.~Ren, and J.~Sun, ``Deep residual learning for image
  recognition,'' in \emph{Proc. CVPR}, 2016, pp. 770--778.

\bibitem{dwt-mec}
S.~Roy, A.~Siarohin, E.~Sangineto, S.~R. Bulo, N.~Sebe, and E.~Ricci,
  ``Unsupervised domain adaptation using feature-whitening and consensus
  loss,'' in \emph{Proc. CVPR}, 2019, pp. 9471--9480.

\bibitem{safn}
R.~Xu, G.~Li, J.~Yang, and L.~Lin, ``Larger norm more transferable: An adaptive
  feature norm approach for unsupervised domain adaptation,'' in \emph{Proc.
  ICCV}, 2019, pp. 1426--1435.

\bibitem{adr}
K.~Saito, Y.~Ushiku, T.~Harada, and K.~Saenko, ``Adversarial dropout
  regularization,'' in \emph{Proc. ICLR}, 2018.

\bibitem{bsp}
X.~Chen, S.~Wang, M.~Long, and J.~Wang, ``Transferability vs. discriminability:
  Batch spectral penalization for adversarial domain adaptation,'' in
  \emph{Proc. ICML}, 2019, pp. 1081--1090.

\bibitem{swd}
C.-Y. Lee, T.~Batra, M.~H. Baig, and D.~Ulbricht, ``Sliced wasserstein
  discrepancy for unsupervised domain adaptation,'' in \emph{Proc. CVPR}, 2019,
  pp. 10\,285--10\,295.

\bibitem{venkateswara2017Deep}
H.~Venkateswara, J.~Eusebio, S.~Chakraborty, and S.~Panchanathan, ``Deep
  hashing network for unsupervised domain adaptation,'' in \emph{Proc. CVPR},
  2017, pp. 5018--5027.

\bibitem{peng2017visda}
X.~Peng, B.~Usman, N.~Kaushik, J.~Hoffman, D.~Wang, and K.~Saenko, ``Visda: The
  visual domain adaptation challenge,'' \emph{arXiv preprint arXiv:1710.06924},
  2017.

\bibitem{saenko2010adapting}
K.~Saenko, B.~Kulis, M.~Fritz, and T.~Darrell, ``Adapting visual category
  models to new domains,'' in \emph{Proc. ECCV}, 2010, pp. 213--226.

\bibitem{richter2016playing}
S.~R. Richter, V.~Vineet, S.~Roth, and V.~Koltun, ``Playing for data: Ground
  truth from computer games,'' in \emph{Proc. ECCV}, 2016, pp. 102--118.

\bibitem{cordts2016cityscapes}
M.~Cordts, M.~Omran, S.~Ramos, T.~Rehfeld, M.~Enzweiler, R.~Benenson,
  U.~Franke, S.~Roth, and B.~Schiele, ``The cityscapes dataset for semantic
  urban scene understanding,'' in \emph{Proc. CVPR}, 2016, pp. 3213--3223.

\bibitem{ros2016synthia}
G.~Ros, L.~Sellart, J.~Materzynska, D.~Vazquez, and A.~M. Lopez, ``The synthia
  dataset: A large collection of synthetic images for semantic segmentation of
  urban scenes,'' in \emph{Proc. CVPR}, 2016, pp. 3234--3243.

\bibitem{lin2014microsoft}
T.-Y. Lin, M.~Maire, S.~Belongie, J.~Hays, P.~Perona, D.~Ramanan,
  P.~Doll{\'a}r, and C.~L. Zitnick, ``Microsoft coco: Common objects in
  context,'' in \emph{Proc. ECCV}, 2014, pp. 740--755.

\bibitem{zhang2017curriculum}
Y.~Zhang, P.~David, and B.~Gong, ``Curriculum domain adaptation for semantic
  segmentation of urban scenes,'' in \emph{Proc. ICCV}, 2017, pp. 2020--2030.

\bibitem{nair2010rectified}
V.~Nair and G.~E. Hinton, ``Rectified linear units improve restricted boltzmann
  machines,'' in \emph{Proc. ICML}, 2010, pp. 807--814.

\bibitem{chen2017deeplab}
L.-C. Chen, G.~Papandreou, I.~Kokkinos, K.~Murphy, and A.~L. Yuille, ``Deeplab:
  Semantic image segmentation with deep convolutional nets, atrous convolution,
  and fully connected crfs,'' \emph{IEEE Transactions on Pattern Analysis and
  Machine Intelligence (TPAMI)}, vol.~40, no.~4, pp. 834--848, 2017.

\bibitem{advent}
T.-H. Vu, H.~Jain, M.~Bucher, M.~Cord, and P.~P{\'e}rez, ``Advent: Adversarial
  entropy minimization for domain adaptation in semantic segmentation,'' in
  \emph{Proc. CVPR}, 2019, pp. 2517--2526.

\bibitem{radford2015unsupervised}
A.~Radford, L.~Metz, and S.~Chintala, ``Unsupervised representation learning
  with deep convolutional generative adversarial networks,'' \emph{arXiv
  preprint arXiv:1511.06434}, 2015.

\bibitem{maas2013rectifier}
A.~L. Maas, A.~Y. Hannun, and A.~Y. Ng, ``Rectifier nonlinearities improve
  neural network acoustic models,'' in \emph{Proc. ICML}, 2013.

\bibitem{bottou2010large}
L.~Bottou, ``Large-scale machine learning with stochastic gradient descent,''
  in \emph{International Conference on Computational Statistics}.\hskip 1em
  plus 0.5em minus 0.4em\relax Springer, 2010, pp. 177--186.

\bibitem{kingma2014adam}
D.~P. Kingma and J.~Ba, ``Adam: A method for stochastic optimization,'' in
  \emph{Proc. ICLR}, 2015.

\bibitem{morerio2017minimal}
P.~Morerio, J.~Cavazza, and V.~Murino, ``Minimal-entropy correlation alignment
  for unsupervised deep domain adaptation,'' in \emph{Proc. ICLR}, 2018.

\bibitem{cat}
Z.~Deng, Y.~Luo, and J.~Zhu, ``Cluster alignment with a teacher for
  unsupervised domain adaptation,'' in \emph{Proc. ICCV}, 2019, pp. 9944--9953.

\bibitem{ican}
W.~Zhang, W.~Ouyang, W.~Li, and D.~Xu, ``Collaborative and adversarial network
  for unsupervised domain adaptation,'' in \emph{Proc. CVPR}, 2018, pp.
  3801--3809.

\bibitem{maaten2008visualizing}
L.~v.~d. Maaten and G.~Hinton, ``Visualizing data using t-sne,'' \emph{Journal
  of Machine Learning Research (JMLR)}, vol.~9, no. Nov, pp. 2579--2605, 2008.

\end{thebibliography}

\end{document}